\documentclass[journal,transmag]{IEEEtran}
\usepackage{amsmath,graphicx}
\usepackage{amssymb}
\usepackage{hyperref}
\usepackage[ruled]{algorithm2e}
\usepackage{algorithmic}
\usepackage{subfig}
\begin{document}

\title{Fast Steerable Principal Component Analysis}


\author{\IEEEauthorblockN{Zhizhen Zhao
\IEEEauthorrefmark{1},
Yoel Shkolnisky \IEEEauthorrefmark{2},
Amit Singer \IEEEauthorrefmark{3}
}
\IEEEauthorblockA{\IEEEauthorrefmark{1}Courant Institute of Mathematical Sciences, New York University, New York, NY, USA}
\IEEEauthorblockA{\IEEEauthorrefmark{2}Department of Applied Mathematics, School of Mathematical Sciences, Tel Aviv University, Tel Aviv, Israel}
\IEEEauthorblockA{\IEEEauthorrefmark{3}Department of Mathematics and Program in Applied and Computational Mathematics, Princeton University, Princeton, NJ, USA}
\thanks{
Corresponding author: Z. Zhao (email: jzhao@cims.nyu.edu).}}

\IEEEtitleabstractindextext{%
\begin{abstract}
Cryo-electron microscopy nowadays often requires the analysis of hundreds of thousands of 2D images as large as a few hundred pixels in each direction. Here we introduce an algorithm that efficiently and accurately performs principal component analysis (PCA) for a large set of two-dimensional images, and, for each image, the set of its uniform rotations in the plane and their reflections. For a dataset consisting of $n$ images of size $L \times L$ pixels, the computational complexity of our algorithm is $O(nL^3 + L^4)$, while existing algorithms take $O(nL^4)$. The new algorithm computes the expansion coefficients of the images in a Fourier-Bessel basis efficiently using the non-uniform fast Fourier transform. We compare the accuracy and efficiency of the new algorithm with traditional PCA and existing algorithms for steerable PCA.
\end{abstract}

}

\maketitle

\IEEEdisplaynontitleabstractindextext
\IEEEpeerreviewmaketitle

\section{Introduction}
Principal component analysis (PCA) is widely used in image analysis and pattern recognition for dimensionality reduction and denoising. In particular, PCA is often one of the first steps~\cite{vanHeel} in the algorithmic pipeline of  cryo-electron microscopy (cryo-EM) single particle reconstruction (SPR)~\cite{Frank} to compress and denoise the acquired 2D projection images in order to eventually determine the 3D structure of a macromolecule. The high level of noise in those images drastically deteriorates the performance of single-image based denoising methods, such as non-local means~\cite{Morel2005} and wavelet thresholding~\cite{Donoho1995}, and so the latter are outperformed by PCA. As any planar rotation of any given projection image is equally likely to appear in the experiment, by either in-plane rotating the detector or the specimen, it makes sense to include all possible rotations of the projection images when performing PCA. The resulting decomposition, termed steerable PCA, consists of principal components which are tensor products of radial functions and angular Fourier modes \cite{Hilai,Perona,Uenohara,Ponce,Zhao13}. Beyond cryo-EM, steerable PCA has many other applications in image analysis and computer vision \cite{Vonesch2013}.

The term ``steerable PCA'' comes from the fact that rotating the principal components is achieved by a simple phase shift of their angular part. The principal components are invariant to any in-plane rotation of the images, therefore finding steerable principal components is equivalent to finding in-plane rotationally invariant principal components.

In cryo-EM data processing, in addition to compression and denoising, steerable PCA is also useful in generating rotationally invariant image features (i.e. bispectrum-like features~\cite{Zhao14}). These are crucial for fast rotationally invariant nearest neighbors search used in efficient computation of class averages~\cite{Zhao14}.
Rotational alignment between image pairs can also be computed more efficiently using the expansion coefficients in a steerable basis.

In this paper, we focus on the action of the group $O(2)$ on digital images by in-plane rotating and possibly reflecting them. The idea of using group actions for constructing group invariant features and filters has been previously proposed in~\cite{Lenz89, Lenz90}. 
This group theoretical framework has been applied to $SO(3)$ and $SU(1,1)$ in~\cite{Lenz98, Lenz05}. The representation of finite groups, such as the dihedral groups, has been used for computing the Karhunen-Lo{\'e}ve expansion of digital images in~\cite{Lenz95}.

Various efficient algorithms for steerable PCA have been introduced~\cite{Jogan, Ponce}. However, steerable PCA of modern cryo-EM datasets that contain hundreds of thousands of large images poses a computational challenge. Also, it is important to ensure that the steerable PCA algorithm is numerically accurate when the input images are noisy. In order to exploit the special separation of variables structure of the principal components in polar coordinates, most algorithms rely on resampling the images on a polar grid.
However, the transformation from Cartesian to polar is non-unitary, and thus changes the statistics of the noise. In particular, resampling transforms uncorrelated white noise to colored noise that may lead to spurious principal components.

Recently,~\cite{Zhao13} addressed this issue by incorporating a sampling criterion into the steerable PCA framework and introduced an algorithm called Fourier-Bessel steerable PCA (FBsPCA). FBsPCA assumes that the underlying clean images (before being possibly contaminated with noise) are bandlimited and essentially compactly supported in a disk. This assumption holds, for example, for 2D projection images of a 3D molecule compactly supported in a ball. It also implies that the images can be expanded in an orthogonal basis for bandlimited functions, such as the Fourier-Bessel basis.
In FBsPCA, the Fourier-Bessel expansion of each image is truncated into a finite series using a sampling criterion that was introduced by Klug and Crowther~\cite{Klug}.
The sampling criterion ensures that the transformation from the Cartesian grid to the truncated Fourier-Bessel expansion is nearly unitary. Moreover, the covariance matrix built from the expansion coefficients of the images and all their possible rotations has a block diagonal structure where the block size decreases as a function of the angular frequency.
The computational complexity of FBsPCA is $O(nL^4)$ operations for $n$ images of size $L \times L$. Notice that, when $n > L^2$, the computational complexity of traditional PCA is $O(nL^4 + L^6)$, where the first term corresponds to forming the $L^2\times L^2$ covariance matrix and the second term corresponds to its eigen-decomposition. 
Although FBsPCA and PCA have a similar computational complexity, FBsPCA leads to better denoising as it takes into account all possible rotations and reflections. This makes FBsPCA more suitable than traditional PCA as a tool for 2D analysis of cryo-EM images~\cite{Zhao13}.
With the enhancement of electron microscope detectors' resolution, a typical image size of a single particle can easily be over $300\times 300$ pixels. 
Thus, FBsPCA is still 
not efficient enough to analyze a large number of images of large size (i.e. large $n$ and large $L$). The bottleneck for this algorithm is 
the first step that computes the Fourier-Bessel expansion coefficients, 
whose computational complexity is $O(nL^4)$.

In this paper we introduce a fast Fourier-Bessel steerable PCA (FFBsPCA) that reduces the computational complexity for FBsPCA from $O(nL^4)$ to $O(nL^3)$ by computing the Fourier-Bessel expansion coefficients more efficiently and accurately. This is achieved by first mapping the images from their Cartesian grid representation to a polar grid representation in the reciprocal (Fourier) domain using the non-uniform fast Fourier transform (NUFFT)~\cite{Dutt, nufft_fessler, Greengard, Fenn07}. 
The polar grid representation enables to efficiently evaluate the Fourier-Bessel expansion coefficients of the images by 1D FFT on concentric circles followed by accurate evaluation of a radial integral with a Gaussian quadrature rule. The overall complexity of computing the Fourier-Bessel coefficients is reduced to $O(nL^3)$ operations. The increased accuracy and efficiency in evaluating the Fourier-Bessel expansion coefficients are the main contributions of this paper.

We note that the Fourier-Bessel expansion coefficients can be computed in $O(nL^2\log L)$ operations using algorithms for rapid evaluation of special functions~\cite{ONeil10} or a fast analysis-based Fourier-Bessel expansion~\cite{Townsend15}.
However, such ``fast'' algorithms may only lead to a marginal improvement for two reasons. First, the break even point for them compared to the direct approach is for relatively large $L$ such as $L=256$ or larger. Second, forming the covariance matrix from the expansion coefficients still requires $O(nL^3)$ operations.

The paper is organized as follows: Section~\ref{sec:FBT} contains the mathematical preliminaries of the Fourier-Bessel expansion, the sampling criterion, and the numerical evaluation of the expansion coefficients. The computation of the steerable principal components is described in Section~\ref{sec:cov}. We present the algorithm and give a detailed computational complexity analysis in Section~\ref{sec:alg}. Various numerical examples concerning the computation time of FFBsPCA compared with FBsPCA and traditional PCA are presented in Section~\ref{sec:results}. In the same section, we demonstrate the performance of FFBsPCA-based denoising using simulated cryo-EM projection images.

Reproducible research: The FFBsPCA is available in the SPR toolbox ASPIRE (\href{http://spr.math.princeton.edu/}{http://spr.math.princeton.edu/}). There are two main functions: \textit{FBCoeff} computes the Fourier Bessel expansion coefficients and \textit{sPCA} computes the steerable PCA basis and the associated expansion coefficients.

\section{Fourier-Bessel Expansion of Bandlimited Images}
\label{sec:FBT}
We say that $f$ has a band limit radius $c$ if its Fourier transform
\begin{equation}
\mathcal{F}(f)(\xi_1,\xi_2) = \int_{\mathbb{R}^2} f(x,y) e^{-2\pi \imath (x \xi_1 + y\xi_2)} \,dx\,dy
\end{equation}
satisfies $\mathcal{F}(f)(\xi_1,\xi_2) = 0$, for $\xi_1^2 + \xi_2^2 > c^2$. In our setup, a digital image $I$ is obtained by sampling a squared-integrable bandlimited function $f$ on a Cartesian grid of size $L \times L$, that is, $I(i_1, i_2) = f(i_1 \Delta, i_2 \Delta)$,
where $i_1, i_2 = -\left\lceil \frac{L-1}{2} \right\rceil, \dots, \left\lfloor \frac{L-1}{2} \right\rfloor$,  and $\Delta$ is the pixel size.

For pixel size $\Delta=1$, the Nyquist-Shannon sampling theorem implies that the Fourier transform of $I$ is supported on the square $[-1/2,1/2) \times [-1/2,1/2)$. In many applications, the support size is effectively smaller due to other experimental considerations, for example, the exponentially decaying envelope of the contrast transfer function in electron microscopy.
Thus, we assume that the band limit radius of all images is $0 < c \leq \frac{1}{2}$. The scaled Fourier-Bessel functions are the eigenfunctions of the Laplacian in a disk of radius $c$ with Dirichlet boundary condition and they are given by
\begin{equation}
\label{eq:FB}
\psi_c^{k, q}(\xi,\theta) =
\begin{cases}
N_{k, q}J_k\left( R_{k, q}\frac{\xi}{c} \right)e^{\imath k \theta} , & \xi \leq c,  \\
\hfil 0 ,  & \xi > c,
\end{cases}
\end{equation}
where $(\xi, \theta)$ are polar coordinates in the Fourier domain (i.e., $\xi_1 = \xi \cos \theta$, $\xi_2 = \xi \sin \theta$, $\xi \geq 0$, and $\theta \in [0,2\pi)$); $N_{k, q} = (c \sqrt{\pi} | J_{k+1}( R_{k, q})|)^{-1}$ is the normalization factor; $J_k$ is the Bessel function of the first kind of integer order $k$; and $R_{k, q}$ is the $q^{\mathrm{th}}$ root of the Bessel function $J_k$.
For a function $f$ with band limit $c$ that is also in $L^2(\mathbb{R}^2)\cap L^1(\mathbb{R}^2)$,
\begin{equation}
\mathcal{F}(f)(\xi, \theta) = \sum_{k = -\infty }^{\infty} \sum_{q = 1}^{\infty} a_{k, q} \psi_c^{k, q}(\xi, \theta),
\label{eq:cont_int}
\end{equation}
which converges pointwise. In Section~\ref{subsec:sc}, we derive a finite truncation rule for the Fourier-Bessel expansion in Eq.~\eqref{eq:cont_int}.

\subsection{Sampling criterion}
\label{subsec:sc}
For digital implementations of Eq.~\eqref{eq:cont_int}, we must truncate it to a finite sum, namely to derive a sampling criterion for selecting $k$ and $q$.

With the following convention for the 2D inverse polar Fourier transform of a function $g(\xi, \theta)$,
\begin{equation}
\label{eq:IPFT}
\mathcal{F}^{-1}(g)(r,\phi) = \int_0^{2\pi} \int_0^\infty g(\xi,\theta) e^{2\pi \imath r \xi \cos (\theta-\phi)} \xi \,d\xi\,d\theta,
\end{equation}
the 2D inverse Fourier transform of the Fourier-Bessel functions, denoted $\mathcal{F}^{-1}(\psi_c^{k, q})$, is given in polar coordinates as
\begin{equation}
\label{eq:IFT_FB}
\mathcal{F}^{-1}(\psi_c^{k, q})(r, \phi) = \frac{ 2c \sqrt{\pi} (-1)^q  R_{k, q} J_k (2\pi c r)}{ \imath^k((2\pi c r)^2 - R_{k, q}^2) } e^{\imath k \phi}.
\end{equation}
The maximum of $|\mathcal{F}^{-1}(\psi_c^{k, q})(r, \phi)|$ in~\eqref{eq:IFT_FB} is obtained near the circle $r = \frac{R_{k, q}}{2\pi c}$ and $\mathcal{F}^{-1}(\psi_c^{k, q})(r, \phi)$ vanishes on concentric circles of radii $r = \frac{R_{k, q'}}{2\pi c}$ with $q'\neq q$. The smallest circle with vanishing $\mathcal{F}^{-1}(\psi_c^{k, q})$ that encircles the maximum of $|\mathcal{F}^{-1}(\psi_c^{k, q})|$ is of radius $r = \frac{R_{k, (q+1)}}{2\pi c}$.

We assume that the underlying clean images (before being possibly contaminated with noise) are essentially compactly supported in a disk of radius $R$. Therefore, we should rule out Fourier-Bessel functions for which the maximum of their inverse Fourier transform resides outside a disk of radius $R$. Otherwise, those functions introduce spurious information from noise. Notice that if the maximum is inside the disk, yet the zero after the maximum is outside the disk, then there is a significant spillover of energy outside the disk. We therefore require the more stringent criterion that the zero after the maximum is inside the disk, namely
\begin{equation}
\label{eq:sc}
\frac{R_{k, (q+1)}}{2\pi c}  \leq  R.
\end{equation}
This sampling argument gives a finite truncation rule for the Fourier-Bessel expansion in Eq.~\eqref{eq:cont_int}, that is
\begin{equation}
\label{criterion1}
R_{k, (q+1)} \leq 2\pi c R.
\end{equation}
For each $k$, we denote by $p_k$ the number of components satisfying Eq.~\eqref{criterion1}. We also denote by $p = \sum_{k = -k_{\max}}^{k_{\max}} p_k$ the total number of components, where $k_{\max}$ is the maximal possible value of $k$ satisfying Eq.~\eqref{criterion1}. The locations of Bessel zeros have been extensively studied, for example, in~\cite[p.517-521]{Watson},~\cite[p.370]{Abramowitz},~\cite{Olver54, Elbert, Breen95}. Several lower and upper bounds for Bessel zeros $R_{k, q}$ were proven by Breen in~\cite{Breen95}, such as
\begin{equation}
R_{k, q} > k + \frac{2}{3} | a_{q - 1}|^{3/2},
\label{eq:Rkq_lb}
\end{equation}
where $a_q$ is the $q$th zero of the Airy function, shown to satisfy 
\begin{equation}
\left[\frac{3}{8}\pi(4q - 1.4)\right]^{2/3} < |a_q | < \left[\frac{3}{8}\pi(4q - 0.965)\right]^{2/3}.
\end{equation}
Using the lower bound for $|a_q|$ and the sampling criterion in Eq.~\eqref{criterion1}, we have the following inequality for $k$ and $p_k$,
\begin{equation}
2 \pi c R \approx R_{k, (p_k+1)} > k + \pi p_k - \frac{1.4 \pi}{4}.
\label{eq:lb}
\end{equation}
Breen also obtained
\begin{equation}
R_{k, q} < (\frac{k}{2} + q - \frac{0.965}{4}) \pi,
\label{eq:Rkq_ub}
\end{equation}
so we get another inequality for $k$ and $p_k$,
\begin{equation}
2 \pi c R \approx R_{k, (p_k+1)}  <  \left(\frac{k}{2} + p_k + \frac{3.035}{4} \right)\pi.
\label{eq:ub}
\end{equation}
Combining Eqs.~\eqref{eq:lb} and~\eqref{eq:ub},
we have the following lower and upper bounds for $p_k$,
\begin{equation}
2cR - \frac{k}{2} - \frac{3.035}{4} < p_k < 2cR - \frac{k}{\pi} + \frac{1.4}{4}.
\label{eq:pkbs}
\end{equation}
The bound for the highest angular frequency $k_{\max}$ is determined by setting $p_k = 1$ in Eq.~\eqref{eq:pkbs}, resulting in
\begin{equation}
4cR -3.517 < k_{\max} < 2 \pi c R -2.042.
\label{eq:kmaxbs}
\end{equation}
Equation~\eqref{eq:pkbs} implies that as the angular frequency $k$ increases, the number of components $p_k$ decreases. Moreover, using the lower and upper bounds for $p_k$ and $k_{\max}$ in Eqs.~\eqref{eq:pkbs} and~\eqref{eq:kmaxbs}, we derive that the total number of selected Fourier-Bessel basis functions is between $8(cR)^2$ and $4 \pi (cR)^2$.
When $c$ is the largest possible band limit, i.e. $c = \frac{1}{2}$, the number of basis functions is between $2 R^2$ and $\pi R^2$, where the latter is approximately the number of pixels inside a disk of radius $R$. 
Also, whenever $c = O(1)$ and $R = O(L)$, we get that $p = O(L^2)$ and $k_{\max} = O(L)$.

Because the bandlimited function $f$ is assumed to be essentially compactly supported, the infinite expansion in Eq.~\eqref{eq:cont_int} is approximated by the finite expansion
\begin{equation}
\label{eq:contapprox}
P_{c, R}\mathcal{F}(f)(\xi, \theta) = \sum_{k = -k_{\max}}^{k_{\max}} \sum_{q = 1}^{p_k} a_{k, q} \psi_c^{k, q}(\xi, \theta),
\end{equation}
where $P_{c, R}$ is the orthogonal projection from $L^2(D_c)$ (the space of $L^2$ functions supported on a disk of radius $c$), to the space of functions spanned by a finite number of Fourier-Bessel functions that satisfy~\eqref{criterion1}.

\subsection{Numerical evaluation of Fourier-Bessel expansion coefficients}
\label{subsec:pft}
Previously in~\cite{Zhao13}, the evaluation of the expansion coefficients $a_{k, q}$ of Eq.~\eqref{eq:contapprox} was done by least squares. Let $\Psi$ be the matrix whose entries are evaluations of the Fourier-Bessel functions at the Cartesian grid points, with rows indexed by the grid points and columns indexed by angular and radial frequencies.
Finding the coefficient vector $a$ as the solution to $\min_a \| \Psi a - I \|^2_2$ requires the computation of $\Psi^* I$, which takes $O(p L^2) = O(L^4)$ operations, because $p=O(L^2)$. In general $a = (\Psi^* \Psi)^{-1}\Psi^* I$, but here $\Psi^* \Psi$ is approximately the identity matrix, due to the orthogonality of the Fourier-Bessel functions.

\begin{figure}
\begin{center}
\includegraphics[width = 0.8\columnwidth]{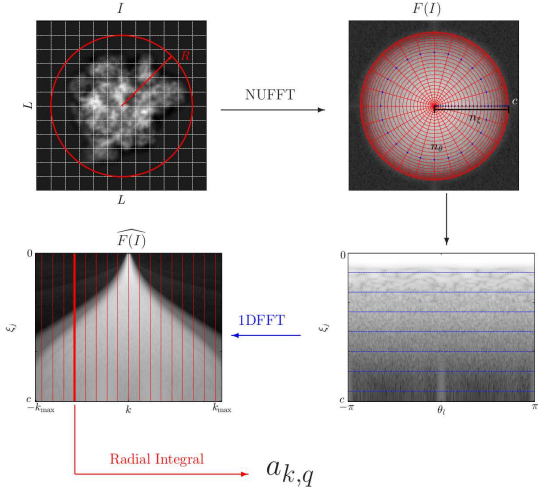}
\end{center}
\caption{Pictorial summary of the procedure for computing the Fourier-Bessel expansion coefficients. The original image (top left) is resampled on a polar Fourier grid (Eq.~\eqref{eq:nufft}) using NUFFT (top right and bottom right) followed by 1D FFT (Eq.~\eqref{eq:1DFFT}) on each concentric circle.
The evaluation of the radial integral (Eq.~\eqref{eq:a}) gives the expansion coefficients $a_{k, q}$. The bow-tie phenomenon illustrated in bottom-left was discussed in~\cite{Rattey81}.}
\label{fig:schematic}
\end{figure}

We introduce here a method that computes the expansion coefficients $a_{k, q}$ in $O(L^3)$ operations instead of $O(L^4)$. The expansion coefficients in Eq.~\eqref{eq:contapprox} are given analytically by
\begin{align}
a_{k, q} &= \int_0^{2\pi} \int_0^c \mathcal{F}(f)( \xi , \theta ) \overline{\psi_c^{k, q}(\xi, \theta)} \xi\, d \xi\, d\theta \nonumber \\
& =  \int_0^c  N_{k, q} J_k\left(R_{k, q} \frac{\xi}{c}\right) \xi\, d\xi \int_0^{2\pi} \mathcal{F}(f)( \xi, \theta)e^{-\imath k \theta} d\theta.
\label{eq:FBint}
\end{align}
We evaluate the last integral numerically using a quadrature rule that consists of equally spaced points in the angular direction and a Gaussian quadrature rule in the radial direction, that is, using the nodes, $\xi_1(j, l) = \xi_j \cos(2\pi l / n_\theta)$, $\xi_2(j,l) = \xi_j \sin(2\pi l / n_\theta)$, $j = 1,\ldots,n_{\xi}$, $l = 0,\ldots, n_\theta-1$ (see Fig.~\ref{fig:schematic}). The values of $n_\xi$ and $n_\theta$ depend on the compact support radius $R$ and the band limit $c$ and are derived later in the paper. To use this quadrature rule, we need to sample the Fourier transform of $f$ at the quadrature nodes. This is approximated by the Fourier coefficients of the image $I$ (consisting of samples of $f$ on a Cartesian grid) at the given quadrature nodes, namely by the Fourier coefficients
\begin{equation}
\label{eq:nufft}
F(I)(\xi_1, \xi_2) = \frac{1}{(2R)^2}\sum_{i_1 = - R }^{R-1} \sum_{i_2 = -R}^{ R-1 }  I (i_1, i_2) e^{-\imath 2\pi (\xi_1 i_1 + \xi_2 i_2)},
\end{equation}
which can be evaluated efficiently using the the nonuniform discrete Fourier transform. The angular integration in Eq.~\eqref{eq:FBint} is then sped up by 1D FFT on the concentric circles, followed by a numerical evaluation of the radial integral with a Gaussian quadrature rule.
As the samples on each concentric circle are equally-spaced, the natural quadrature weights for the angular integral are $\frac{2\pi}{n_\theta}$, with the nodes taken at $\theta_l = \frac{2 \pi l}{n_\theta}$ for $l = 0, \dots, n_\theta-1$. The angular integration using one-dimensional FFT on each concentric circle thus yields
\begin{equation}
\widehat{F(I)}( \xi_j, k ) = \frac{ 2 \pi }{ n_\theta }\sum_{l = 0}^{n_\theta - 1} F(I)(\xi_j, \theta_l) e^{-\imath \frac{ 2\pi k  l }{n_\theta}}.
\label{eq:1DFFT}
\end{equation}
The radial integral is evaluated using the Gauss-Legendre quadrature rule~\cite[Chap. 4]{NRFT77}, which determines the locations of $n_\xi$ points $\{\xi_j\}_{j=1}^{n_\xi}$ on the interval $[0, c]$ and the associated weights $w(\xi_j)$. The integral in Eq.~\eqref{eq:FBint} is thus approximated by
\begin{equation}
\label{eq:a}
a_{k, q} \approx \sum_{j=1}^{n_\xi} N_{k, q}J_k\left(R_{k, q} \frac{\xi_j}{c}\right) \widehat{F(I)}( \xi_j, k ) \xi_j w( \xi_j).
\end{equation}
Since $I$ is real valued and $J_{-k}(x) = (-1)^k J_{k} (x)$, we get that $a_{-k, q} = a^*_{k, q}$ and thus we only need to evaluate coefficients with $k\geq 0$.

The procedure for numerical evaluation of the Fourier-Bessel expansion coefficients is illustrated in Fig.~\ref{fig:schematic}.
In practice, we have
observed that using $n_\xi = 4 c R$ and $n_\theta = 16 c R $ results in highly-accurate numerical evaluation of the integral in Eq.~\eqref{eq:FBint}.

\begin{figure}[h!]
\begin{center}
\subfloat[]{
\includegraphics[width=0.5\columnwidth]{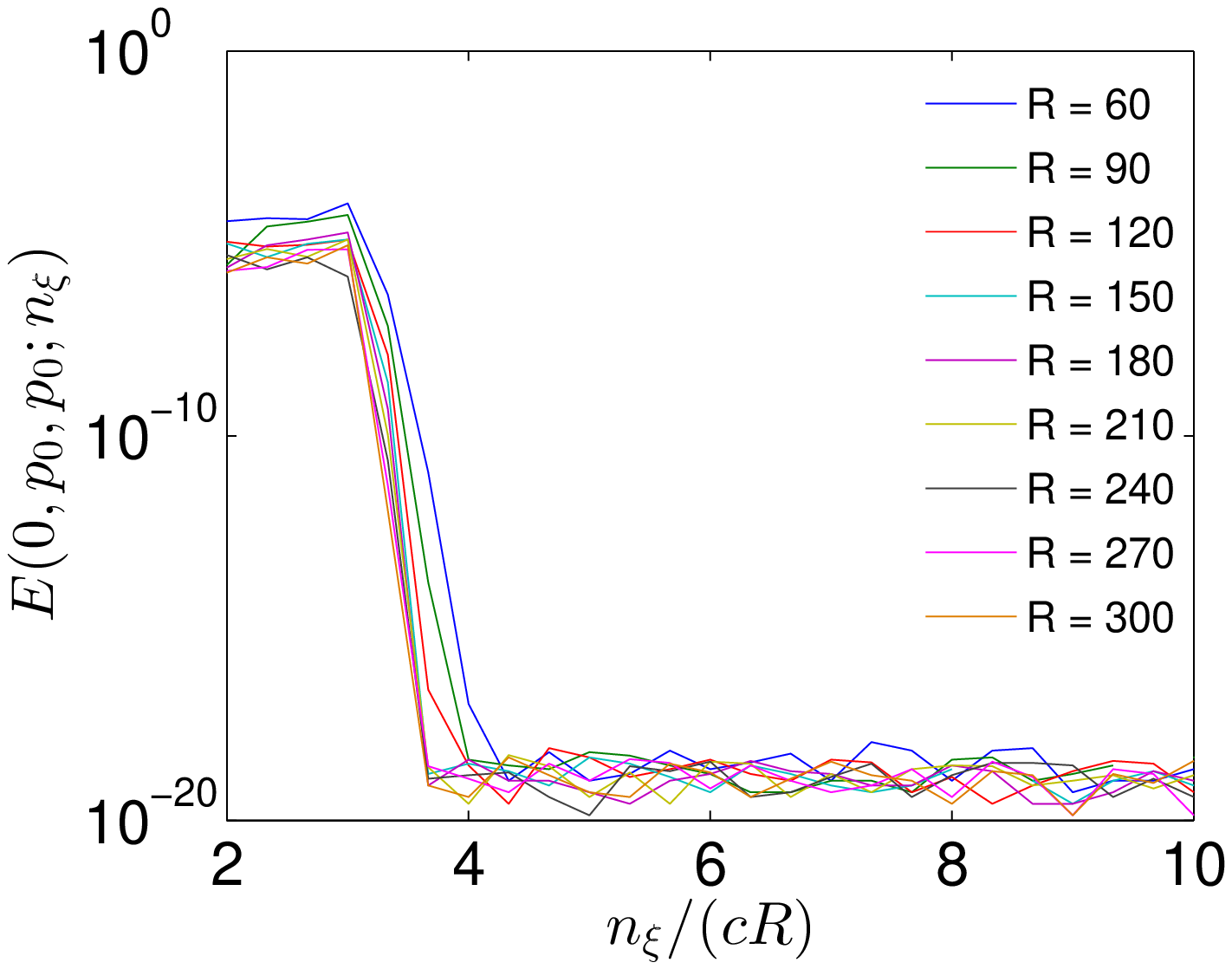}%
\label{fig:sample_r}
}
\subfloat[]{
\includegraphics[width=0.5\columnwidth]{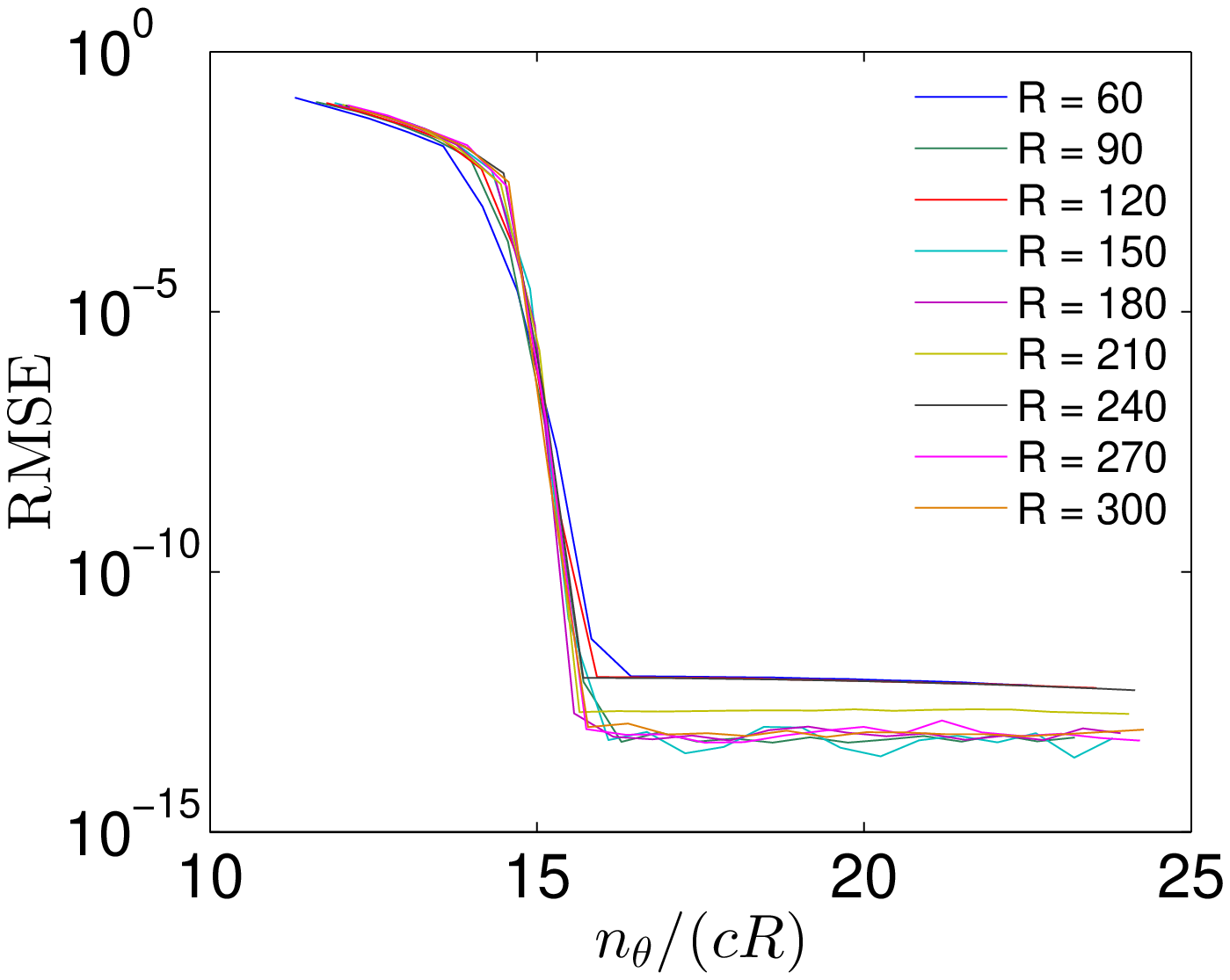}%
\label{fig:sample_theta}
}
\end{center}
\caption{\protect \subref{fig:sample_r} Error, as a function of $n_\xi$, in the numerical evaluation of the integral $G(0, p_0, p_0)$ in Eq.~\eqref{eq:gint}. \protect \subref{fig:sample_theta} Error, as a function of $n_\theta$, in the evaluation of the integral in Eq.~\eqref{eq:FBint}.}
\label{fig:sample_r_theta}
\end{figure}

If our image can be expressed in terms of the truncated Fourier-Bessel expansion in Eq.~\eqref{eq:contapprox}, the approximation error in the radial integral comes from the numerical evaluation of the integrals
\begin{equation}
\label{eq:gint}
G(k, q_1, q_2) = \int_0^c J_{k}\left(R_{k, q_1}\frac{\xi}{c}\right) J_{k}\left(R_{k, q_2}\frac{\xi}{c}\right) \xi d \xi,
\end{equation}
where the approximation error using $n_\xi$ points is
\begin{align}
&E(k, q_1, q_2; n_\xi) \nonumber \\
&= \left|\sum_{j = 1}^{n_\xi} J_{k}\left(R_{k, q_1}\frac{\xi_j}{c}\right) J_{k}\left(R_{k, q_2}\frac{\xi_j}{c}\right) \xi_j w(\xi_j) - G(k, q_1, q_2) \right|.
\end{align}
Asymptotically, a Bessel function behaves like a decaying cosine function with frequency $\frac{R_{k, q}}{2 \pi}$ for $R_{k, q}r\gg | k^2 - \frac{1}{4}|$~\cite{Abramowitz},
\begin{equation}
\label{asym}
J_k ( R_{k, q} r) \sim \sqrt{\frac{2}{\pi R_{k, q} r}}\,\cos ( R_{k, q} r -\frac{k \pi}{2} - \frac{\pi}{4}).
\end{equation}
For a fixed $n_\xi$, the largest approximation error occurs when $k = 0$ and $q_1 = q_2 = p_0$, since $J_0\left(R_{0, p_0} \frac{\xi}{c}\right)$ is the most oscillatory function within the band limit.
The Nyquist rate of $\xi J_0^2\left(R_{0, p_0} \frac{\xi}{c}\right)$ is $2 \frac{2R_{0, p_0}}{2\pi} \approx 4cR$  and we need to sample at Nyquist rate, or higher. Therefore, we choose $n_\xi =\left\lceil 4cR \right\rceil$. Fig.~\ref{fig:sample_r} justifies this choice as the error decays dramatically to $10^{-17}$ before $n_\xi = \left\lceil 4cR \right\rceil$.

To choose $n_\theta$, we computed the root mean squared error (RMSE) in evaluating the expansion coefficients for simulated images composed of white Gaussian noise with various $R$ and $n_\theta$, while $c = 1/2$. We oversampled on the radial lines by $n_\xi = \left\lceil 10 c R\right\rceil$ and the ground truth for the angular integral in Eq.~\eqref{eq:FBint} was computed by Eq.~\eqref{eq:1DFFT} via oversampling in the angular direction by $n_\theta = 60 c R$. We observe that when $n_\theta \geq 16cR$, the estimation error for the Fourier-Bessel expansion coefficients becomes negligible (see Fig.~\ref{fig:sample_theta}). Notice that Eq.~\eqref{eq:kmaxbs} implies that $k_{\max} < 2 \pi cR$. The corresponding Nyquist rate is bounded by $4\pi cR$. We therefore sample at a slightly higher rate of $n_\theta = 16 cR$ to ensure numerical accuracy.

\begin{figure}
\begin{center}
\subfloat[$R = 60$, $c = 1/3$, $L = 121$]{
\includegraphics[width=0.5\columnwidth]{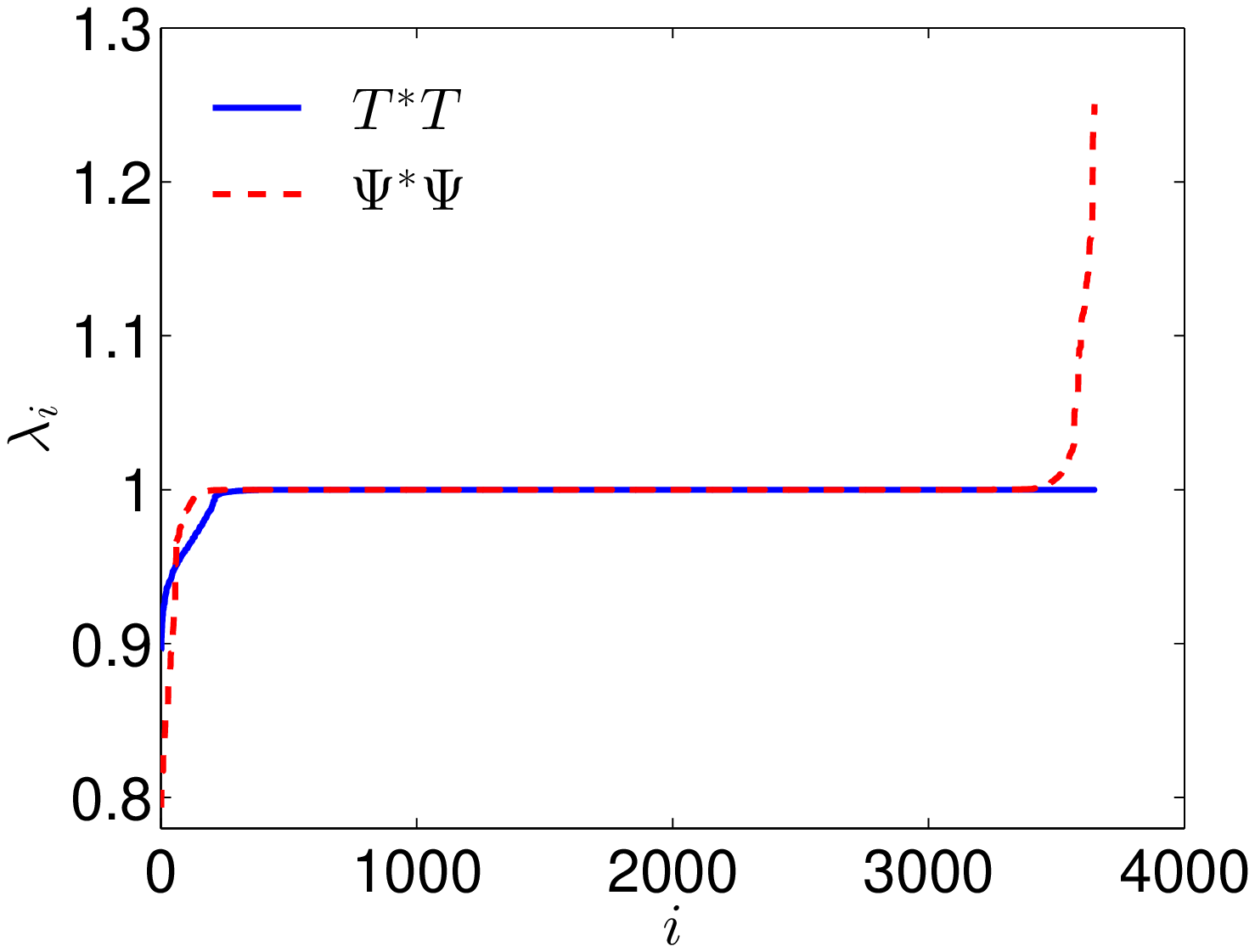}%
\label{subfig:L60}
}
\subfloat[$R = 90$, $c = 1/3$, $L = 181$]{
\includegraphics[width=0.5\columnwidth]{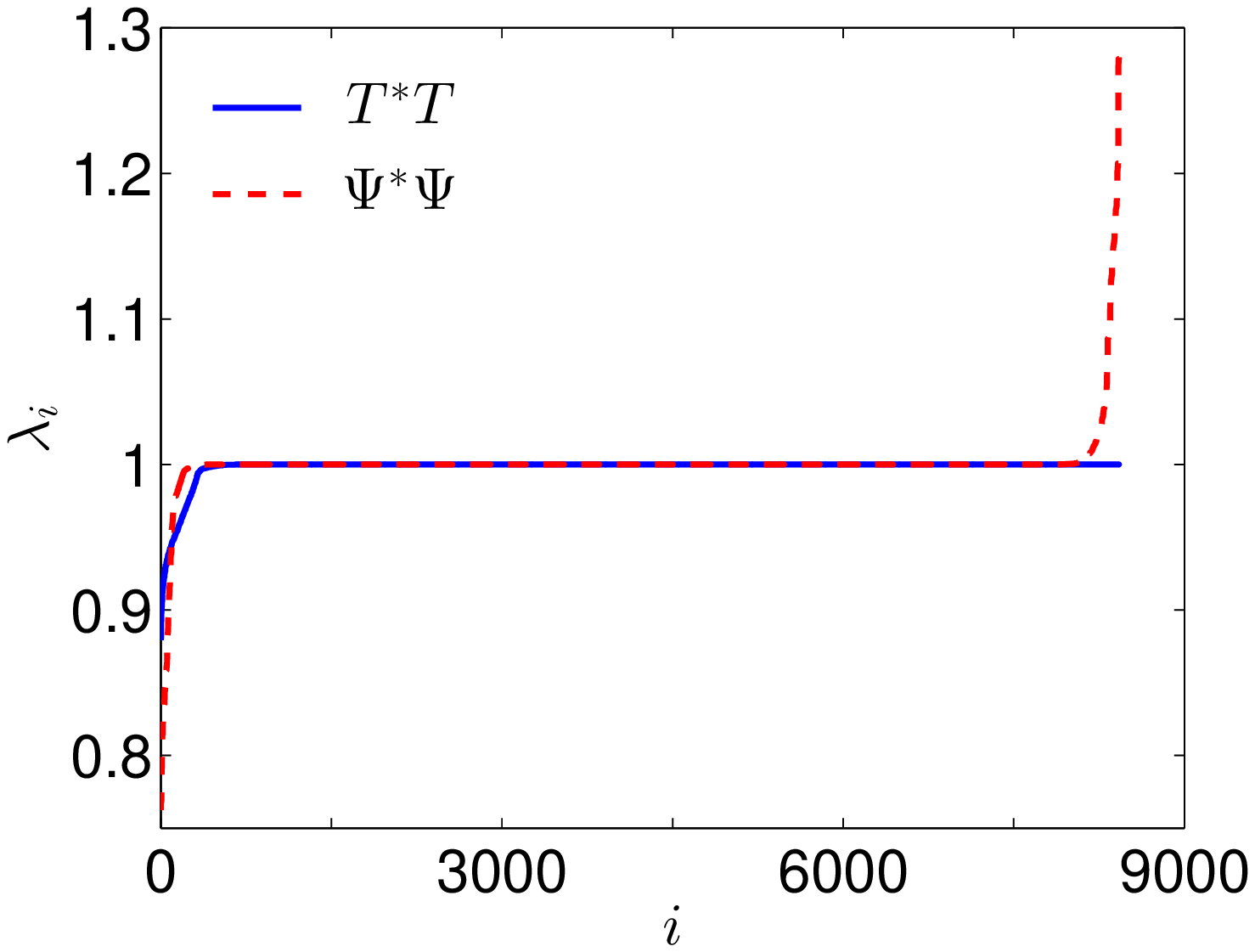}%
\label{subfig:L90}
}
\end{center}
\caption{Eigenvalues of $T^*T$ and $\Psi^* \Psi$, where $T^*$  and $\Psi^*$ are the truncated Fourier-Bessel transforms using numerical integration and least squares respectively. These are also the spectra of the population covariance matrices of transformed white noise images. Most eigenvalues are close to 1, indicating that the truncated Fourier-Bessel transform is almost unitary. Thus white noise remains approximately white.
}
\label{fig:TTT}
\end{figure}
Now that we are able to numerically evaluate $a_{k, q}$ with high accuracy, we can study the spectral behavior of the finite Fourier-Bessel expansion of the images. We define $a$ as the vector that contains the expansion coefficients $a_{k, q}$ computed in Eq.~\eqref{eq:a} and denote by $T^*$ the transformation that maps an image $I$ to its finite Fourier-Bessel expansion coefficients through Eqs.~\eqref{eq:nufft}, \eqref{eq:1DFFT} and~\eqref{eq:a}, that is,
\begin{equation}
\label{eq:transf}
a = T^* I.
\end{equation}
Ideally we would like $T^*$ to be a unitary transformation, that is $T^*T = I$, so that the transformation from the images to the coefficients preserves the noise statistics. Numerically, we observe that the majority of the eigenvalues of $T^*T$ are 1 and the smallest eigenvalues are also close to 1 (see blue solid line in Fig.~\ref{fig:TTT}). The transformation $T^*$ is close to unitary because it is a numerical approximation of an expansion in an orthogonal basis (Fourier-Bessel), and the sampling criterion prevents aliasing. In Fig.~\ref{fig:TTT}, the eigenvalues of $\Psi^* \Psi$ are also plotted for comparison. It can be observed that $T^*T$ has fewer eigenvalues that deviate from 1. 
Although the Fourier-Bessel functions are orthogonal as continuous functions, their discrete sampled versions are not necessarily orthogonal, hence $\Psi^* \Psi$ deviates from the identity matrix. The fact that $T^*T$ is closer to the identity than $\Psi^* \Psi$ implies that the numerical evaluation of the expansion coefficient vector $a$ as $T^* I$ is more accurate than estimating it as $\Psi^*I$.  We compare the numerical accuracy explicitly with an example. We choose a signal $f$ that satisfies $\mathcal{F}f(\xi, \theta) = \psi_c^{1, 5}(\xi, \theta)$ for $c = 0.5$ and $R = 30$, $a_{k, q} = 1$, for $k = 1$ and $q = 5$, and otherwise, $a_{k, q} = 0$. The evaluation method from~\cite{Zhao13} is applied here in Fourier space. It first evaluates discrete samples of $\mathcal{F}(f)$ and the Fourier-Bessel basis on a Cartesian grid of size $2R \times 2R$, and then projects the discrete samples onto the basis. The root mean squared error (RMSE) is $7.2\times 10^{-5}$ and the maximum absolute error is $4.0\times 10^{-4}$. Using the numerical evaluation in Eq.~\eqref{eq:a}, we get that $\text{RMSE}=1.2 \times 10^{-16}$ and the maximum absolute error is $2.7 \times 10^{-15}$.

Computing the polar Fourier transform of an image of size $L \times L$ 
 on a polar grid with $n_\xi \times n_\theta$ points in Eq.~\eqref{eq:nufft} is implemented efficiently using NUFFT~\cite{Dutt, nufft_fessler, Greengard, Fenn07}, whose computational complexity is $O(L^2 \log L + n_\xi n_\theta)$. Since $n_\theta = 16cR = O(L)$ and $n_\xi = 4cR = O(L)$, $n_\xi \times n_\theta = O(L^2)$ and the complexity of the discrete polar Fourier transform is $O(L^2 \log L)$.
The complexity of the 1D FFTs in Eq.~\eqref{eq:1DFFT} is $O( n_\xi n_\theta \log n_\theta)$, because there are $n_\xi$ concentric circles with $n_\theta$ samples on each circle. Both $n_\xi$ and $n_\theta$ are of $O(L)$, so the total complexity of the 1D FFTs is also $O(L^2 \log L)$. Evaluating  Eq.~\eqref{eq:a} (the quadrature rule for the radial integral in Eq.~\eqref{eq:FBint}) for all $k$ and $q$ requires a total of $O(L^3)$ operations using a direct method, because there are $O(L^2)$ basis functions to integrate, and each function is integrated using $O(L)$ quadrature points. However, this complexity can be reduced to $O(L^2 \log L)$ using a fast Bessel transform~\cite{ONeil10, Townsend15}. In summary, the computational complexity of computing the Fourier-Bessel expansion coefficients of an image of size $L\times L$ is $O(L^3)$ operations, or $O(L^2 \log L)$ using a ``fast'' transform.

\section{Steerable Principal Components}
\label{sec:cov}
Given a dataset of $n$ images $\{I_i\}_{i = 1}^n$, we denote by $f_i$ the underlying 
bandlimited function that corresponds to the $i$'th image $I_i$.  Under the action of the group $O(2)$, the function $f_i$ is transformed to $f_i^{\alpha,\beta}$, where $\alpha \in [0,2\pi)$ is the counter-clockwise rotation angle and $\beta$ denotes reflection and takes values in $\{+,-\}$. More specifically, $f_i^{\alpha, +} (r, \phi) = f_i(r, \phi-\alpha)$ and $f_i^{\alpha, -}(r, \phi) = f_i(r, \pi - (\phi - \alpha))$. The images $I_i^{\alpha, +}$ and $I_i^{\alpha, -}$ are obtained by sampling $f_i^{\alpha, +}$ and $f_i^{\alpha, -}$ respectively.

The Fourier transform of $f_i$ commutes with the action of the group $O(2)$, namely,
$\mathcal{F}(f_i^{\alpha, +}) (\xi, \theta) = \left ( \mathcal{F}(f_i) \right )^{\alpha, +}(\xi, \theta) = \mathcal{F}(f_i)(\xi, \theta - \alpha)$, and
$\mathcal{F}(f_i^{\alpha, -}) (\xi, \theta) = \left (\mathcal{F}(f_i)\right )^{\alpha, -}(\xi, \theta) = \mathcal{F}(f_i)(\xi, \pi - (\theta - \alpha))$. The transformation of the images under rotation and reflection can be represented by the transformation of their Fourier-Bessel expansion coefficients in Eq.~\eqref{eq:cont_int}.
Under counter-clockwise rotation by an angle $\alpha$, $\mathcal{F}(f_i^{\alpha, +})$ is given by
\begin{align}
\label{eq:rot}
\mathcal{F}(f_i^{\alpha, +}) (\xi, \theta) &  = \sum_{k, q} a^i_{k, q} \psi_c^{k, q} (\xi, \theta - \alpha) \nonumber \\
& = \sum_{k, q} a^i_{k, q} e^{-\imath k \alpha} \psi_c^{k, q}(\xi, \theta).
\end{align}
Therefore a planar rotation introduces a phase shift in the expansion coefficients.
Under rotation and reflection,
\begin{align}
& \mathcal{F}(f_i^{\alpha, -}) (\xi, \theta) = \sum_{k, q} a^i_{k, q} \psi_c^{k, q} (\xi, \pi - (\theta - \alpha)) \nonumber \\
& = \sum_{k, q} a^i_{k, q} N_{k, q} J_k\left(R_{k, q} \frac{\xi}{c}\right) e^{\imath k (\pi - \theta + \alpha)} \nonumber \\
& = \sum_{k, q} a^i_{k, q} N_{k, q} (-1)^k J_k\left(R_{k, q} \frac{\xi}{c}\right) e^{\imath (-k) \theta} e^{\imath k \alpha} \nonumber \\
& = \sum_{k, q} a^i_{k, q} e^{\imath k \alpha} \psi_c^{-k, q} (\xi, \theta) = \sum_{k, q} a^{i}_{-k, q} e^{-\imath k \alpha}\psi_c^{k, q}(\xi, \theta), \label{eq:refl}
\end{align}
namely, the expansion coefficient $a^i_{k, q}$ changes to $a^i_{-k, q} e^{-\imath k \alpha}$.

If we augment the collection of bandlimited functions $\{f_i\}_{i = 1}^n$ by all possible rotations and reflections, the Fourier transform of the sample mean of the augmented collection, denoted $f_{\mathrm{mean},}$ becomes,
\begin{equation}
\label{eq:mean1}
\mathcal{F}(f_{\mathrm{mean}})(\xi, \theta) = \frac{1}{2n}\sum_{i = 1}^n \sum_{\beta \in \{+, -\}} \frac{1}{2 \pi}\int_{0}^{2 \pi}\mathcal{F}( f_i^{\alpha, \beta})(\xi, \theta) d\alpha.
\end{equation}
Using the properties in Eqs.~\eqref{eq:rot} and \eqref{eq:refl}, we have
\begin{align}
\label{eq:mean2}
& \mathcal{F}(f_{\mathrm{mean}})(\xi, \theta) = \frac{1}{2n} \sum_{i = 1}^{n}\frac{1}{2 \pi} \nonumber \\ & \quad \times \int_{0}^{2 \pi} \sum_{k = -\infty}^{\infty} \sum_{q=1}^{\infty}\left[a^i_{k, q} + a^i_{-k, q} \right] e^{-\imath k \alpha} \psi_c^{k, q} (\xi, \theta) d \alpha \nonumber \\
& = \sum_{q=1}^{ \infty } \left(\frac{1}{n}\sum_{i=1}^n a_{0,q}^i\right) \psi_c^{0,q}(\xi,\theta).
\end{align}
As expected, the sample mean is radially symmetric, because $\psi_c^{0,q}$ is only a function of $\xi$ but not of $\theta$.

The rotationally invariant covariance kernel $\mathcal{C}((\xi, \theta), (\xi', \theta'))$ built from Fourier transformed functions with all their possible in-plane rotations and reflections is defined as
\begin{multline}
\mathcal{C}((\xi, \theta), (\xi', \theta')) = \frac{1}{4 \pi n} \times \\ \sum_{i=1}^{n} \sum_{\beta\in\{+,-\}}
 \int_0^{2\pi} \left(\mathcal{F}(f_i^{\alpha, \beta})(\xi, \theta) - \mathcal{F}(f_{\mathrm{mean}})(\xi, \theta)\right)\\
 \times \overline{\left(\mathcal{F}(f_i^{\alpha, \beta})(\xi', \theta') - \mathcal{F}(f_{\mathrm{mean}})(\xi', \theta')\right)} d\alpha. 
\end{multline}
From Eq.~\eqref{eq:mean2} it follows that if we express $\mathcal{F}(f_i)$ and $\mathcal{F}(f_\mathrm{mean})$ in terms of the Fourier-Bessel basis and the associated expansion coefficients, subtracting the sample mean is equivalent to subtracting $\frac{1}{n}\sum_{j=1}^n a_{0,q}^j $ from the coefficients $a^i_{0, q}$, while keeping other coefficients unchanged. Therefore, we first update the zero angular frequency coefficients by $a^i_{0, q} \gets a^i_{0, q} - \frac{1}{n}\sum_{j=1}^n a_{0,q}^j$, and then
\begin{align}
& \quad \mathcal{C}((\xi, \theta), (\xi', \theta')) = \frac{1}{4 \pi n}\sum_{i=1}^n \nonumber \\
& \int_0^{2\pi}\sum_{k = -\infty}^{\infty} \sum_{q = 1}^{\infty} \sum_{k' = -\infty}^{\infty} \sum_{q' = 1}^{\infty}\left(a^i_{k, q}\psi_c^{k, q}(\xi, \theta) \overline{a^i_{k', q'} \psi_c^{k', q'}(\xi', \theta')} \right. \nonumber \\
& \left.\quad +  a^i_{-k, q}\psi_c^{k, q}(\xi, \theta) \overline{a^i_{-k', q'}\psi_c^{k', q'}(\xi', \theta')} \right) e^{-\imath (k-k')\alpha} d \alpha  \nonumber \\
& = \sum_{k = -\infty}^{\infty} \sum_{q = 1}^{\infty}\sum_{k' = -\infty}^{\infty} \sum_{q' = 1}^{\infty}\psi_c^{k, q}(\xi, \theta) C_{(k, q), (k', q')} \overline{\psi_c^{k', q'}(\xi', \theta')},
\label{eq:covfcovFB}
\end{align}
where 
\begin{align}
& \quad C_{(k, q), (k', q')}\nonumber \\ &= \frac{1}{4\pi n} \sum_{i=1}^{n} \int_0^{2\pi} \left(a^i_{k, q } \overline{a^i_{k', q'}} + a^i_{-k, q} \overline{a^i_{- k', q'}}\right) e^{-\imath (k-k') \alpha}  d \alpha  \nonumber
\\ &=  \delta_{k,k'}\frac{1}{n} \sum_{i=1}^{n} \mathrm{Re} \left\{a^i_{k, q } \overline{a^i_{k', q'}} \right\}.
\label{eq:cov}
\end{align}
$\delta_{k, k'}$ comes from the integral over $\alpha \in [0, 2\pi)$. 
The covariance matrix in Eq.~\eqref{eq:cov} is positive semi-definite and block diagonal because the non-zero entries of $C$ correspond only to $k=k'$. 
Since the images are well approximated by the subspace spanned by a finite number of Fourier-Bessel basis functions (see Eq.~\eqref{eq:contapprox}), $C_{(k, q), (k', q')}$ are close to zero when $(k, q)$ or $(k', q')$ do not satisfy the sampling criterion in Eq.~\eqref{criterion1}. Therefore, we have a finite matrix representation $C$ of $\mathcal{C}$. Moreover, it suffices to consider $k\geq 0$, because $C_{(k,q),(k,q')} = C_{(-k,q),(-k,q')}$.  
Thus, the covariance matrix in Eq.~\eqref{eq:cov} can be written as the direct sum $C = \bigoplus_{k=0}^{k_{\max}} C^{(k)}$, where $C^{(k)}$ is by itself a sample covariance matrix of size $p_k \times p_k$, given by,
\begin{equation}
C^{(k)}_{q, q'} = \frac{1}{n} \sum_{i=1}^{n} \mathrm{Re} \left\{a^i_{k, q } \overline{a^i_{k, q'}} \right\}.
\end{equation}

Let us denote by $A^{(k)}$ the matrix of expansion coefficients, obtained by putting the coefficients
$a^i_{k, q}$ 
for all $q$ and all $i$ into a matrix, where the columns are indexed by the image number $i$ and the rows are indexed by the radial index $q$. The coefficient matrix $A^{(k)}$ for $k\neq 0$ is of size $p_k \times n$ and the covariance matrix for $k\neq 0$ is,
\begin{equation}
\label{eq:ck}
C^{(k)} = \frac{1}{n} \mathrm{Re} \left\{A^{(k)} (A^{(k)})^* \right\},
\end{equation}
where $A^*$ is the conjugate transpose ($A_{ij}^* = \bar{A}_{ji}$).
The case $k=0$ is special because the expansion coefficients satisfy $a_{0,q} = \overline{a_{0,q}}$, and so $A^{(0)}$ is a matrix of size $p_0 \times n$ and
\begin{equation}
\label{eq:ck_0}
C^{(0)} = \frac{1}{n}A^{(0)}(A^{(0)})^*.
\end{equation}

We compute the eigenvalues $\lambda^{(k)}_1 \ge \lambda^{(k)}_2 \cdots \geq \lambda_{p_k}^{(k)}$ and eigenvectors $u^{(k)}_1, u^{(k)}_2, \ldots, u^{(k)}_{p_k}$ of the covariance matrices $C^{(k)}$. Because $\mathcal{C}$ and $C$ are related through Eq.~\eqref{eq:covfcovFB} and $C$ is block diagonal as in Eq.~\eqref{eq:cov}, 
$\mathcal{C}((\xi, \theta), (\xi', \theta'))$ is well approximated by $\sum_{k=-k_{\max}}^{k_{\max}} \Psi^{(k)}(\xi, \theta) C^{(k)} (\Psi^{(k)})^*(\xi', \theta')$, where $\Psi^{(k)}$ contains Fourier-Bessel functions with angular frequency $k$. Equation~\eqref{eq:covfcovFB} reveals that the eigenfunctions of $\mathcal{C}$, which are the steerable principal components, can be expressed as linear combinations of the Fourier-Bessel functions with the coefficients given by the eigenvectors of the matrix $C$,
\begin{align}
\label{eq:sPCArt}
g^{k, l}(\xi, \theta) &= \sum_{q=1}^{p_k} \psi_c^{k, q}(\xi, \theta)u^{(k)}_l(q) \nonumber \\
& = \sum_{q=1}^{p_k} N_{k, q}J_k\left(R_{k, q} \frac{\xi}{c}\right)u^{(k)}_l(q)e^{\imath k \theta}.
\end{align}
Therefore the radial parts of the steerable principal components  
\begin{equation}
\label{eq:sPCArad}
f^{k, l}(\xi) = \sum_{q=1}^{p_k} N_{k, q}J_k\left(R_{k, q} \frac{\xi}{c}\right)u^{(k)}_l(q)
\end{equation} are linear combinations of the Bessel functions within the same angular frequency. The associated expansion coefficients for $I_i$ are
\begin{equation}
c^i_{k, l} = \sum_{q=1}^{p_k} a^i_{k, q} u_l^{(k)}(q),\quad \text{for } i = 1, \dots, n.
\label{eq:sPCAcoeff}
\end{equation}

The computational complexity for forming the matrix $C^{(k)}$ is $O(np_k^2)$.
The complexity for eigendecomposition of $C^{(k)}$ is $O(p_k^3)$, since the size of the covariance matrix is $p_k \times p_k$.
Using the upper and lower bounds for $p_k$ in Eq.~\eqref{eq:pkbs} and assuming $c = O(1)$ and $R = O(L)$, we get $\sum_k p_k^2 = O(L^3)$ and $\sum_k p_k^3 = O(L^4)$. Therefore, the complexity for forming the covariance matrix $C$ is $ O(n\sum_k p_k^2) = O(nL^3)$ and the complexity of its full eigendecomposition is $O(\sum_k p_k^3) = O(L^4)$.
Equations~\eqref{eq:ck} and~\eqref{eq:ck_0} show that instead of constructing the covariance matrices $C^{(k)}$ to compute the principal components, we can perform singular value decomposition (SVD) on the coefficient matrix $A^{(k)}$ directly and take the left singular vectors as the principal components. The computational complexity for full rank SVD on $A^{(k)}$ is $O(np_k^2)$ and 
the total complexity of SVD of all coefficient matrices is $O(n\sum_k p_k^2) = O(nL^3)$.

\section{Algorithm and computational complexity}
\label{sec:alg}
The new algorithm introduced in this paper is termed fast Fourier-Bessel steerable PCA (FFBsPCA). The algorithm is composed of two steps. In the first step, Fourier-Bessel expansion coefficients are computed according to Algorithm~\ref{alg:FBCoeff}. The input to the algorithm includes an image dataset, the band limit $c$, and the compact support radius $R$. The second step (Algorithm~\ref{alg:sPCAI}) takes the Fourier-Bessel expansion coefficients from Algorithm~\ref{alg:FBCoeff} as input and computes the steerable PCA radial functions and the expansion coefficients of the images in the new steerable basis. Algorithm~\ref{alg:sPCAI} is the same as the corresponding part of the algorithm in~\cite{Zhao13}.
\begin{algorithm}
\caption {Fast Fourier-Bessel Expansion}
\begin{algorithmic}[1]
\REQUIRE $n$ images $I_1,\ldots,I_n$ sampled on a Cartesian grid of size $L\times L$ with compact support radius $R$ and band limit $c$. \\
\STATE (Precomputation) Select $(k, q)$'s that satisfy the sampling criterion of Eq.~(\ref{criterion1}). Fix $n_\xi = \left\lceil 4cR \right\rceil$ and $n_\theta = \left\lceil 16 cR \right\rceil$.
\STATE (Precomputation) Find $n_\xi$ Gaussian quadrature points and weights on the interval $[0, c]$ and evaluate $N_{k, q}J_{k}(R_{k, q} \frac{\xi_j}{c})$, $j = 1, \dots, n_\xi$, for all selected $(k, q)$'s.\\
\STATE Compute $F(I_i)$ (Eq.~\eqref{eq:nufft}) on a polar grid of size  $n_\xi \times n_\theta$ by NUFFT for each $i = 1, \ldots, n$.\\
\STATE For each $F(I_i)$, compute $a^i_{k, q}$ using Eqs.~(\ref{eq:1DFFT}) and~(\ref{eq:a}).\\
\RETURN $a^i_{k, q}$ for all selected $(k, q)$'s.
\end{algorithmic}
\label{alg:FBCoeff}
\end{algorithm}

\begin{algorithm}
\caption {Steerable PCA}
\begin{algorithmic}[1]
\REQUIRE Fourier-Bessel expansion coefficients $a^i_{k, q}$ for $n$ images and the maximum angular frequency $k_{\max}$.
\STATE Compute the coefficient vector of the mean image $a^{\text{mean}}_{0, q} = \frac{1}{n} \sum_j a^j_{0, q}$. Then, set $a^i_{0, q} \leftarrow a^i_{0, q}-a^{\text{mean}}_{0, q}$.
\FOR {$k = 0, 1, \dots, k_{\max}$}
\STATE Construct the coefficient matrix $A^{(k)}$. \\
\STATE Compute the covariance matrix $C^{(k)}$, its eigenvalues $\lambda^{(k)}_1 \ge \lambda^{(k)}_2 \cdots \geq \lambda_{p_k}^{(k)}$, and eigenvectors,  $u^{(k)}_1,\ldots, u^{(k)}_{p_k}$; or perform SVD of $A^{(k)}$ and take the left singular vectors $u^{(k)}_1, \ldots, u^{(k)}_{p_k}$.\\
\STATE Compute the radial eigenvectors $f^{k,l}(\xi_j)$ for $j = 1, \dots, n_\xi$ using Eq.~\eqref{eq:sPCArad}. \label{line5} \\
\STATE Compute the expansion coefficients of the images in the new steerable basis $c^i_{k,l}$ using Eq.~\eqref{eq:sPCAcoeff}. \label{line6}\\
\ENDFOR
\RETURN For all $(k, l)$, $u_l^{(k)}$, $\lambda_l^{(k)}$, $f^{k, l}$, and $c^i_{k, l}$ $i=1,\ldots,n$.
\end{algorithmic}
\label{alg:sPCAI}
\end{algorithm}

The analysis of the computational complexity of FFBsPCA is as follows.
The precomputation that generates all radial basis functions requires $O(L^3)$ operations because there are $O(L^2)$ basis functions, each of which is sampled over $O(L)$ points.
Computing the Fourier-Bessel expansion coefficients $a^i_{k, q}$ in Eq.~\eqref{eq:a} for all images takes $O(nL^3)$ operations (or $O(nL^2\log L)$ with a fast Bessel transform) as discussed in Section~\ref{subsec:pft}.

The complexity of constructing the covariance matrix $C$ and computing its full eigendecomposition is $O(nL^3 + L^4)$ as described in Section~\ref{sec:cov}.
Another method for computing the principal components is by SVD of the coefficient matrices. Full rank SVD on all coefficient matrices requires $O(nL^3)$ floating point operations (see Section~\ref{sec:cov}).

To generate the new steerable basis, we take linear combinations of the Bessel functions as in line~\ref{line5} of Algorithm~\ref{alg:sPCAI}, which takes $O(L^4)$ operations. Computing the steerable PCA expansion coefficients $c^i_{k, l}$ for $i = 1\dots, n$ (line~\ref{line6} in Algorithm~\ref{alg:sPCAI}) requires $O(n L^3)$ operations by taking linear combinations of the Fourier-Bessel expansion coefficients as in Eq.~\eqref{eq:sPCAcoeff}. Therefore the total computational complexity of FFBsPCA is $O(nL^3 + L^4)$.

The complexity of FBsPCA introduced in~\cite{Zhao13} is $O(nL^4)$. Thus, FFBsPCA is faster than FBsPCA. For PCA, when the number of images is smaller than the number of pixels in the compact support disk, we form $X^TX$ and compute its eigendecomposition and the complexity is $O(n^2L^2 + n^3)$.  However, as the number of images grows, the complexity of PCA switches to $O(nL^4 + L^6)$ since it becomes more efficient to compute the eigendecomposition of $XX^T$. Therefore the computational complexity of traditional PCA, without taking into account all rotations and reflections is $O(\min\{n^2L^2 + n^3, nL^4 + L^6\})$.  When $n> O(L)$, FFBsPCA is more efficient than the traditional PCA.

FFBsPCA is easily adapted for parallel computation. The computation of Fourier-Bessel expansion coefficients in Algorithm~\ref{alg:FBCoeff} can run on multiple workers in parallel, where each worker is allocated with a subset of the images and Fourier-Bessel radial basis functions. 
In addition, in Algorithm~\ref{alg:sPCAI}, the radial eigenfunctions and the steerable PCA expansion coefficients can also be efficiently computed in parallel for each angular index $k$.

\section{Numerical Experiments}
\label{sec:results}
We compare the running times of FFBsPCA, FBsPCA and traditional PCA, where the latter is computed without the images' in-plane rotations and reflections. The algorithms are implemented in MATLAB on a machine with 60 cores, running at 2.3 GHz, with total RAM  of 1.5TB.

We first simulated $n = 24,000$ images with different radii of compact support $R$, while the band limit is fixed at $c = 1/2$. For small $R$, since FFBsPCA performs polar Fourier transformation, it appears slightly slower than FBsPCA. However when $R$ increases, FFBsPCA is computationally more efficient (see Tab.~\ref{tab:T_varL}).
\begin{table}
\begin{center}
\begin{tabular}{| c | c | c | c |}
\hline
$R$  & PCA & FBsPCA & FFBsPCA \\
\hline
30  &  8     & {\bf 7}      & 51 \\
\hline
60  &  214     & \textbf{50}   & 87 \\
\hline
90  &  1,636    &  168         & \textbf{148}\\
\hline
120 &  1,640    &  413         & \textbf{234}\\
\hline
150 &  1,808     &  757        & \textbf{371}\\
\hline
180 &  1,988     &  1,437      & \textbf{657}\\
\hline
210 &  2,106     &  2,274      & \textbf{695}\\
\hline
240 &  2,188      & 3,827      &\textbf{892}\\
\hline
\end{tabular}
\end{center}
\caption{Running times (in seconds) as a function of $R$ for $n = 2.4\times 10^4$, $c = 1/2$, and $L = 2R$.}
\label{tab:T_varL}
\end{table}
We next fixed the size of the images while using $R = 150$ and $c = 1/2$, and varied the number of images $n$.  Table~\ref{tab:T_varN} shows that the running time of FBsPCA and FFBsPCA grows linearly with $n$. 
\begin{table}
\begin{center}
\begin{tabular}{| c | c | c | c |}
\hline
$n$ ($\times 10^3$)  & PCA & FBsPCA & FFBsPCA \\
\hline
 1  &  $ {\bf 0.05} $   & 1.2  &  1.1 \\
\hline
 2  &    $ {\bf 0.1} $  & 2.1  &  1.3 \\
\hline
 4  &   $ {\bf 0.3}$ & 3.5  &  1.8 \\
\hline
 8  &   $ {\bf 1.3} $ & 4.3  &  2.4 \\
\hline
 16 &   $ 9.8 $ &  8.7   &  \textbf{4.6} \\
\hline
 32 &   $ 59.1 $ &  17.9   &  \textbf{8.0} \\
\hline
 64 &   $ 424.7 $ &  35.7   &  \textbf{14.4} \\
\hline
 128 &   $ 653.7 $ &  74.2   &  \textbf{30.6} \\
\hline
\end{tabular}
\end{center}
\caption{Running times (in minutes) as a function of $n$ for image size $300 \times 300$ pixels ($L=300$), with $R = 150$ and $c = 1 / 2 $.}
\label{tab:T_varN}
\end{table}

To show that our new algorithm can handle large datasets efficiently, we simulated a large dataset with $10^5$ images of size $300\times 300$ pixels. The images consist entirely of Gaussian noise with mean 0 and variance 1. We assume that the compact support in the image domain is $R = 150$ and the band limit in Fourier domain is $c = 1/2$. In Table~\ref{tab:T_big}, the total running time is divided into three parts: precomputation, Fourier-Bessel expansion (Algorithm~\ref{alg:FBCoeff}), and steerable PCA (Algorithm~\ref{alg:sPCAI}).
\begin{table}
\begin{center}
\begin{tabular}{| c | c |}
\hline
Steps  &  Time (sec)\\
\hline
Precomputation  &  7 \\
\hline
NUFFT and Fourier-Bessel Expansion  & 1,438  \\
\hline
Steerable PCA    & 42  \\
\hline \hline
\textbf{Total} & \textbf{1487} \\
\hline
\end{tabular}
\end{center}
\caption{Timing for FFBsPCA on a large dataset with $n = 10^5$ images. Each image is of size $300 \times 300$ pixels, $R = 150$ and $c = 1/2$. We computed the full eigendecomposition in Algorithm~\ref{alg:sPCAI}.}
\label{tab:T_big}
\end{table}
Fourier Bessel expansion took about 24 minutes, during which $91\%$ of the time was spent on mapping images to polar Fourier grid, where we used the software package~\cite{Fenn07} downloaded from \href{https://www-user.tu-chemnitz.de/~potts/nfft/}{https://www-user.tu-chemnitz.de/$\sim$potts/nfft/}. Numerical evaluation of the angular integration by 1D FFT  and the radial integration by a direct method took $6.4\%$ and $2.6\%$ of the time respectively. Steerable PCA took $42$ seconds.

\begin{figure}[h!]
\begin{center}
\subfloat[Clean]{
\includegraphics[width=0.4\columnwidth]{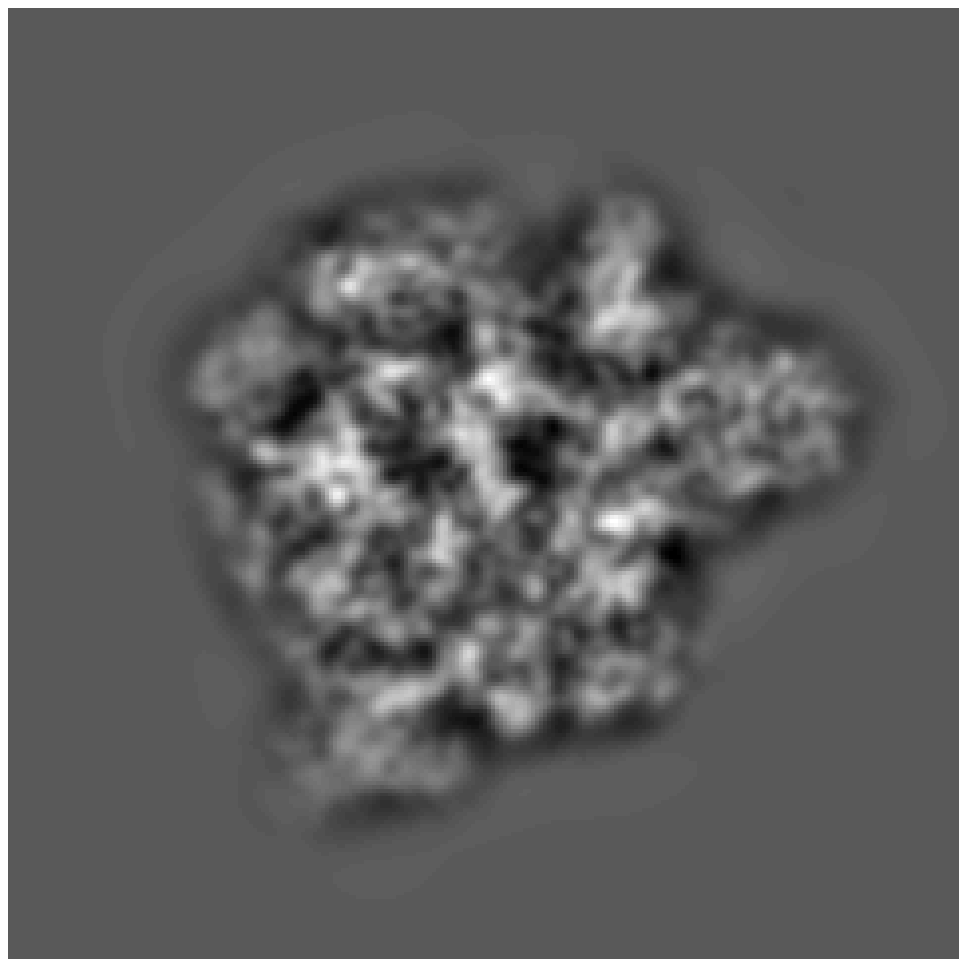}%
\label{fig:clean}
}
\subfloat[SNR$= 1/30$]{
\includegraphics[width=0.4\columnwidth]{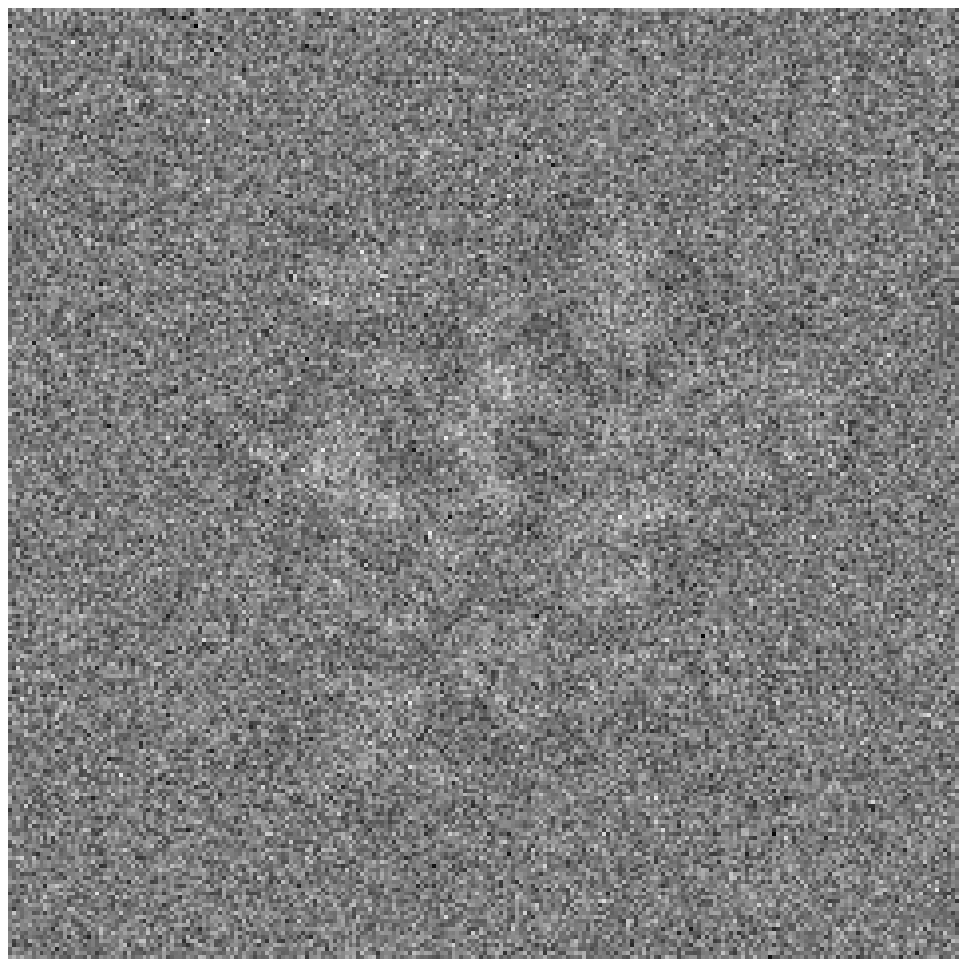}%
\label{fig:noisy}
}
\end{center}
\caption{Simulated projection images of the human mitochondrial large ribosomal subunit. Image size is $240 \times 240$ pixels.}
\label{fig:Simulation}
\end{figure}

In our third experiment, we simulated $n=10^5$ clean projection images from a reconstructed volume of a human mitochondrial large ribosomal subunit, downloaded from the electron microscopy data bank~\cite{AlanBrown14} (EMDB-2762). The original volume in the data bank is of size $320 \times 320 \times 320$ voxels. We preprocessed the volume such that its center of mass is at the origin and cropped out a volume of size $240 \times 240 \times 240$ voxels that contains the particle. Each projection image is of size $240 \times 240$ pixels. We simulated both the vanishing behavior of the CTF at low frequencies and the blurring effect due to the Gaussian envelope of the CTF. This was done by convolving the images with the inverse Fourier transform of
\begin{equation}
\min(\pi \lambda z f^2 + a,\, 1) \exp(-B f^2),
\label{eq:CTFenv} 
\end{equation}where $f$ is the frequency, $\lambda$ is the wavelength of the electron beam, $z$ is the defocus, and $a$ is the phase of the CTF introduced by microscope. This stems from the analytic form of the CTF given by $\sin(\pi\lambda z f^2 + a)\exp(-B f^2)$. 
For the simulations we chose $\lambda = 0.0197\text{\AA}$, $z = 2.5\mu m$, $a = 0.1 \mathrm{rad}$, and $B = 100 \text{\AA}^2$. Our clean images (see Fig.~\ref{fig:clean}) are the projection images filtered by the filter in Eq.~\eqref{eq:CTFenv} and they were then corrupted by additive white Gaussian noise at SNR$=1/30$, corresponding to noise variance of $\sigma^2 = 9$ (see Fig. \ref{fig:noisy}).

\begin{figure}[h!]
\begin{center}
\subfloat[Estimating $R$]{
\includegraphics[width=0.4\columnwidth]{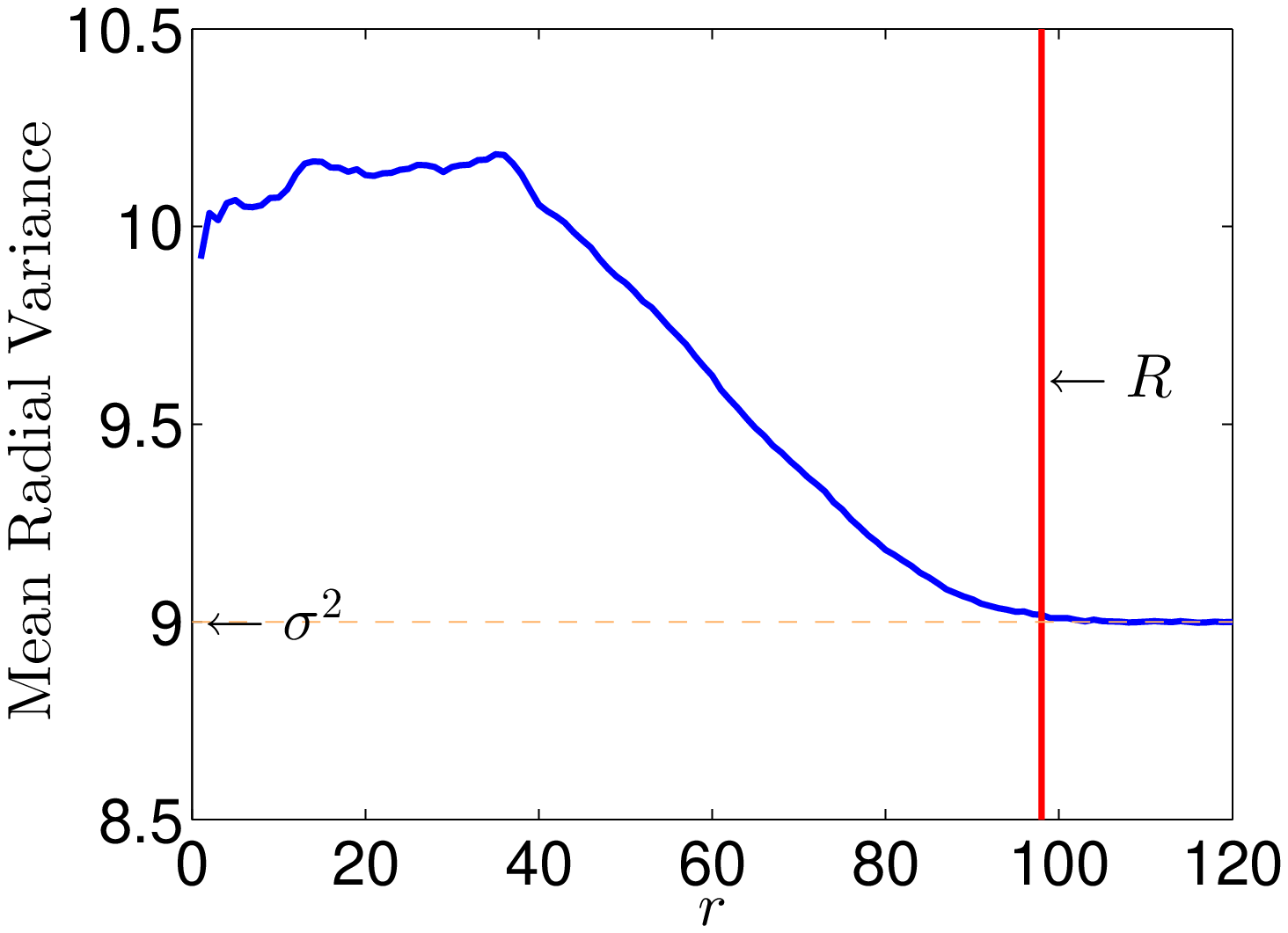}%
\label{fig:R}
}
\subfloat[Estimating $c$]{
\includegraphics[width=0.4\columnwidth]{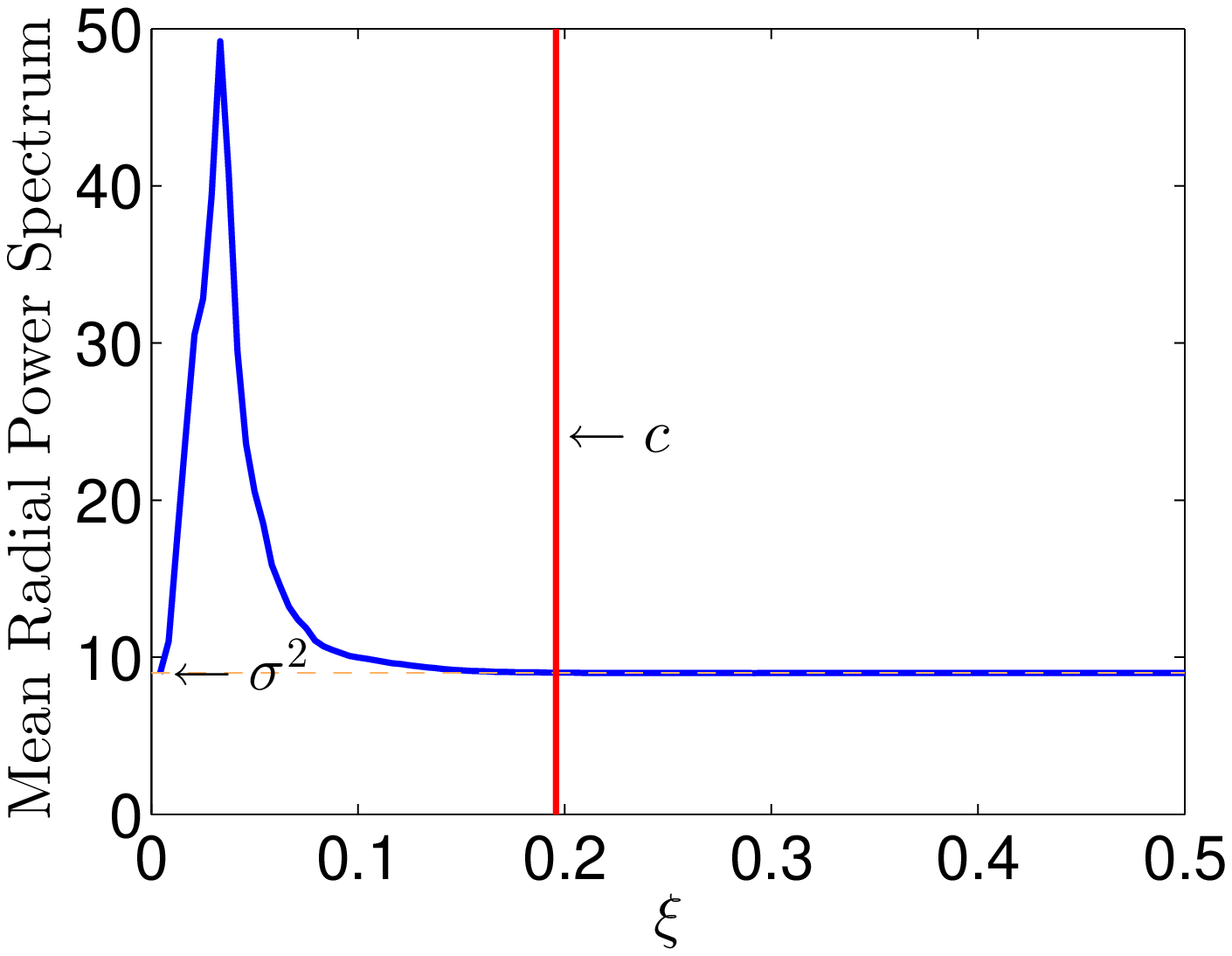}%
\label{fig:c}
}
\end{center}
\caption{Estimating $R$ and $c$ from $n = 10^5$ simulated noisy projection images of a human mitochondrial large ribosomal subunit. Each image is of size $240 \times 240$ pixels. \protect\subref{fig:R}~Mean radial variance of the images. The curve levels off at about $\sigma^2 = 9$ when $r \geq 98$. The radius of compact support is chosen as $R = 98$. \protect\subref{fig:c}~Mean radial power spectrum. The curve levels off at $\sigma^2 = 9$ when $\xi \geq 0.196$. The band limit is chosen as $c = 0.196$.}
\label{fig:cR}
\end{figure}
We estimated the radius of compact support of the particle in real domain and the band limit in Fourier domain from the noisy images in the following way. We first subtracted the mean image of the dataset from each image. Then we computed the 2D variance map of the dataset averaged in the angular direction, to get the mean radial variance (see Fig.~\ref{fig:R}). At large $r$, the mean radial variance levels off at 9, which corresponds to the noise variance. We subtracted the noise variance from the estimated mean radial variance and computed the cumulative variance by integrating the mean radial variance over $r$ with a Jacobian weight $r d r$. The fraction of the cumulative variance reaches $99.9\%$ at $r = 98$, and therefore $R$ was chosen to be $98$. In the Fourier domain, we computed the angular average of the mean 2D power spectrum. The curve in Fig.~\ref{fig:c} also levels off at the noise variance when $\xi$ is large. We used the same method as before to compute the cumulative radial power spectrum. The fraction reaches $99.9\%$ at $\xi = 0.196$, therefore the band limit is chosen to be $c = 0.196$.

\begin{figure}
\begin{center}
\begin{tabular}{ccc}
\includegraphics[width=0.3\columnwidth]{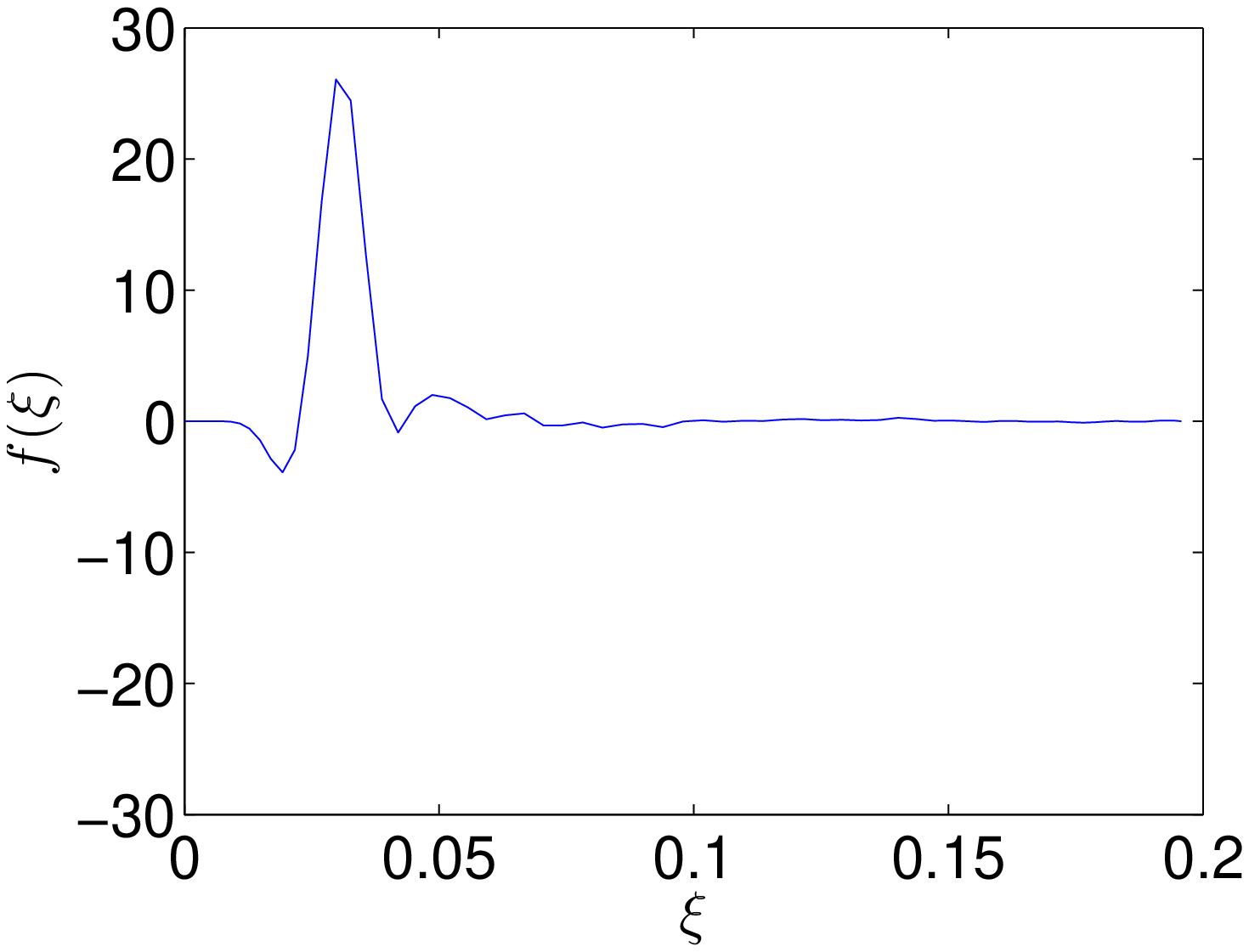} &
\includegraphics[width=0.3\columnwidth]{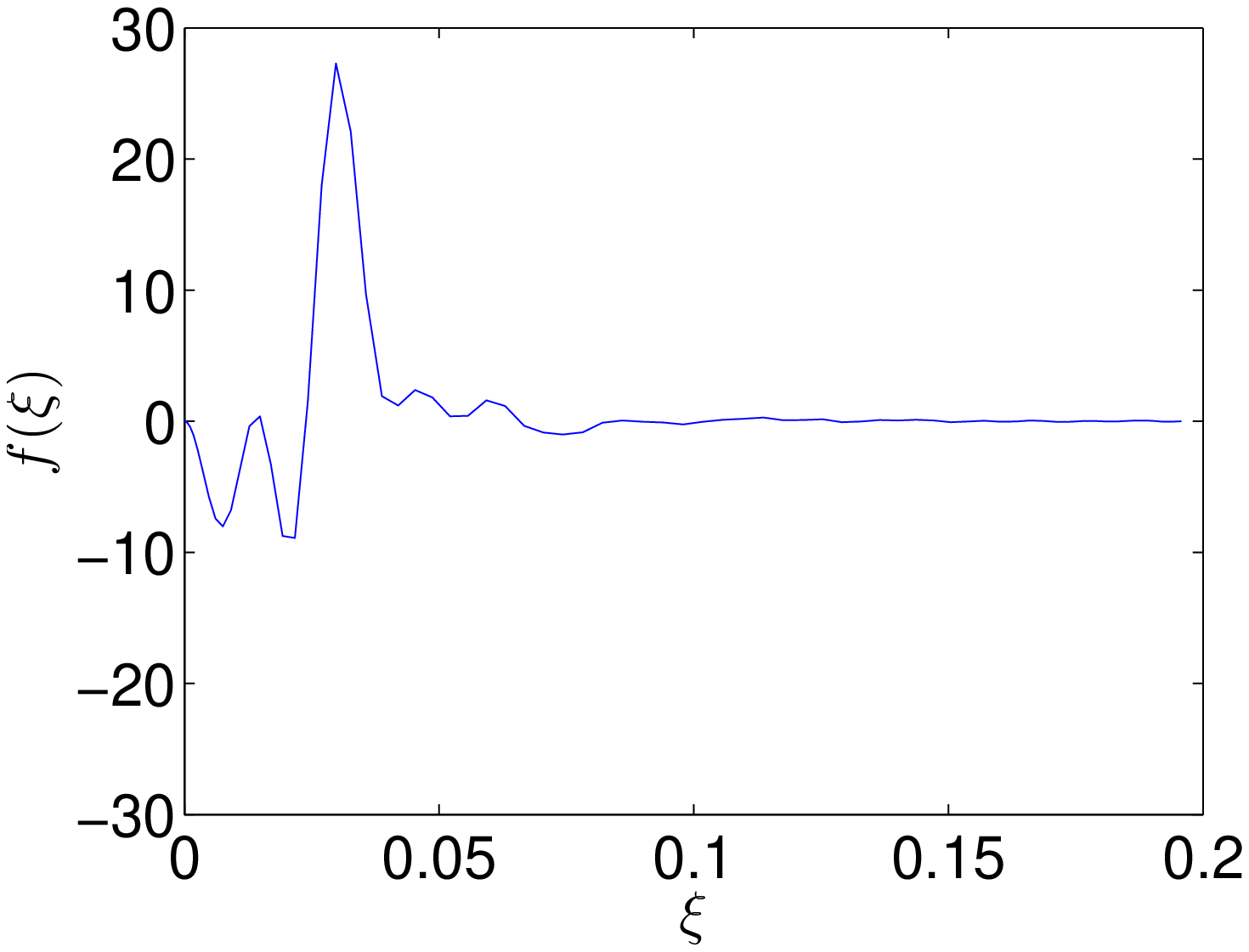} &
\includegraphics[width=0.3\columnwidth]{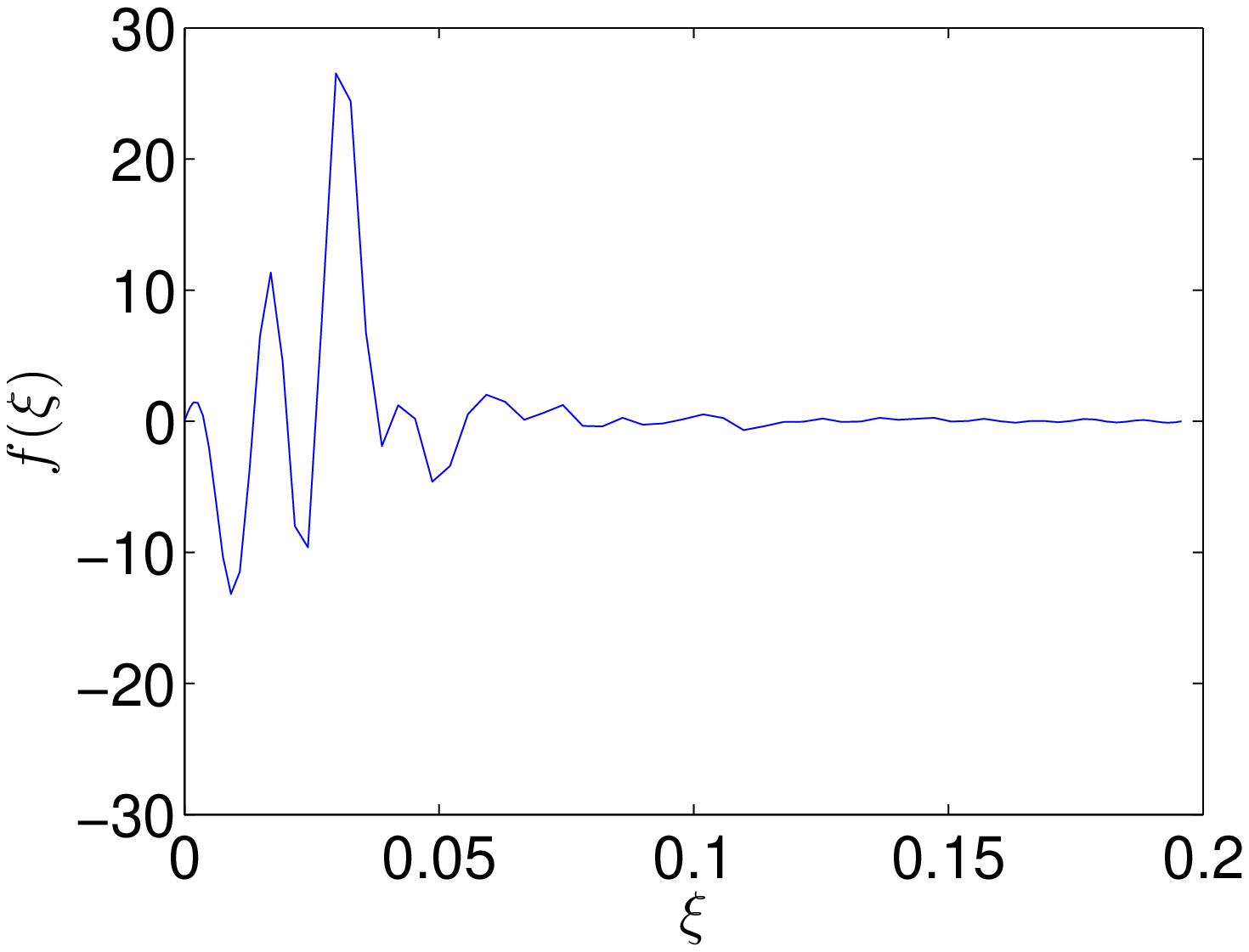} \\
$ k = 8 $, $l = 1$ & $ k = 2 $, $l = 1$ & $ k = 1 $, $l = 1$ \\
$\lambda = 163.8$ &  $\lambda = 160.0$  &$\lambda = 158.1$ \\
\includegraphics[width=0.3\columnwidth]{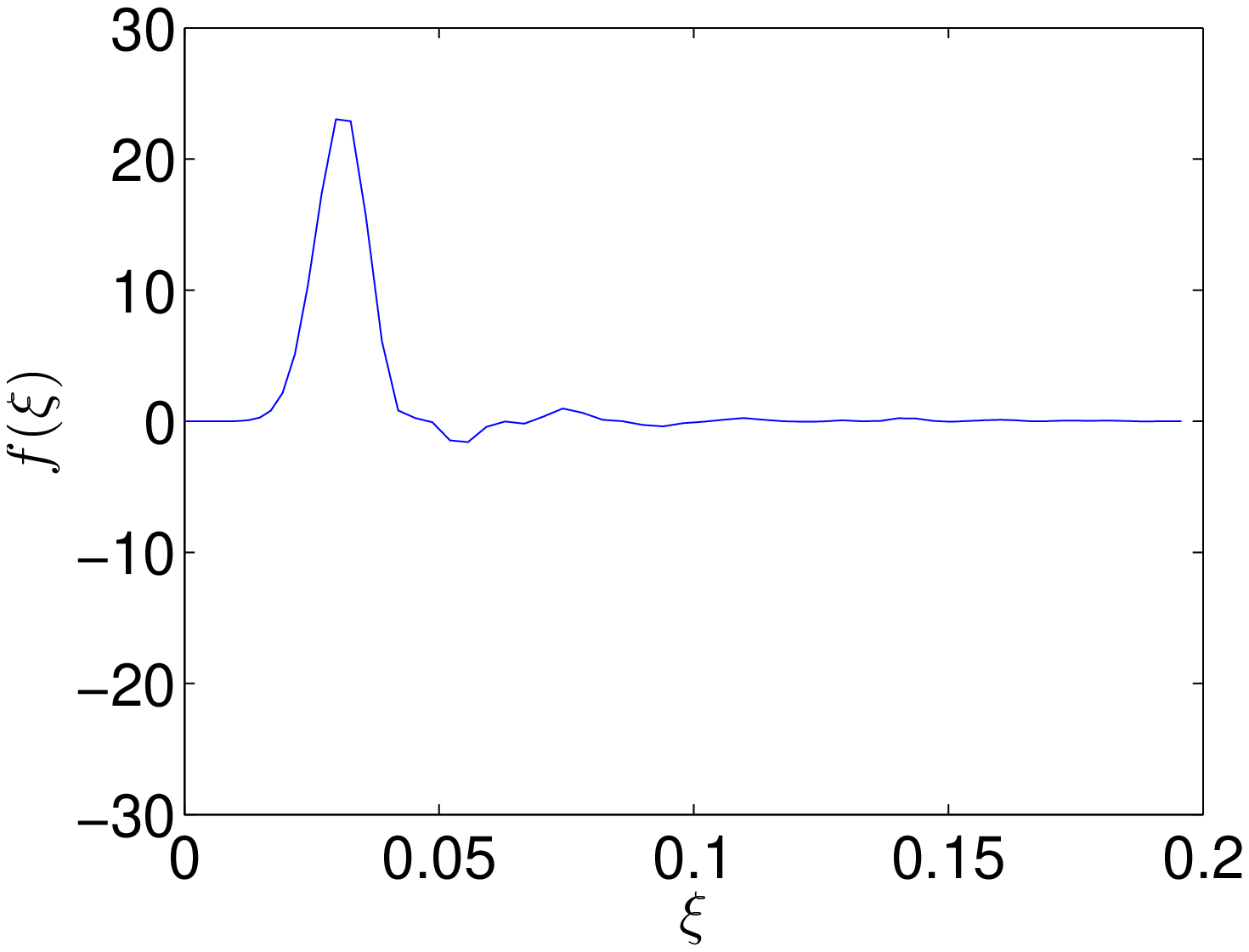} &
\includegraphics[width=0.3\columnwidth]{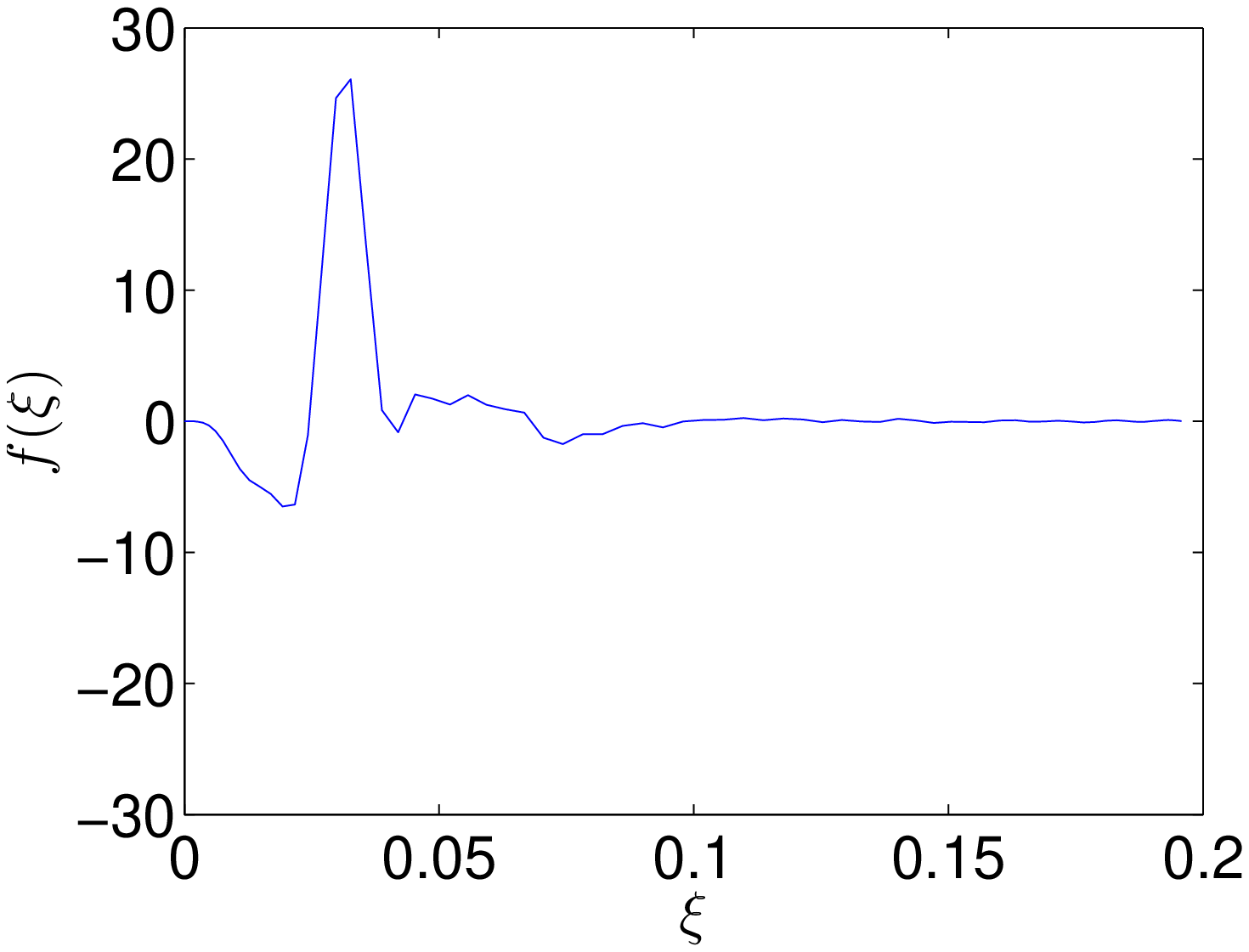} &
\includegraphics[width=0.3\columnwidth]{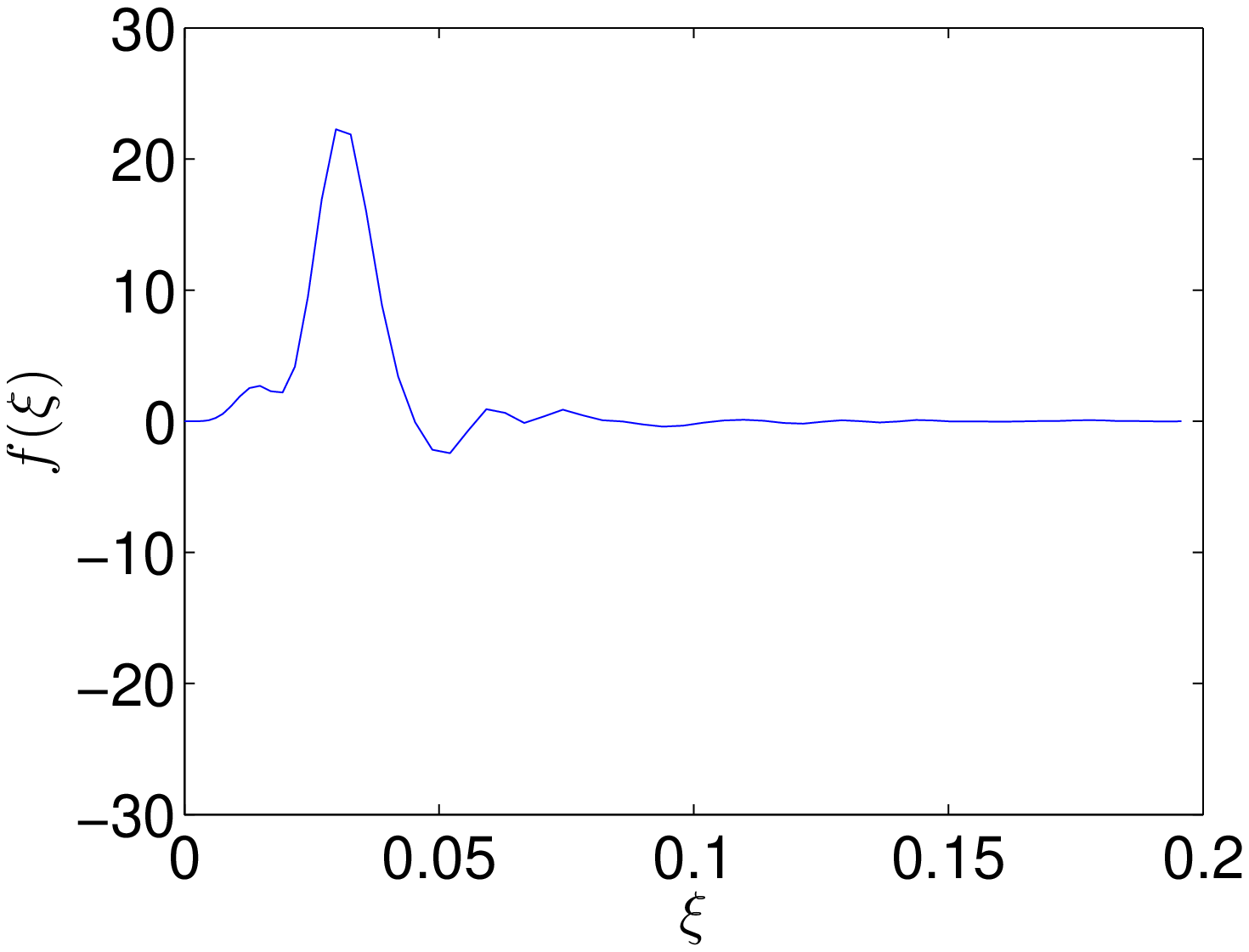} \\
$ k = 9 $, $l = 1$ & $ k = 4 $, $l = 1$ & $ k = 5 $, $l = 1$ \\
$\lambda = 153.9$ &  $\lambda = 153.4$ & $\lambda = 153.2$ \\
\includegraphics[width=0.3\columnwidth]{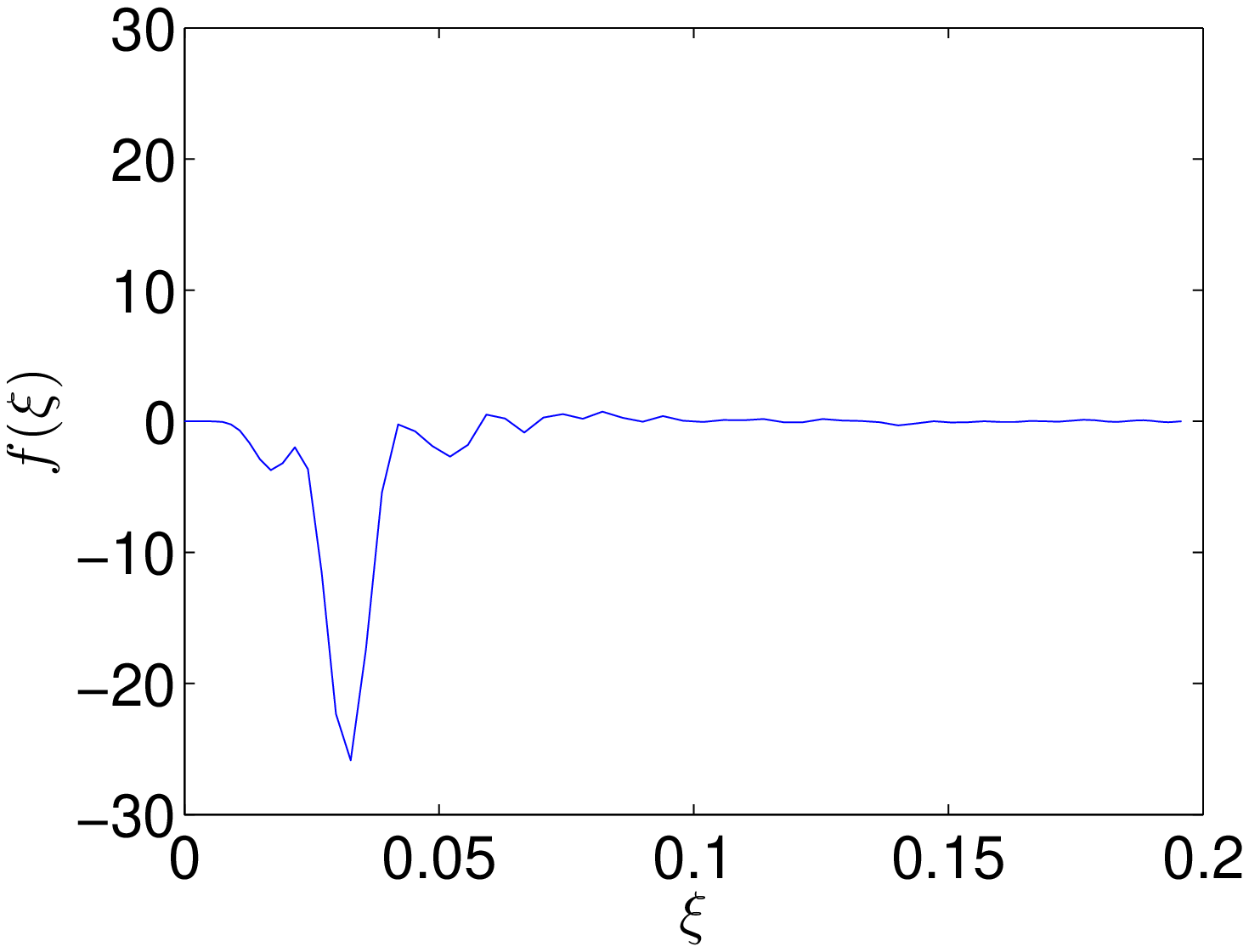} &
\includegraphics[width=0.3\columnwidth]{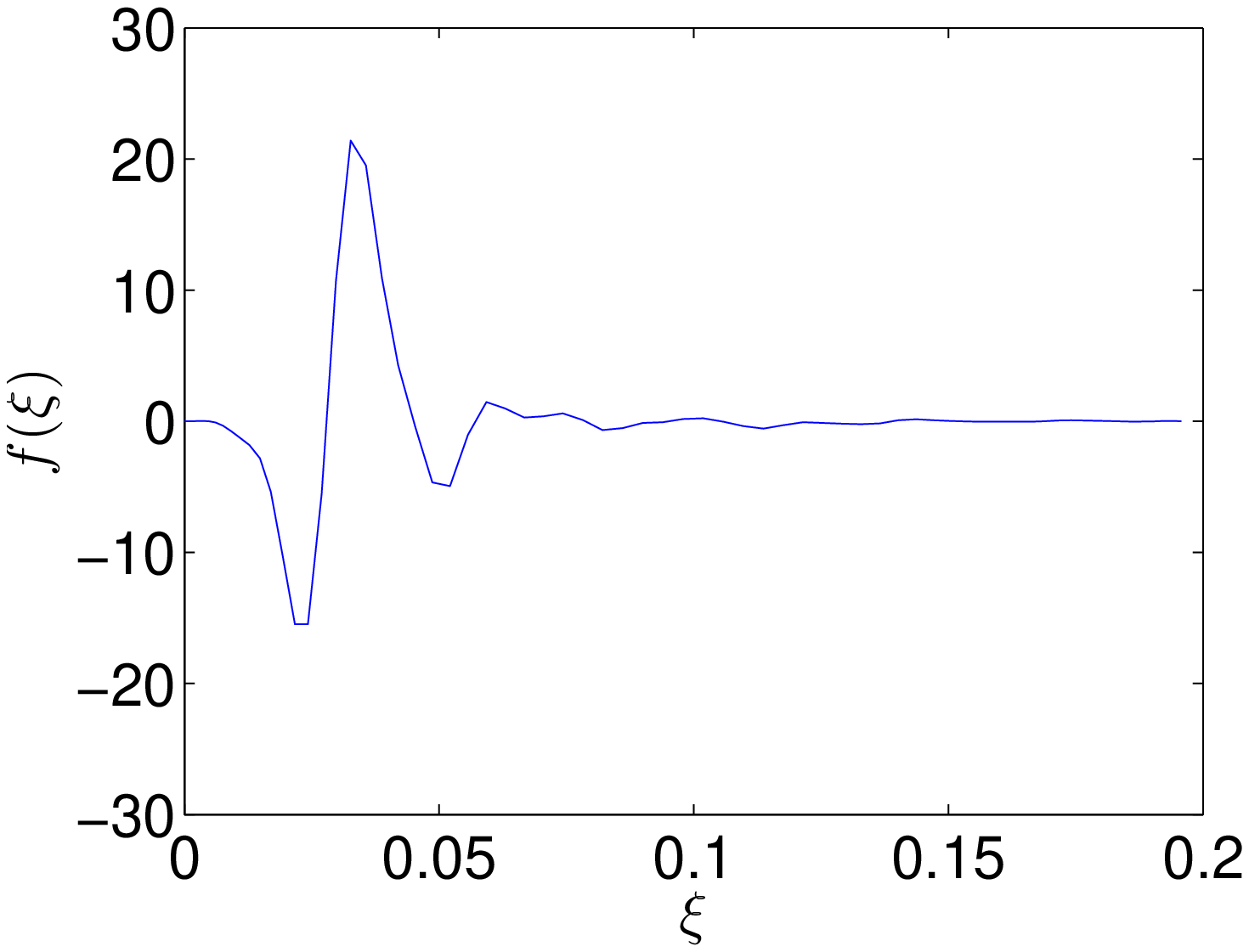} &
\includegraphics[width=0.3\columnwidth]{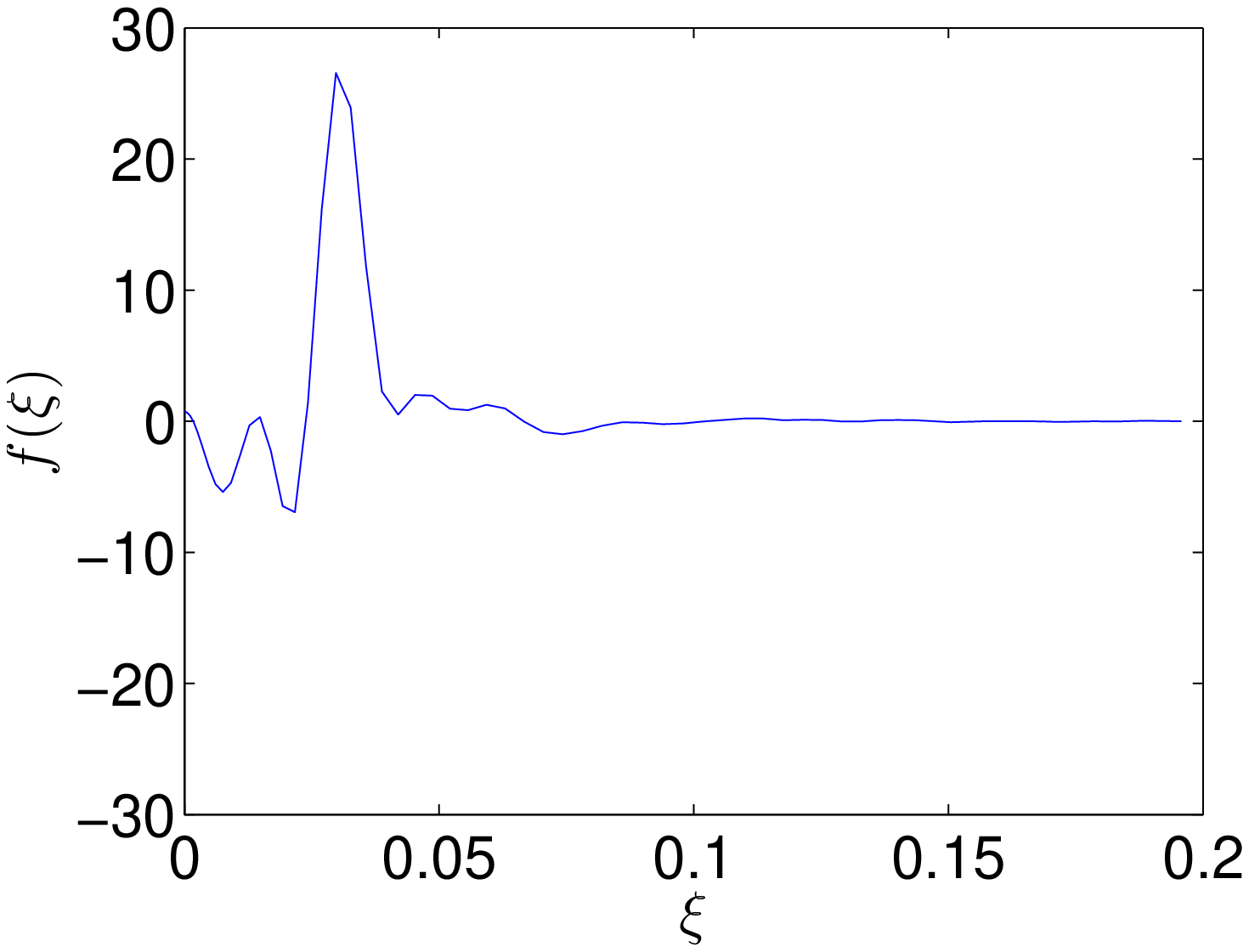} \\
$ k = 6 $, $l = 1$ & $ k = 3 $, $l = 1$ & $ k = 0 $, $l = 1$ \\
$\lambda = 150.3$ & $\lambda = 144.7$ & $\lambda = 143.0$
\end{tabular}
\end{center}
\caption{FFBsPCA principal radial functions in Fourier domain. The dataset contains $n = 10^5$ simulated human mitochondrial large ribosomal subunit projection images corrupted by additive white Gaussian noise with SNR$= 1/30$. Image size is $240 \times 240$ pixels, $R = 98$, $c = 0.196$.  Each radial function is labeled with angular index $k$, radial order $l$, and eigenvalue $\lambda$.}
\label{fig:spca_radial}
\end{figure}
The radial functions of the top nine principal components are shown in Fig.~\ref{fig:spca_radial}. Each radial function is indexed by $k$ and $l$, where $k$ determines the angular Fourier mode and $l$ is the order of the radial function within the same $k$. Taking the tensor product of the radial functions and their corresponding angular Fourier modes gives the two dimensional principal components in Fourier domain.
It took about 9 minutes in total to get the steerable PCA radial components and the associated expansion coefficients. In particular, Fourier-Bessel expansion coefficients were computed in 9 minutes and the steerable PCA took 12 seconds.

\begin{figure}[htb]
\begin{center}
\begin{tabular}{cccc}
\includegraphics[width=0.2\columnwidth]{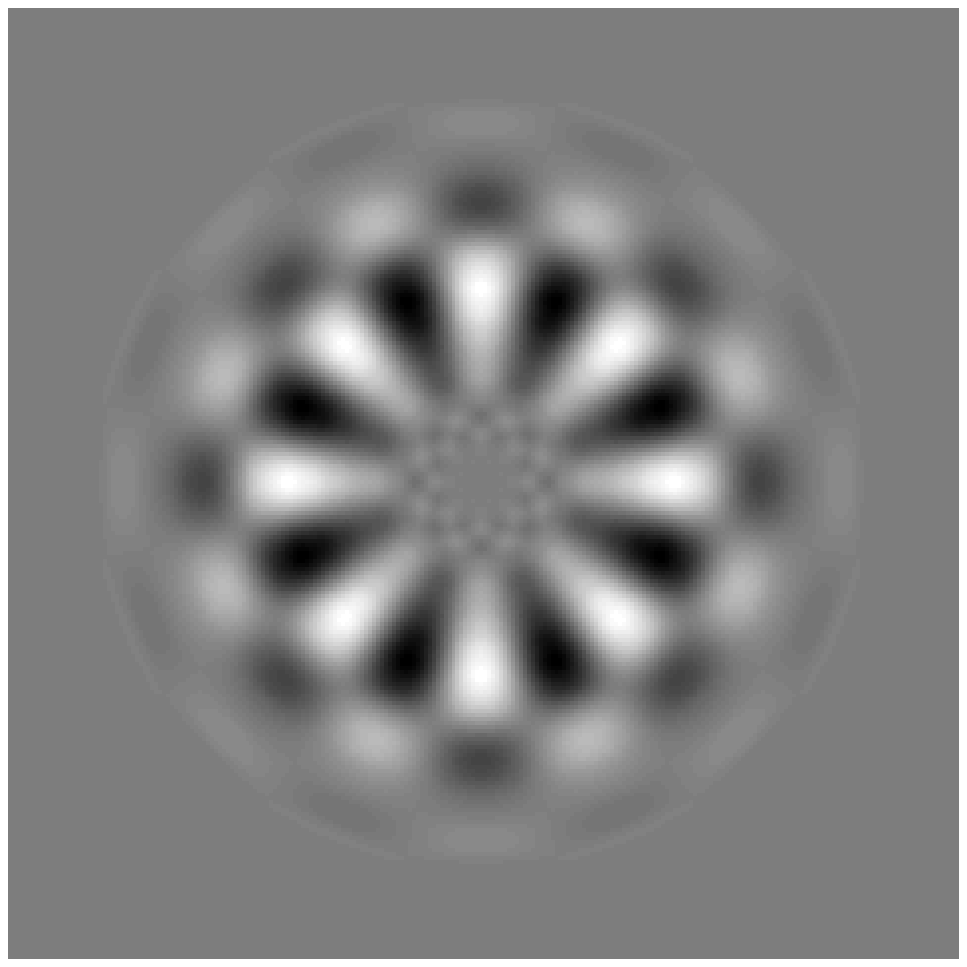} &
\includegraphics[width=0.2\columnwidth]{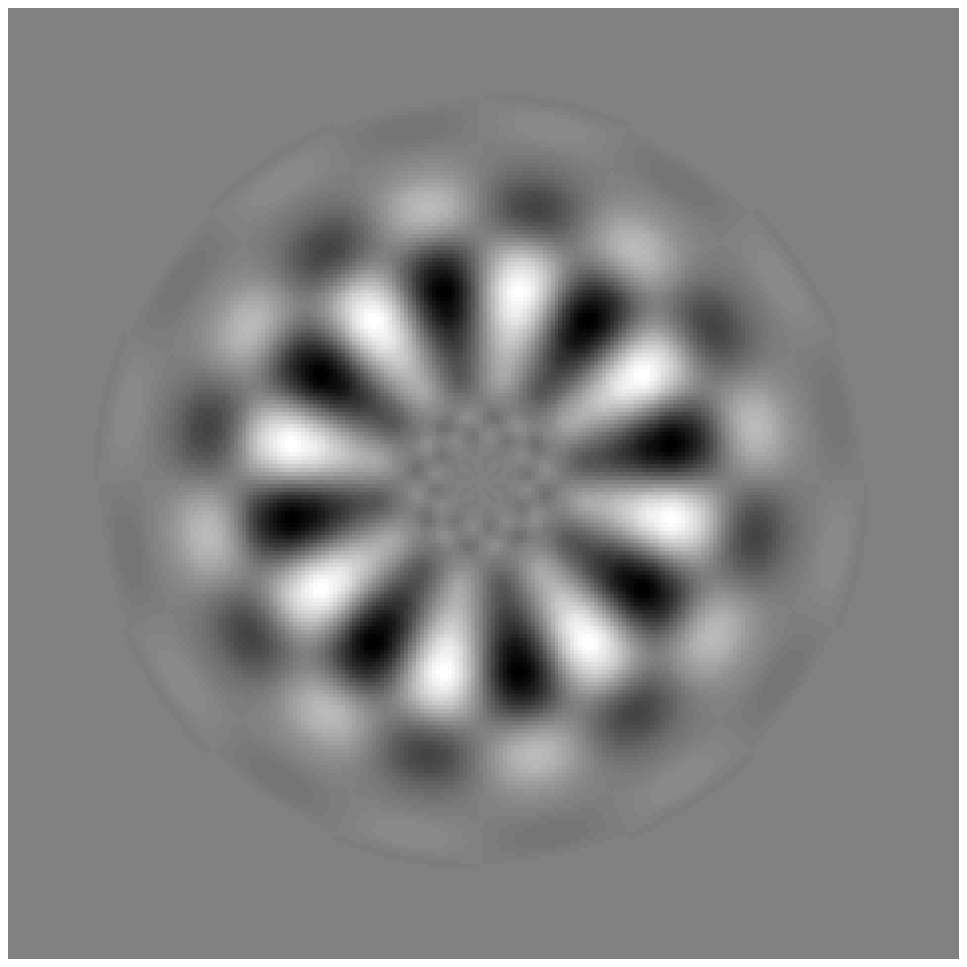} &
\includegraphics[width=0.2\columnwidth]{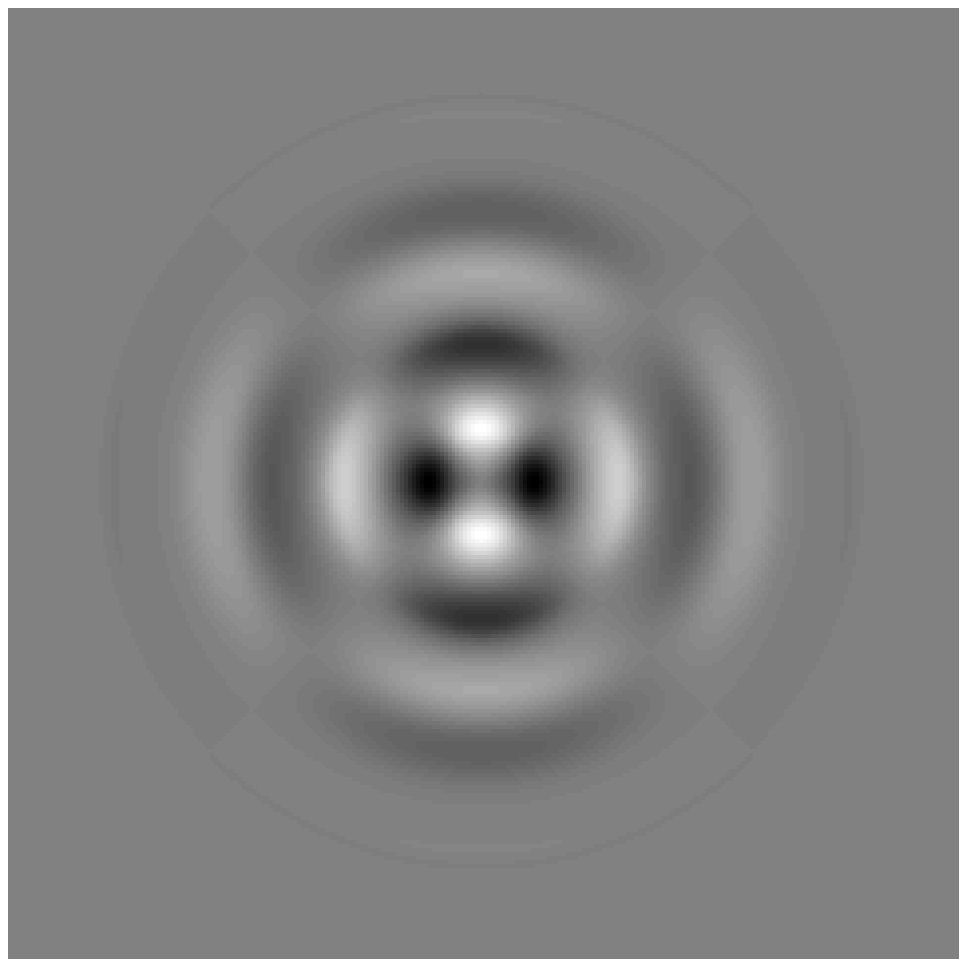} &
\includegraphics[width=0.2\columnwidth]{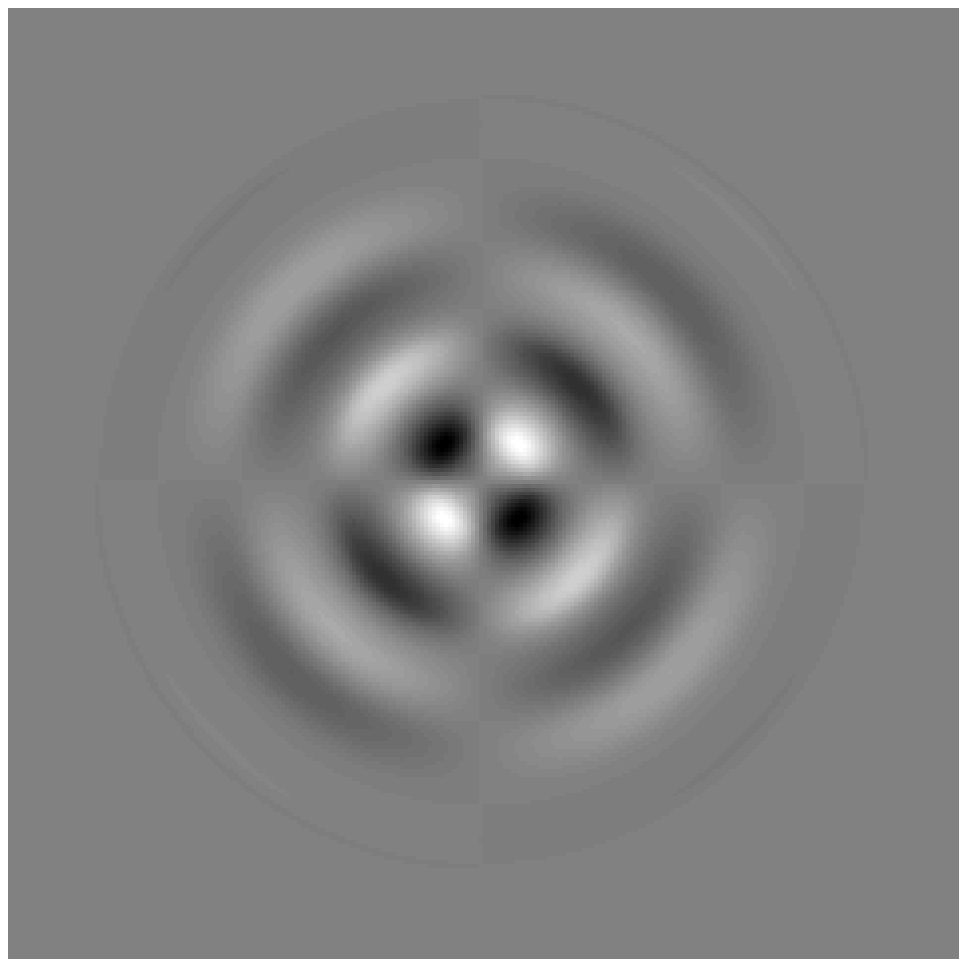} \\
$\lambda = 163.8$ & $\lambda = 163.8 $ & $\lambda = 160.0 $ & $\lambda = 160.0 $ \\
\includegraphics[width=0.2\columnwidth]{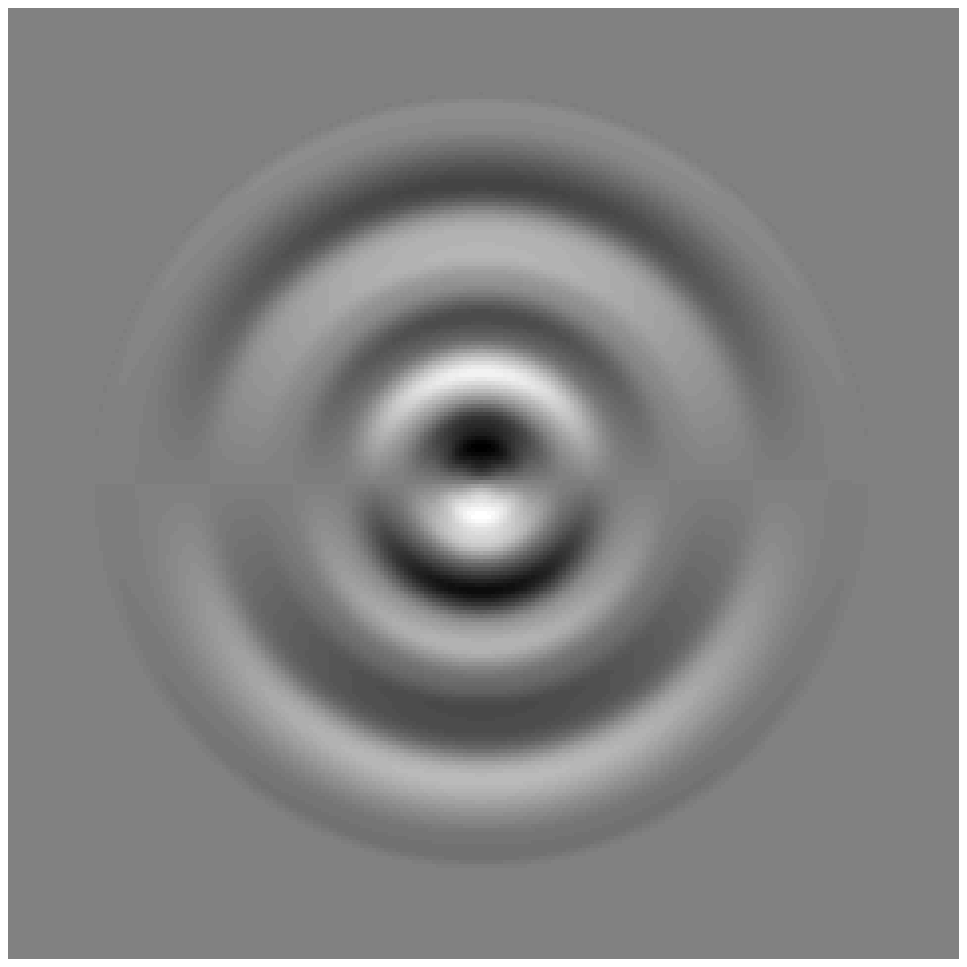} &
\includegraphics[width=0.2\columnwidth]{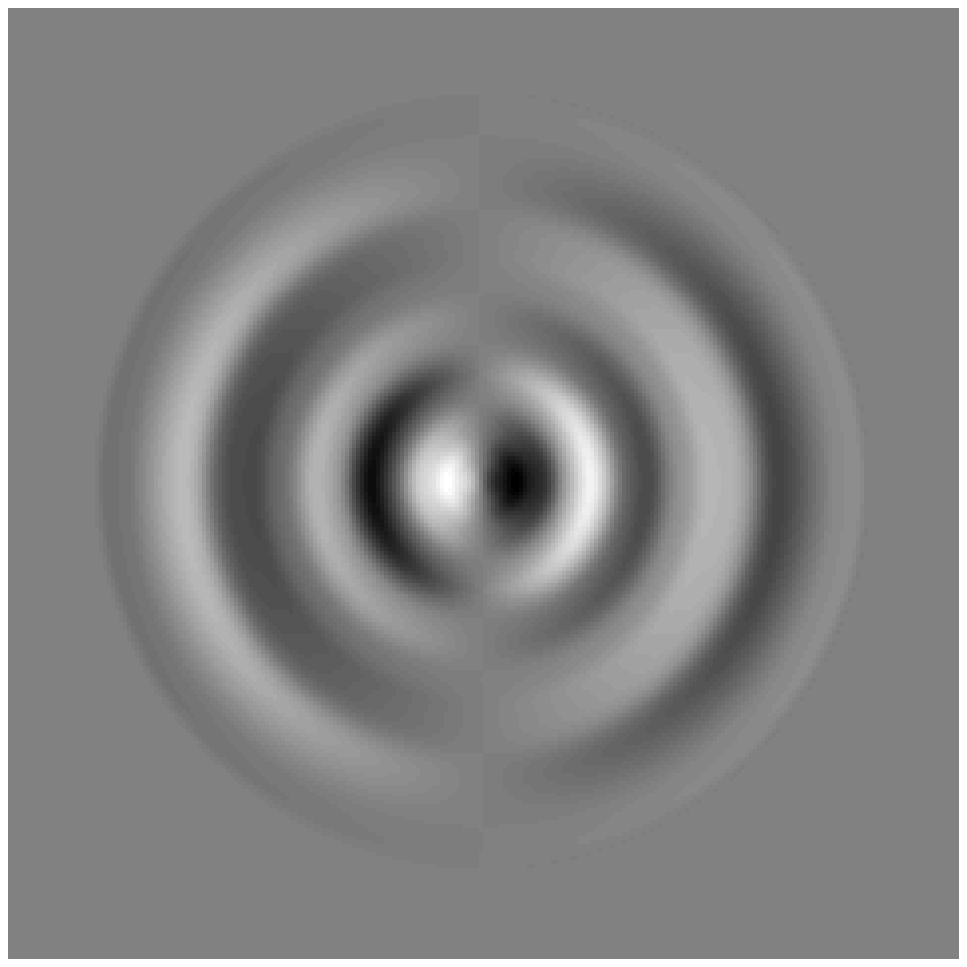} &
\includegraphics[width=0.2\columnwidth]{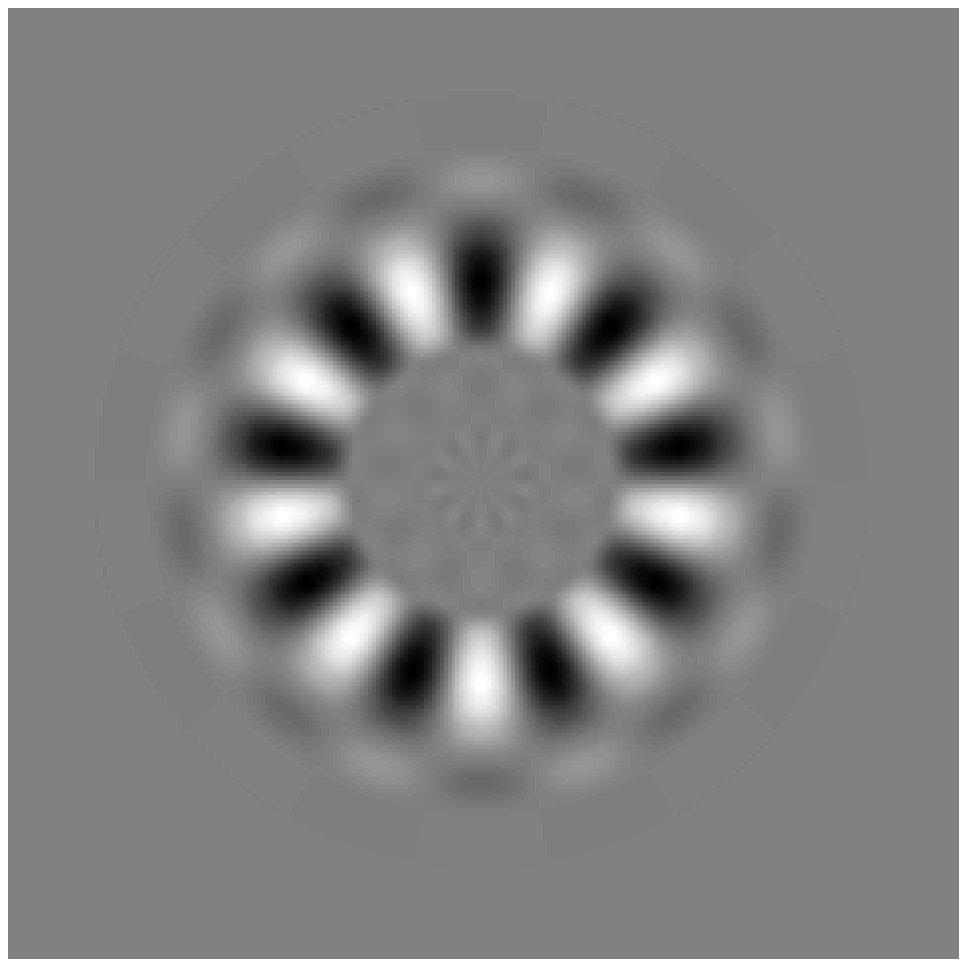} &
\includegraphics[width=0.2\columnwidth]{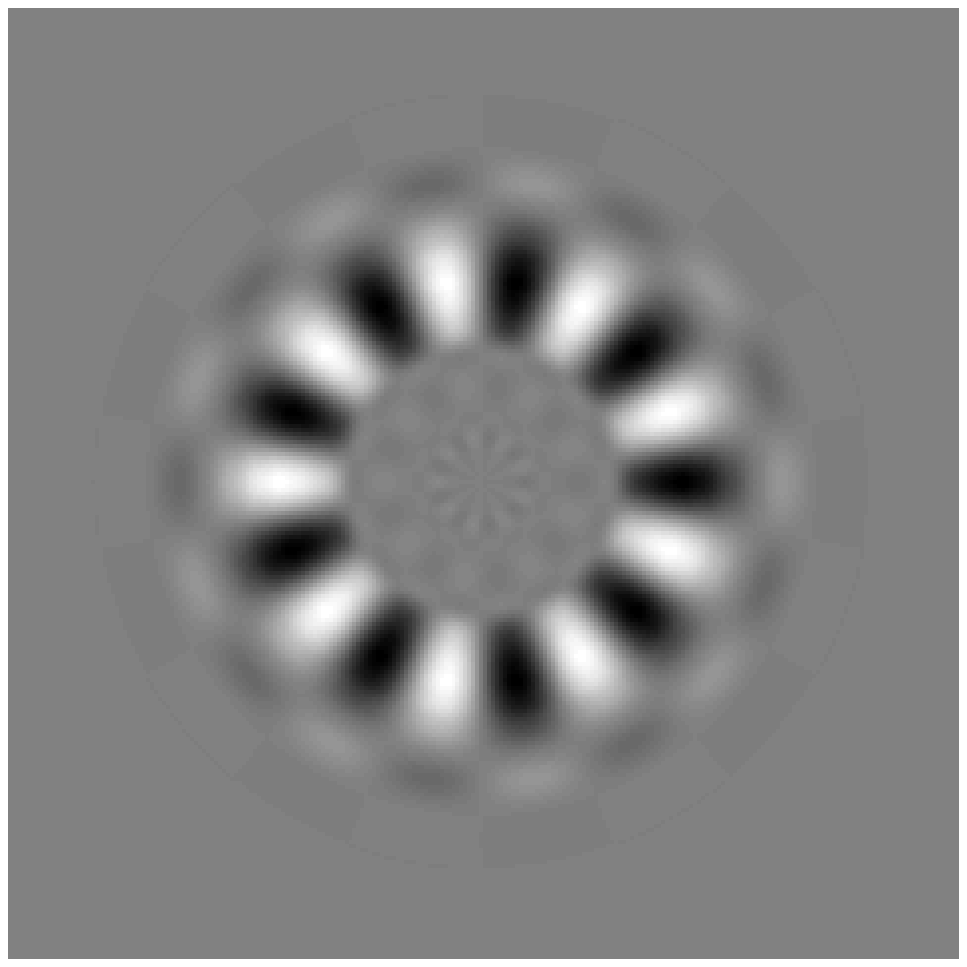} \\
$\lambda = 158.1 $ & $\lambda = 158.1 $ & $\lambda = 153.9$ & $\lambda = 153.9 $ \\
\includegraphics[width=0.2\columnwidth]{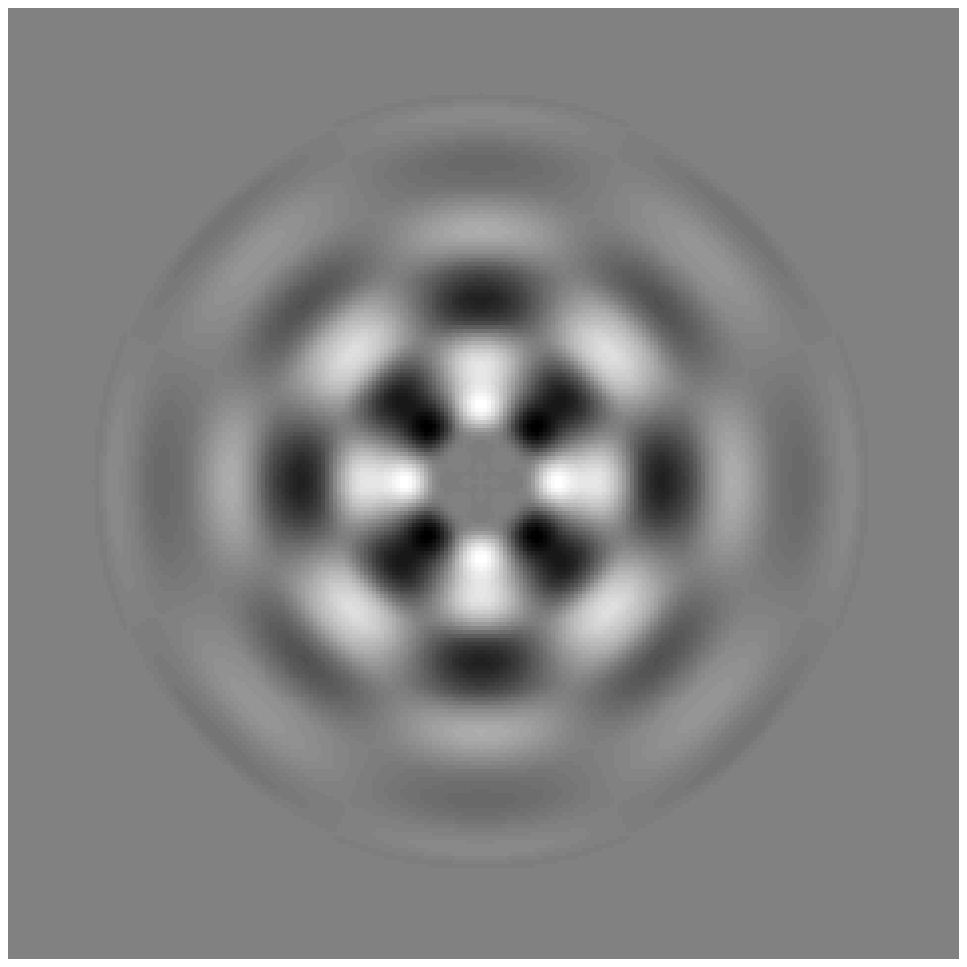} &
\includegraphics[width=0.2\columnwidth]{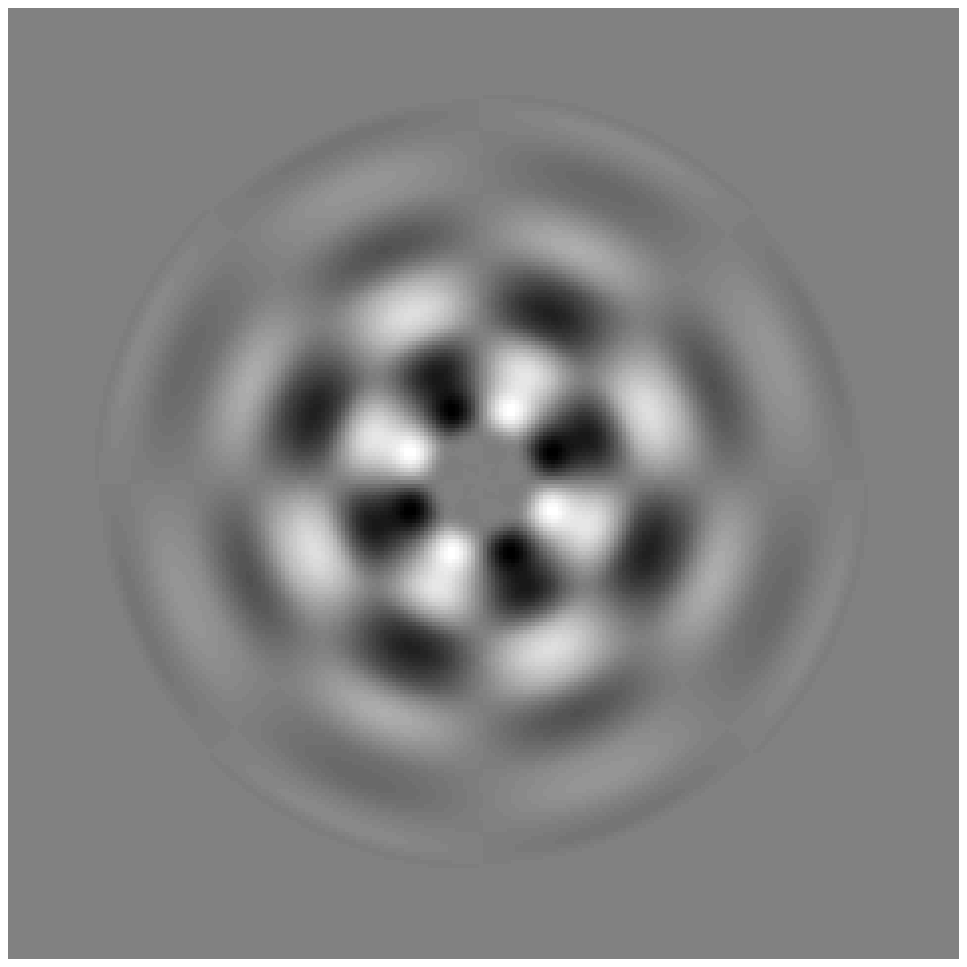} &
\includegraphics[width=0.2\columnwidth]{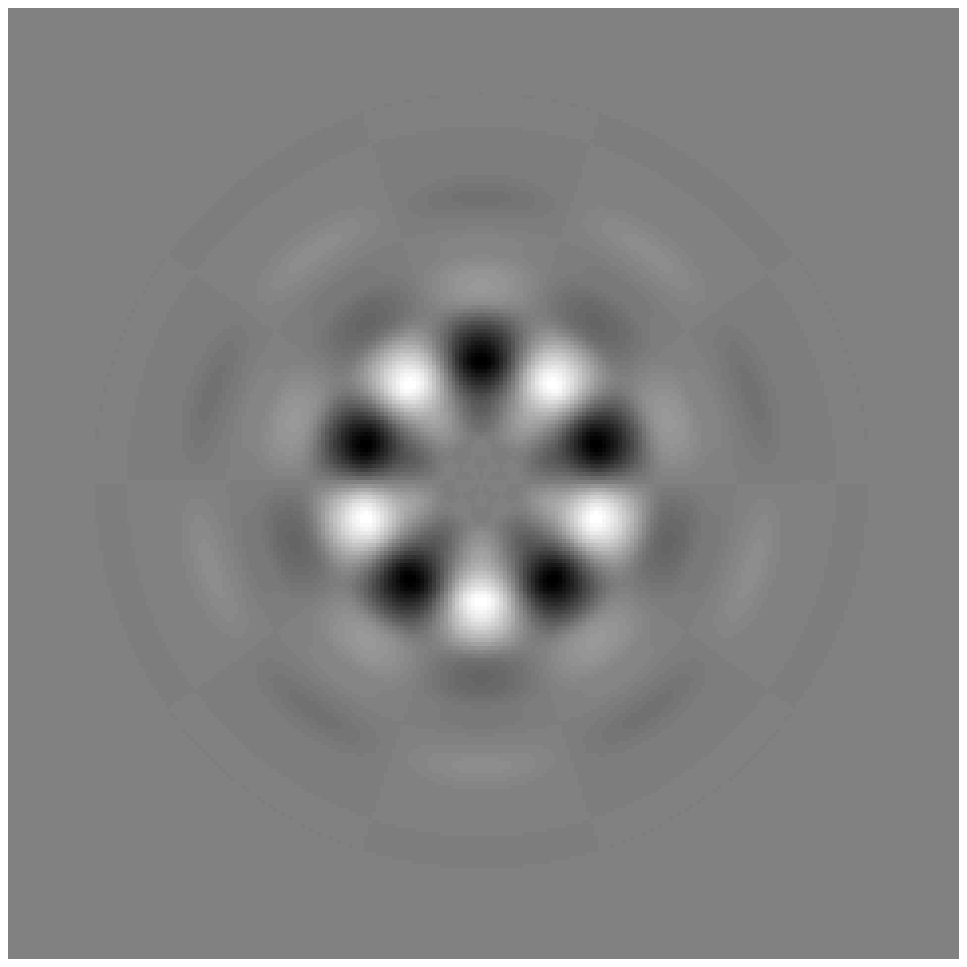} &
\includegraphics[width=0.2\columnwidth]{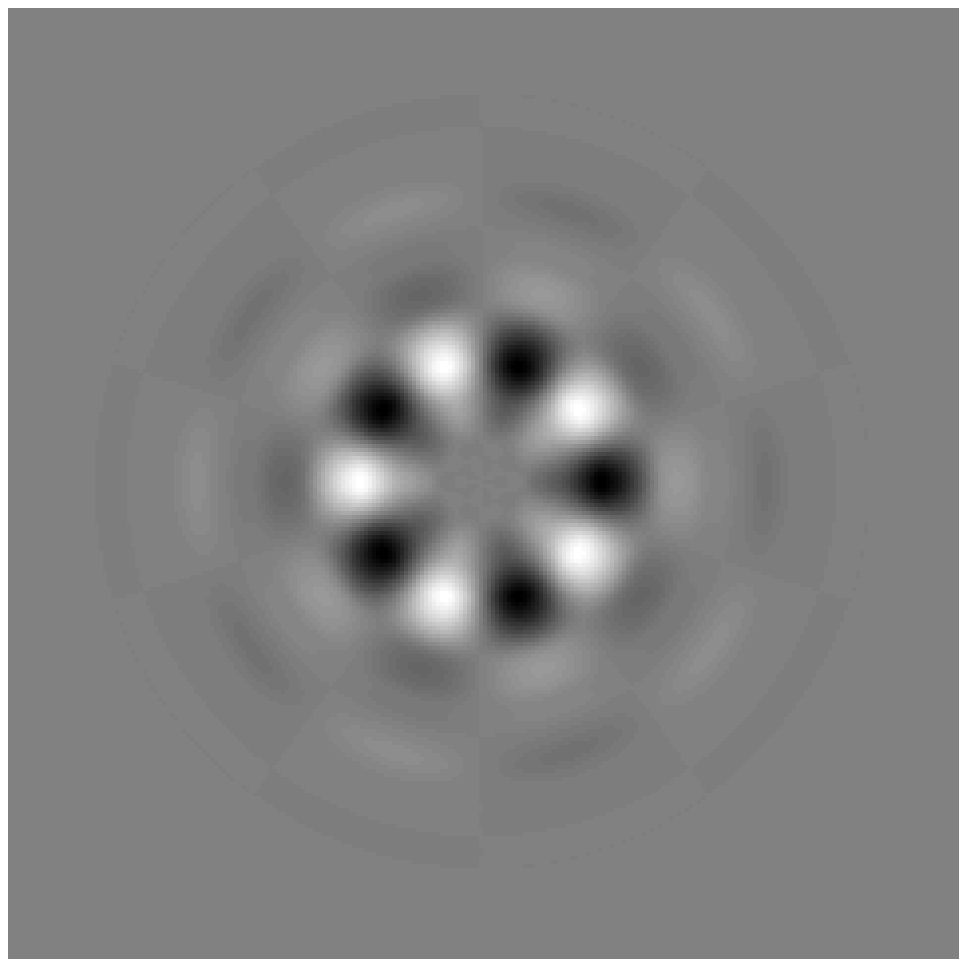} \\
$\lambda = 153.4$ & $\lambda = 153.4 $ & $\lambda = 153.2 $ & $\lambda = 153.2$ \\
\includegraphics[width=0.2\columnwidth]{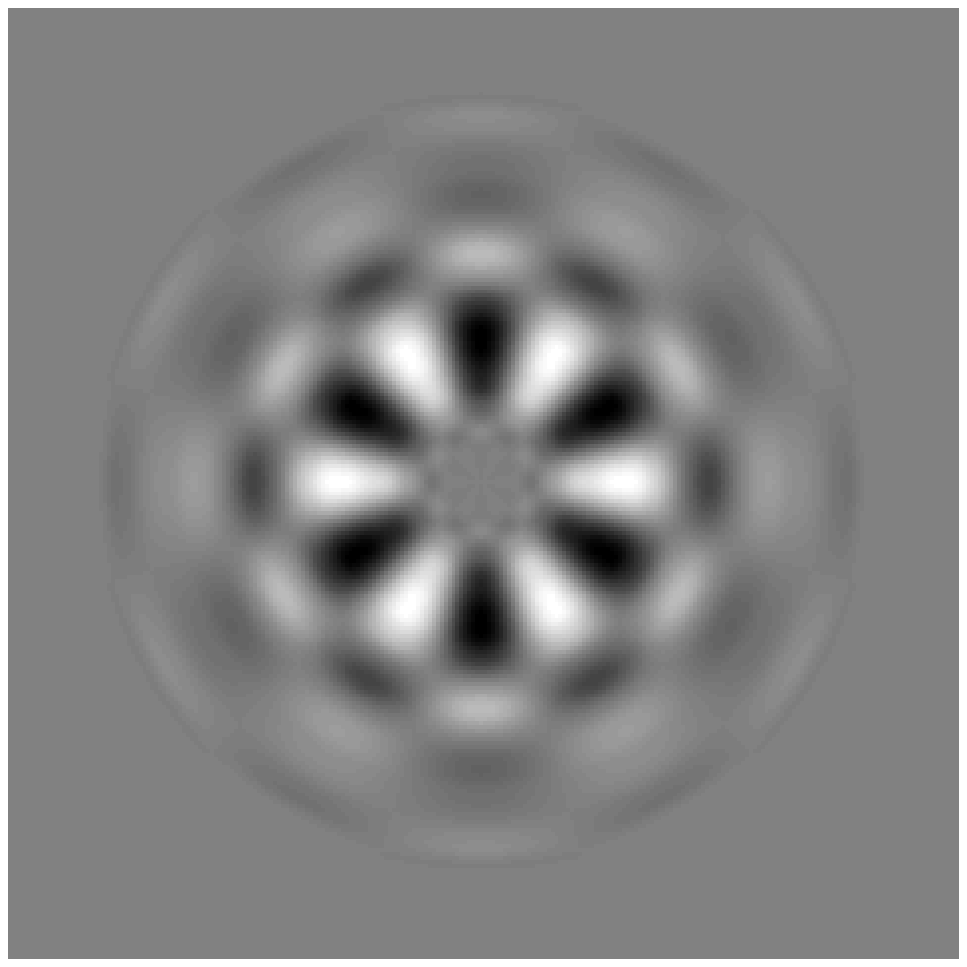} &
\includegraphics[width=0.2\columnwidth]{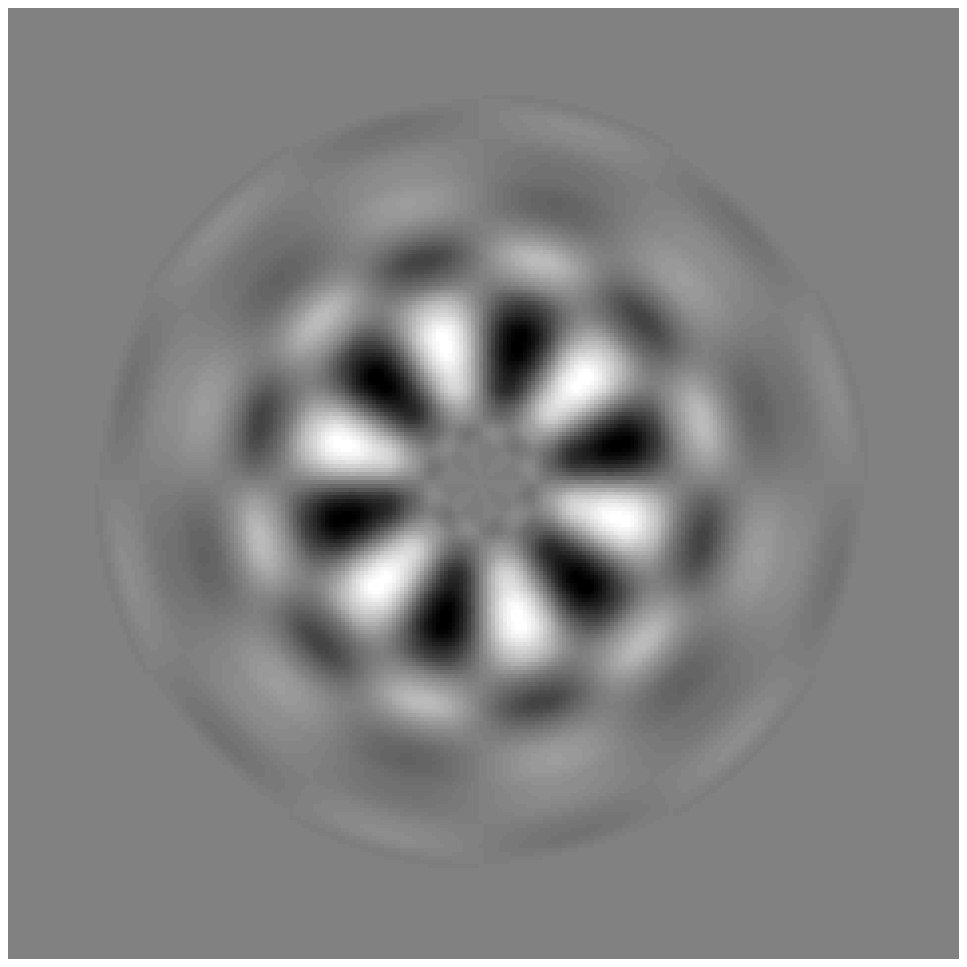} &
\includegraphics[width=0.2\columnwidth]{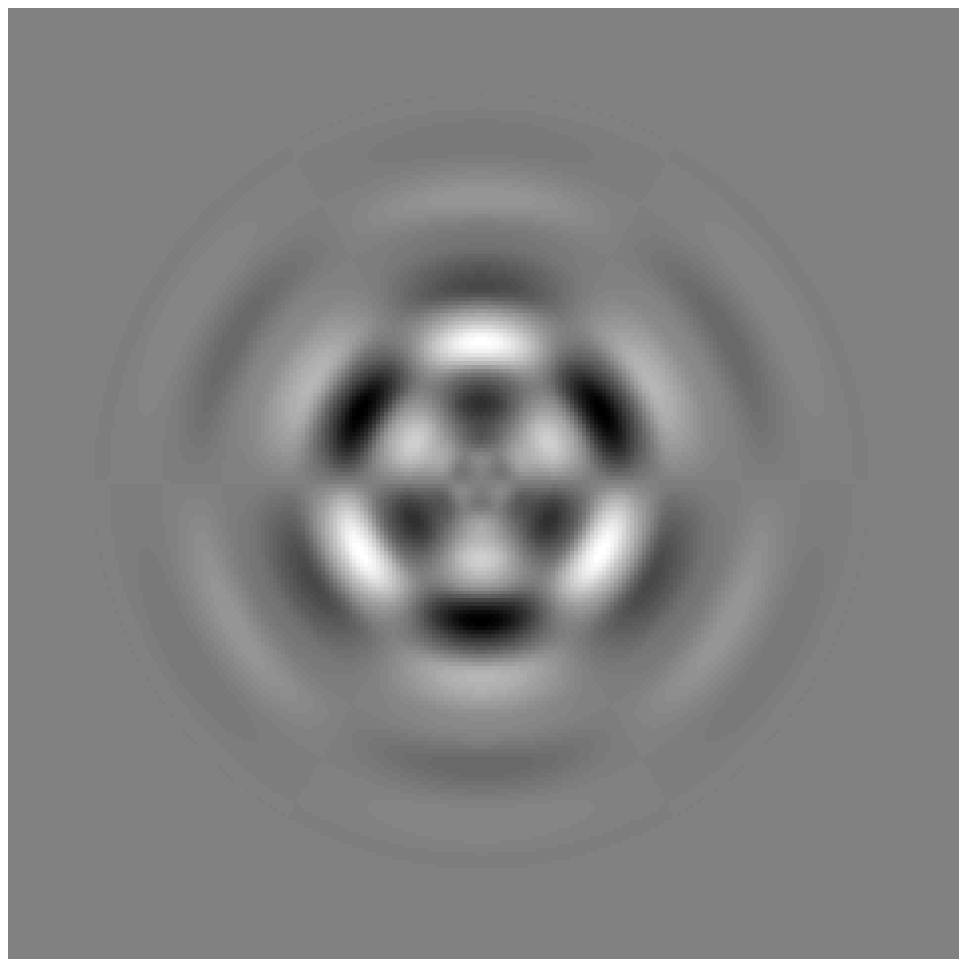} &
\includegraphics[width=0.2\columnwidth]{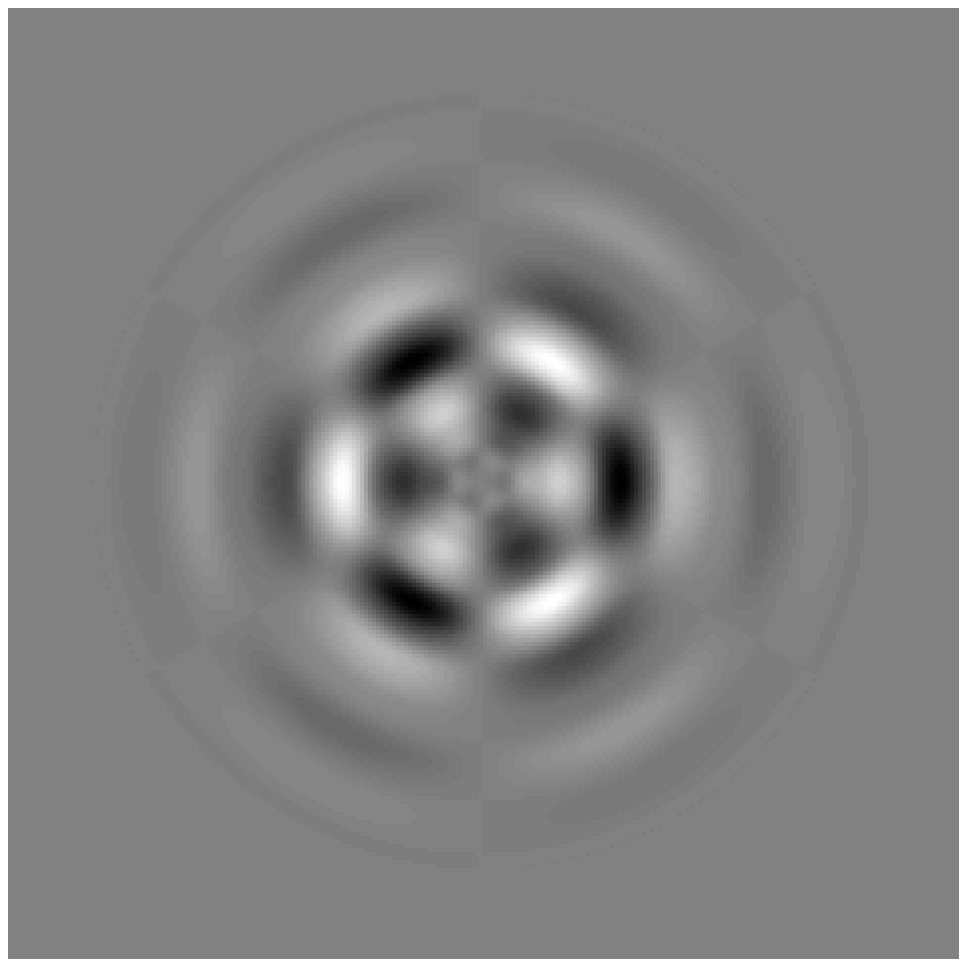} \\
$\lambda = 150.3 $ & $\lambda = 150.3  $ & $\lambda = 144.7$ & $\lambda = 144.7 $
\end{tabular}
\end{center}
\caption{FFBsPCA principal components (eigenimages in real domain) corresponding to Figure~\ref{fig:spca_radial}.}
\label{fig:spca_basis_real}
\end{figure}
\begin{figure}[htb]
\begin{center}
\begin{tabular}{cccc}
\includegraphics[width=0.2\columnwidth]{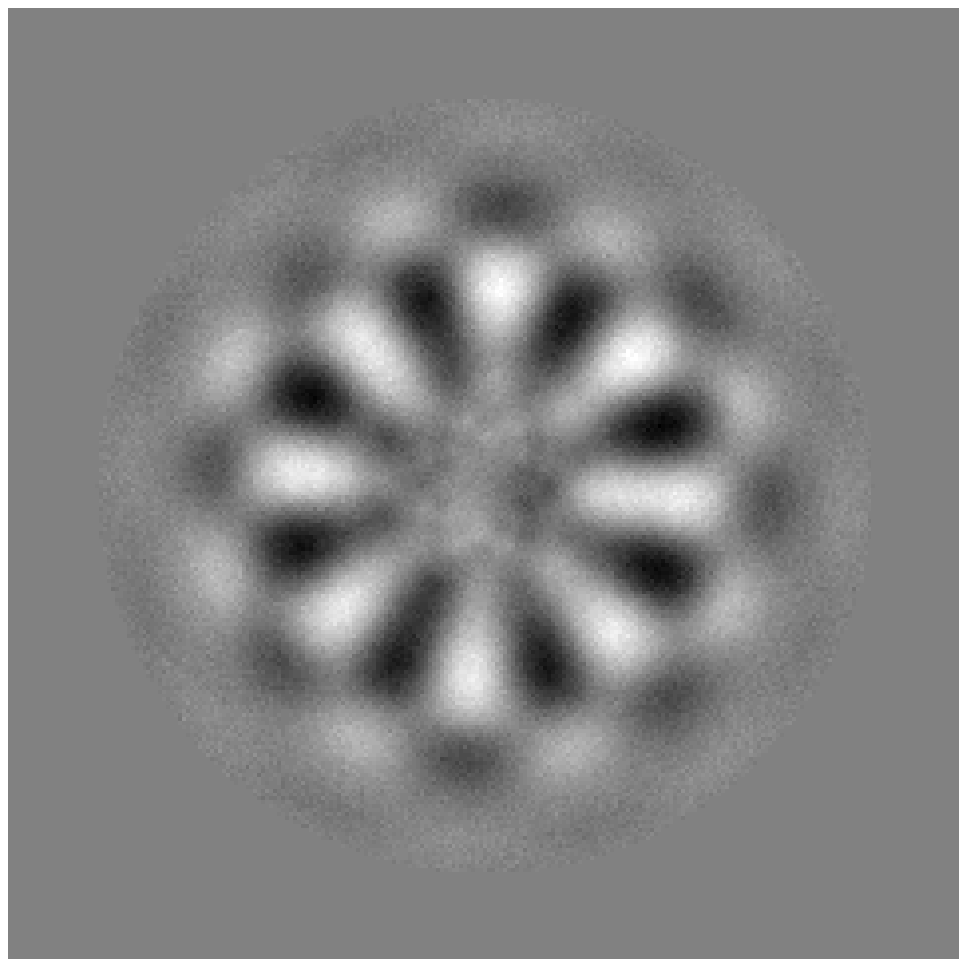} &
\includegraphics[width=0.2\columnwidth]{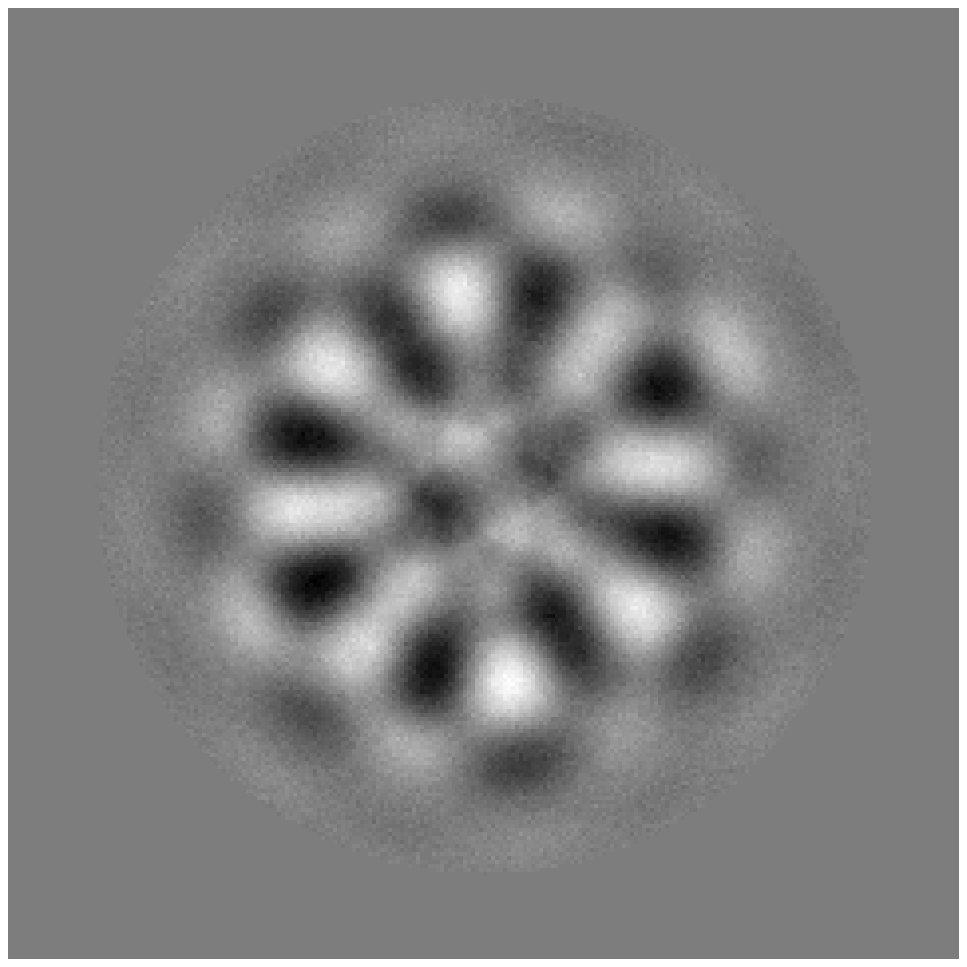} &
\includegraphics[width=0.2\columnwidth]{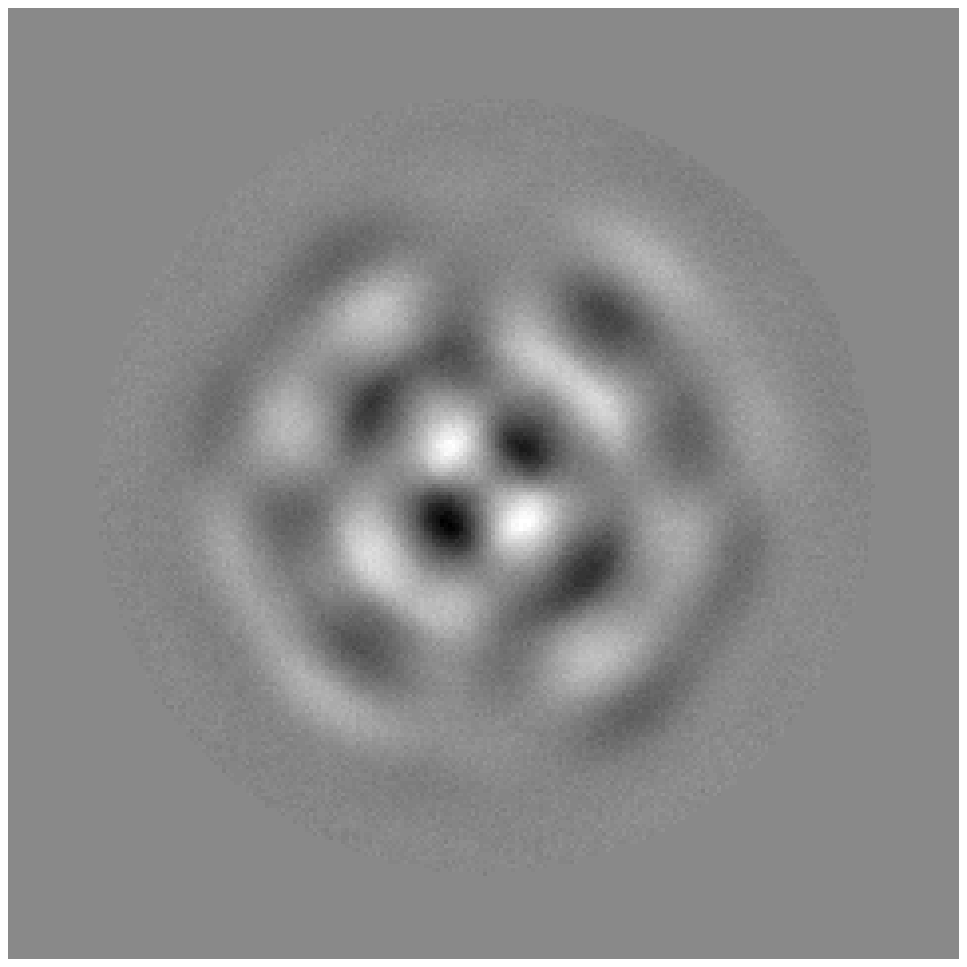} &
\includegraphics[width=0.2\columnwidth]{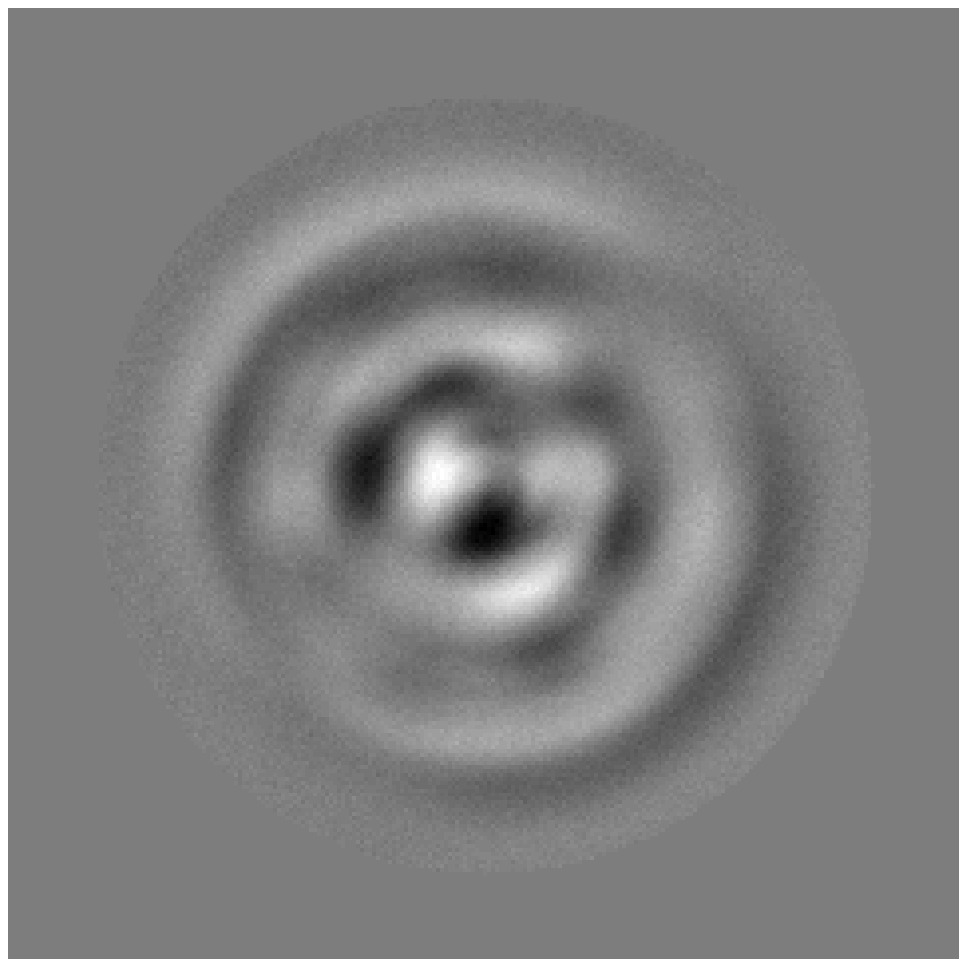} \\
$\lambda = 168.0 $ & $\lambda = 167.0$ & $\lambda = 164.7$ & $\lambda = 162.6 $ \\
\includegraphics[width=0.2\columnwidth]{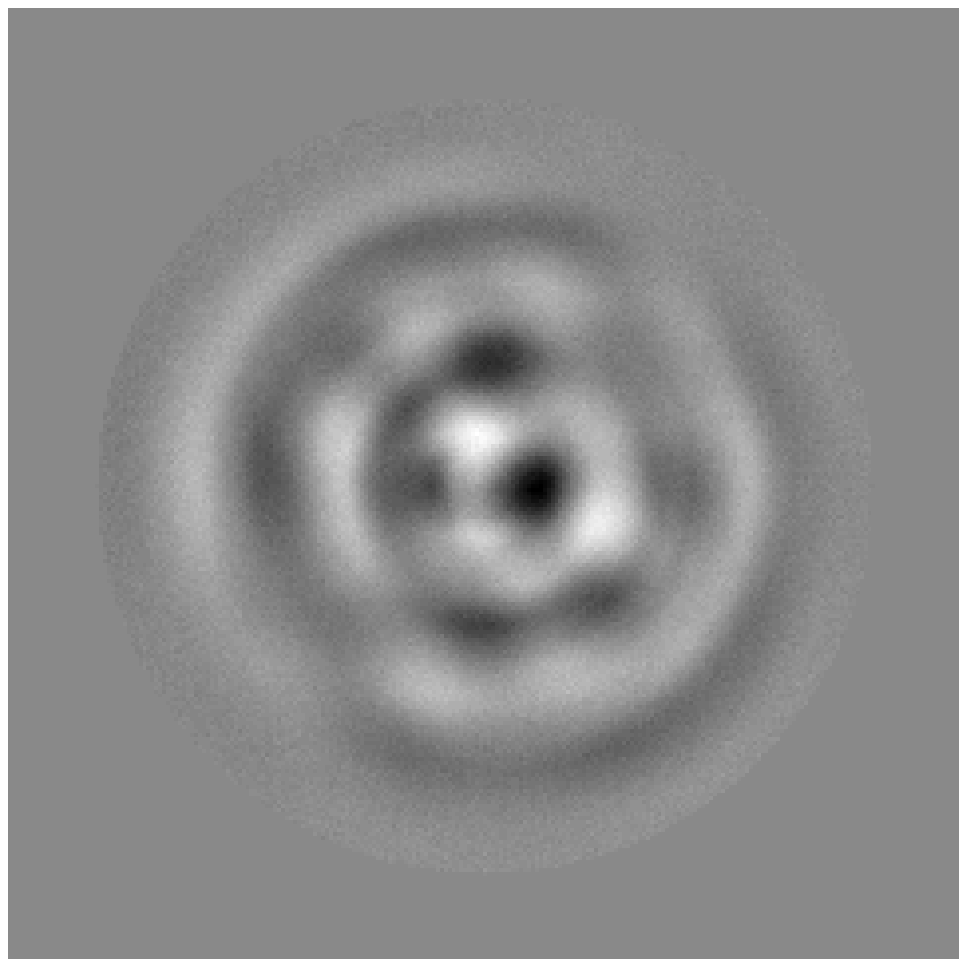} &
\includegraphics[width=0.2\columnwidth]{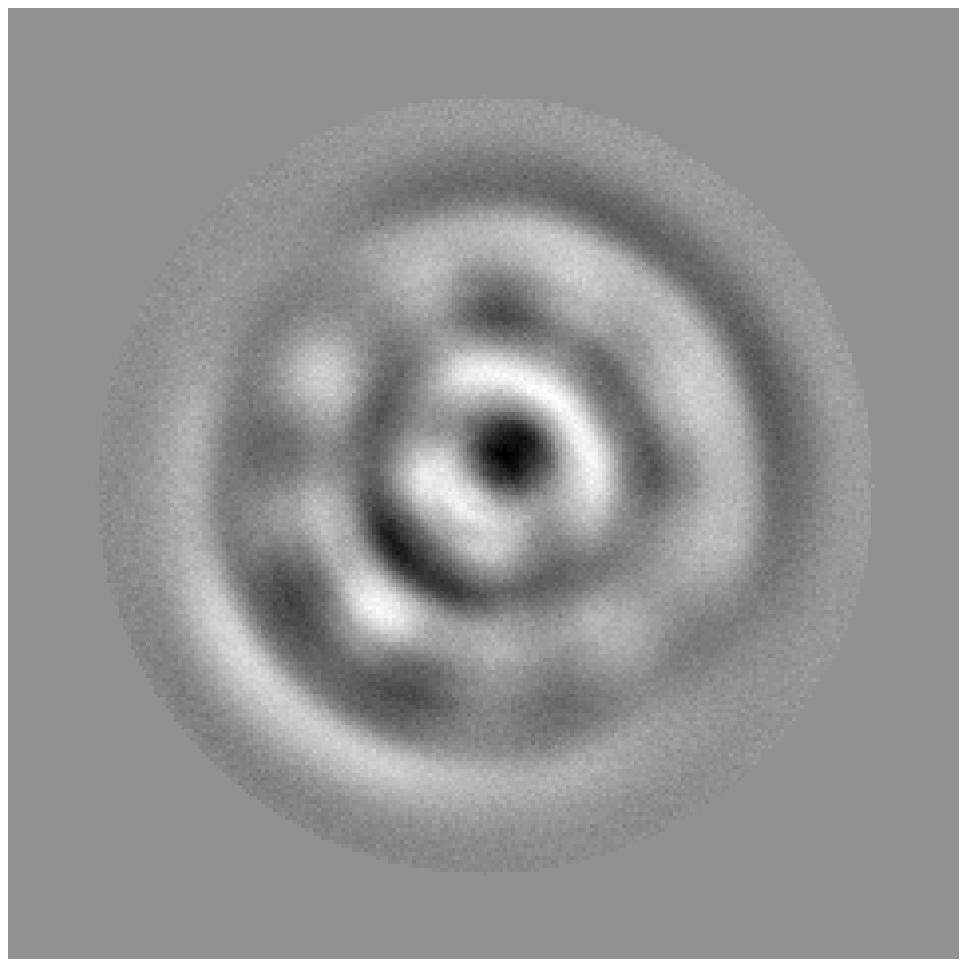} &
\includegraphics[width=0.2\columnwidth]{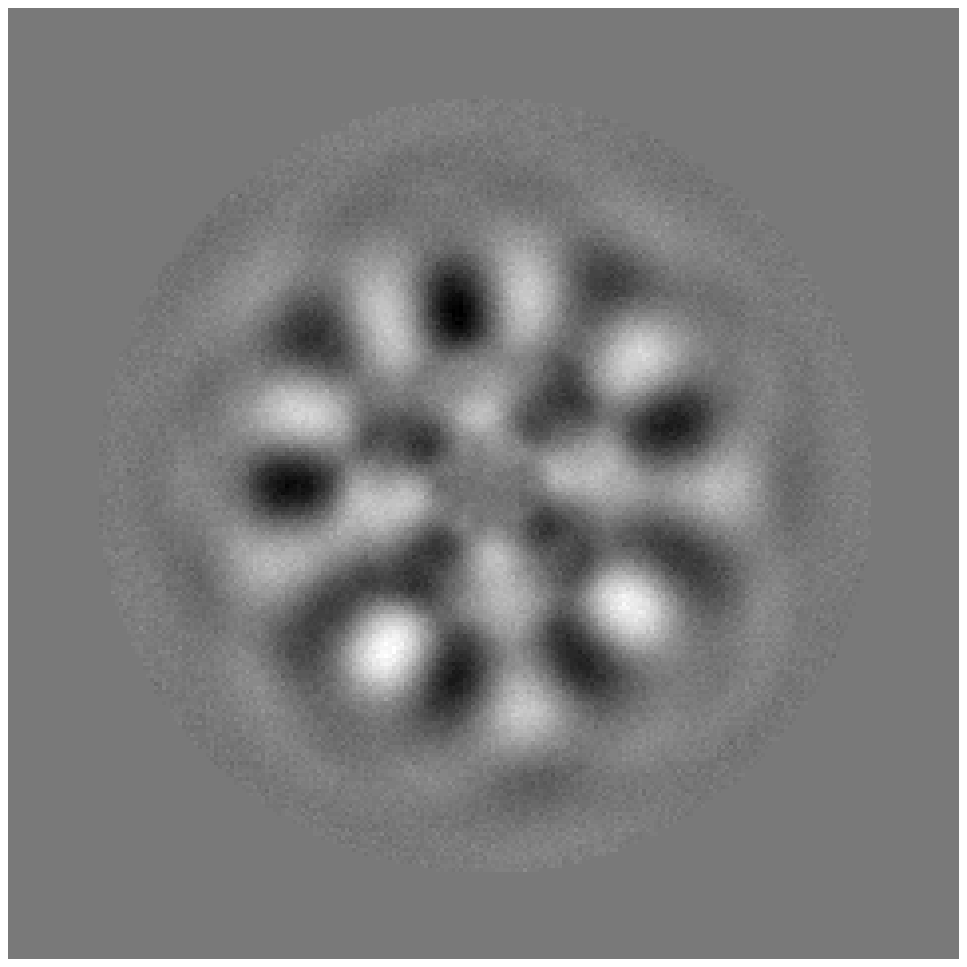} &
\includegraphics[width=0.2\columnwidth]{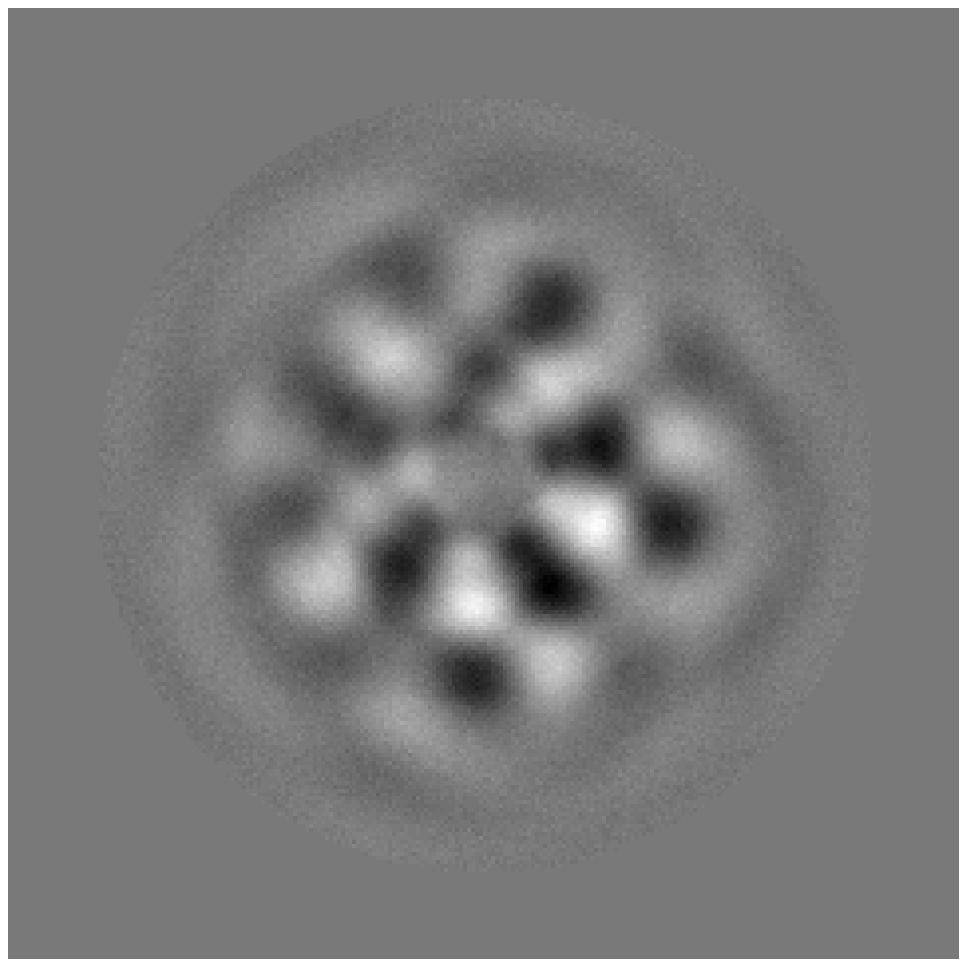} \\
$\lambda = 161.9 $ & $\lambda = 161.3$ & $\lambda = 158.7 $ & $\lambda = 157.6 $ \\
\includegraphics[width=0.2\columnwidth]{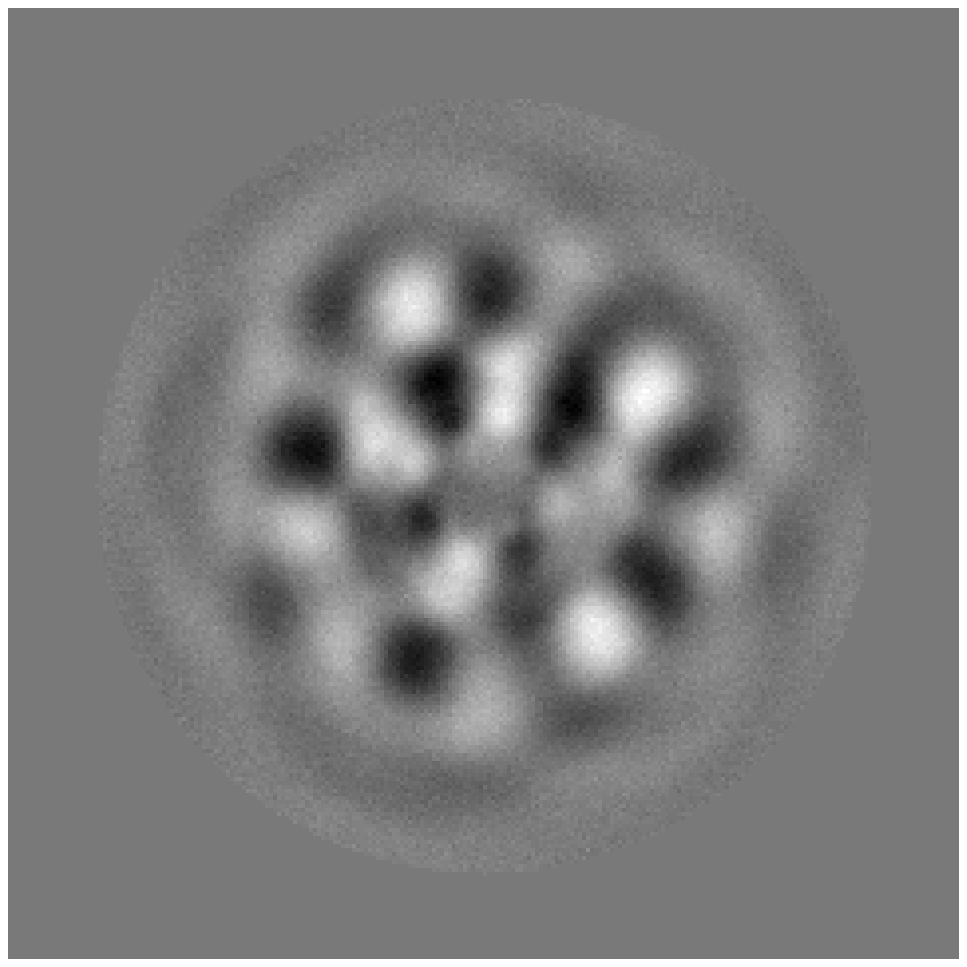} &
\includegraphics[width=0.2\columnwidth]{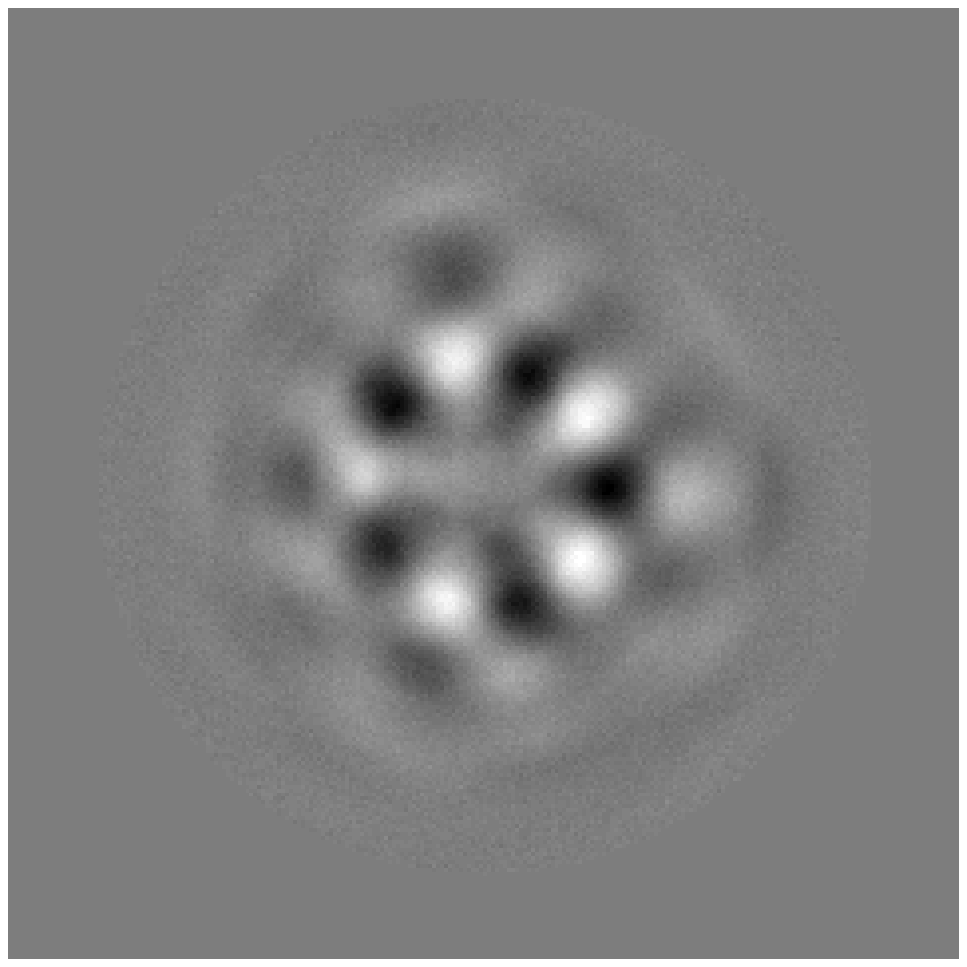} &
\includegraphics[width=0.2\columnwidth]{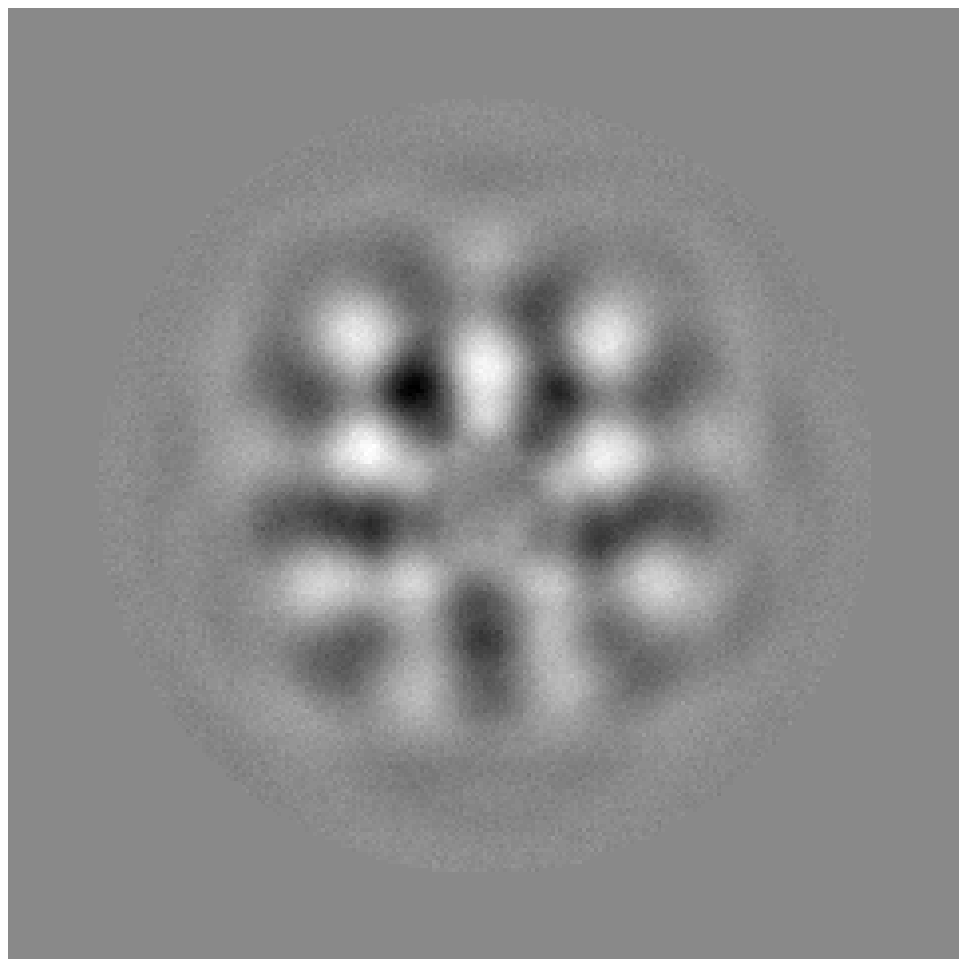} &
\includegraphics[width=0.2\columnwidth]{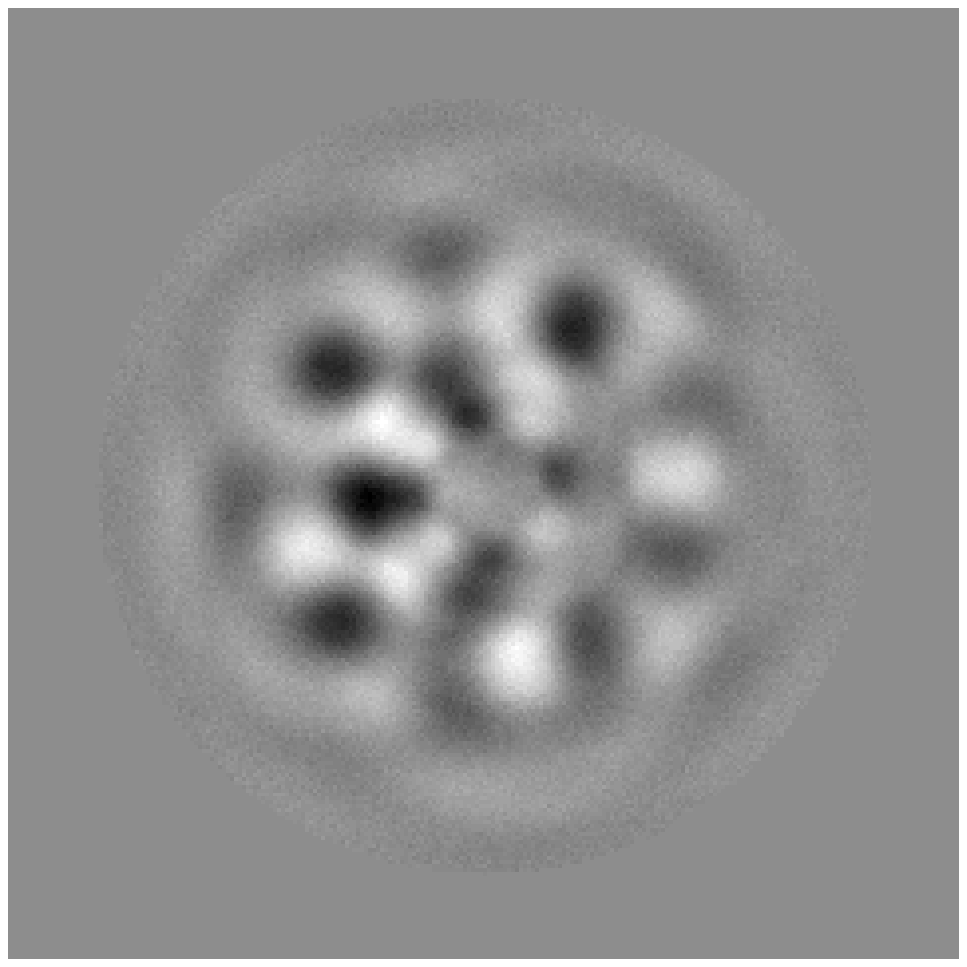} \\
$\lambda = 157.1$ & $\lambda = 156.0 $ & $\lambda = 155.9$ & $\lambda = 154.9 $ \\
\includegraphics[width=0.2\columnwidth]{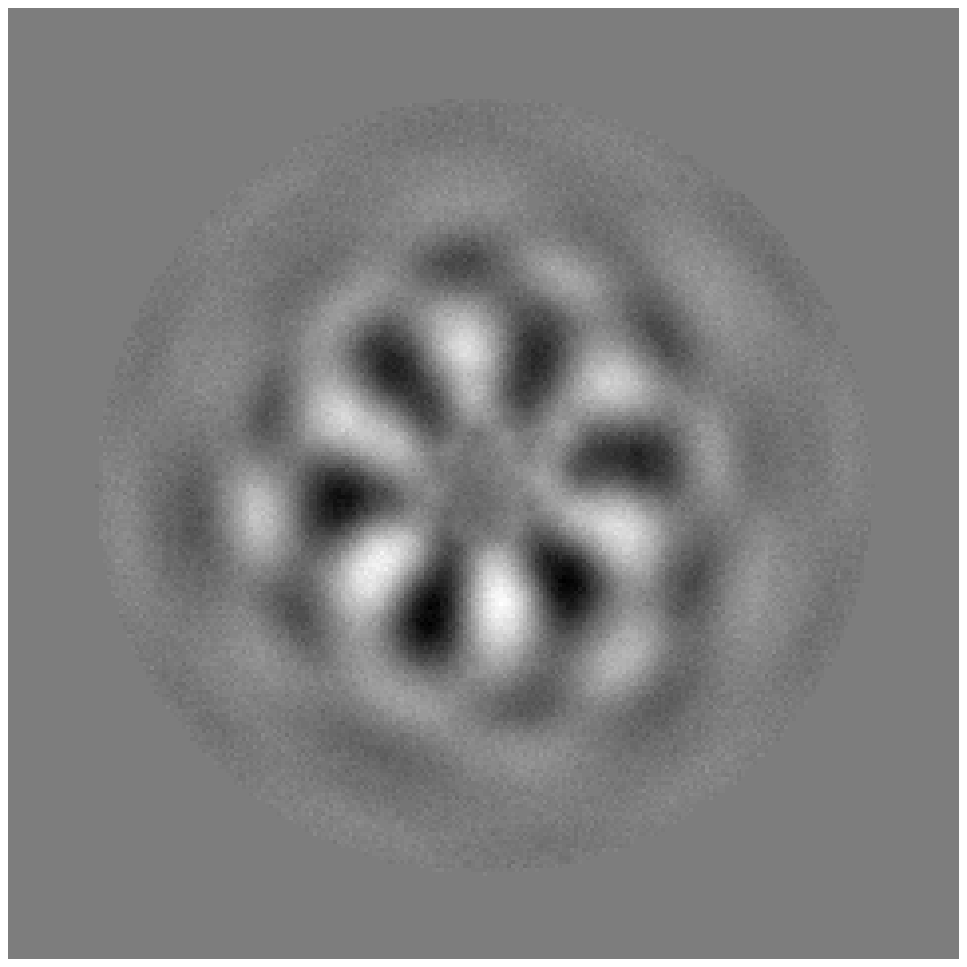} &
\includegraphics[width=0.2\columnwidth]{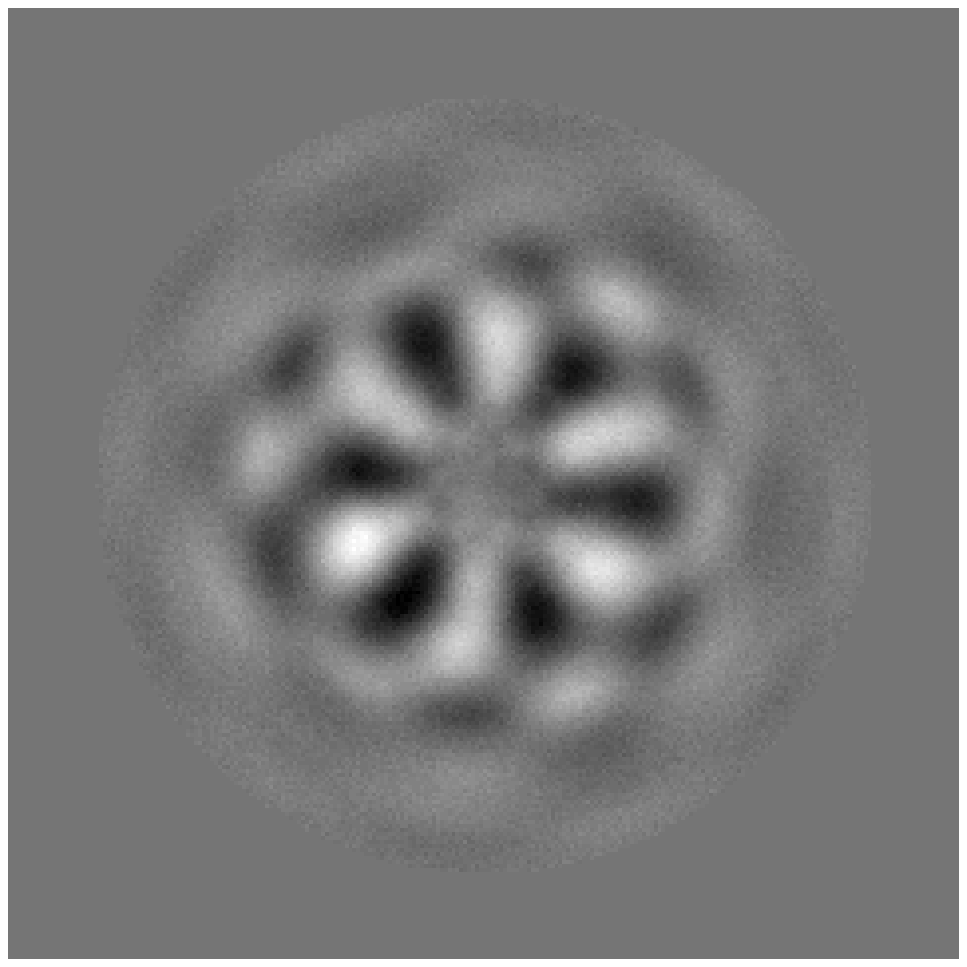} &
\includegraphics[width=0.2\columnwidth]{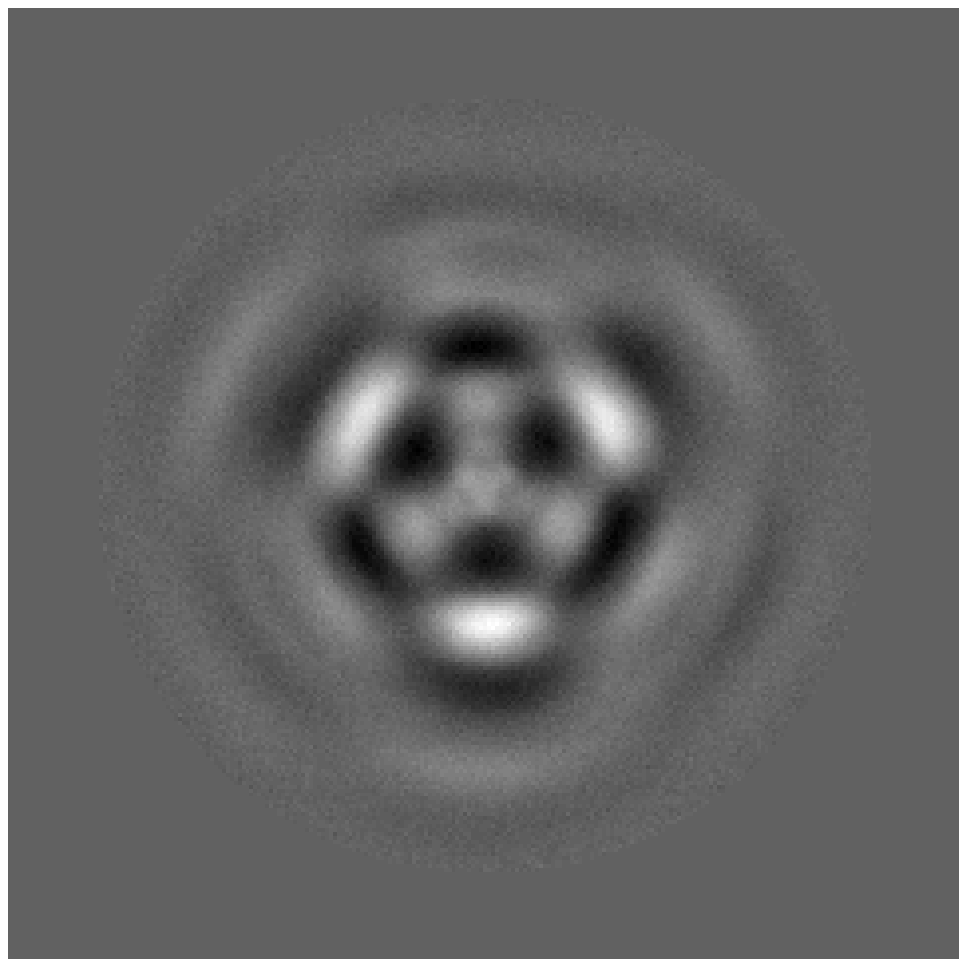} &
\includegraphics[width=0.2\columnwidth]{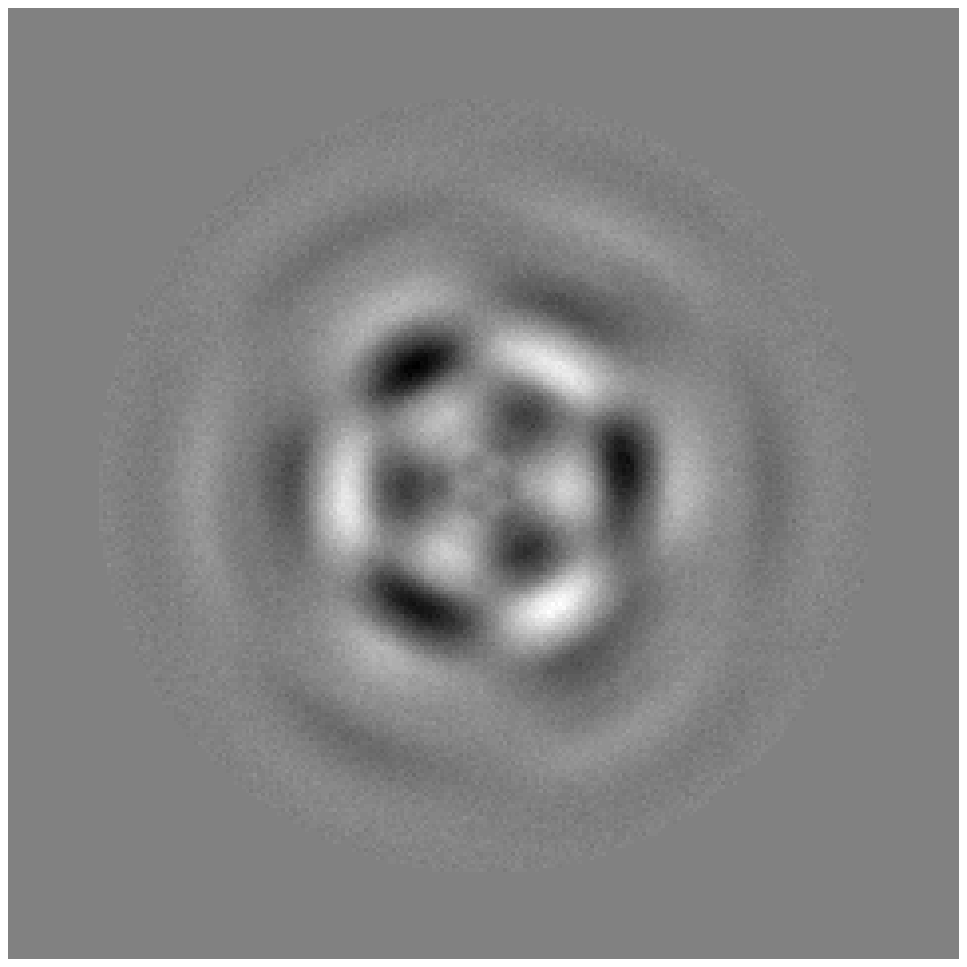} \\
$\lambda = 154.0 $ & $\lambda = 152.3$ & $\lambda = 148.5$ & $\lambda = 147.1$
\end{tabular}
\end{center}
\caption{Traditional PCA principal components in real image domain for the same dataset used in Figures~\ref{fig:spca_radial} and~\ref{fig:spca_basis_real}.}
\label{fig:pca_basis}
\end{figure}

We computed the traditional PCA and FBsPCA on the same dataset in real image domain (see Fig.~\ref{fig:pca_basis} for PCA components), which took 60 minutes and 16 minutes respectively. In order to compare the principal components computed by FFBsPCA with those computed by traditional PCA, we take the inverse Fourier transform of the FFBsPCA components. We do not compute the inverse polar Fourier transform directly, since such a transform is ill-conditioned. Instead, since the FFBsPCA components are linear combinations of the Fourier-Bessel functions as in Eq.~\eqref{eq:sPCArt}, we evaluate the steerable principal components on a Cartesian grid in real space using the linear combinations of $\mathcal{F}^{-1}(\psi_c^{k, q})$, given by Eq.~\eqref{eq:IFT_FB}. 
Those principal components are shown in Fig.~\ref{fig:spca_basis_real}.
Some of the top sixteen principal components computed from traditional PCA and FFBsPCA look very similar, for example, the first three and the last four principal components (see Fig.~\ref{fig:spca_basis_real} and Fig.~\ref{fig:pca_basis}). Because the gap between the eigenvalues of the traditional PCA is very small for the components in the middle two rows of Fig.~\ref{fig:pca_basis}, those components become degenerate and therefore look different from the corresponding components in Fig.~\ref{fig:spca_basis_real}.

\begin{figure}[htb]
\begin{center}
\subfloat[clean]{
\includegraphics[width=0.3\columnwidth]{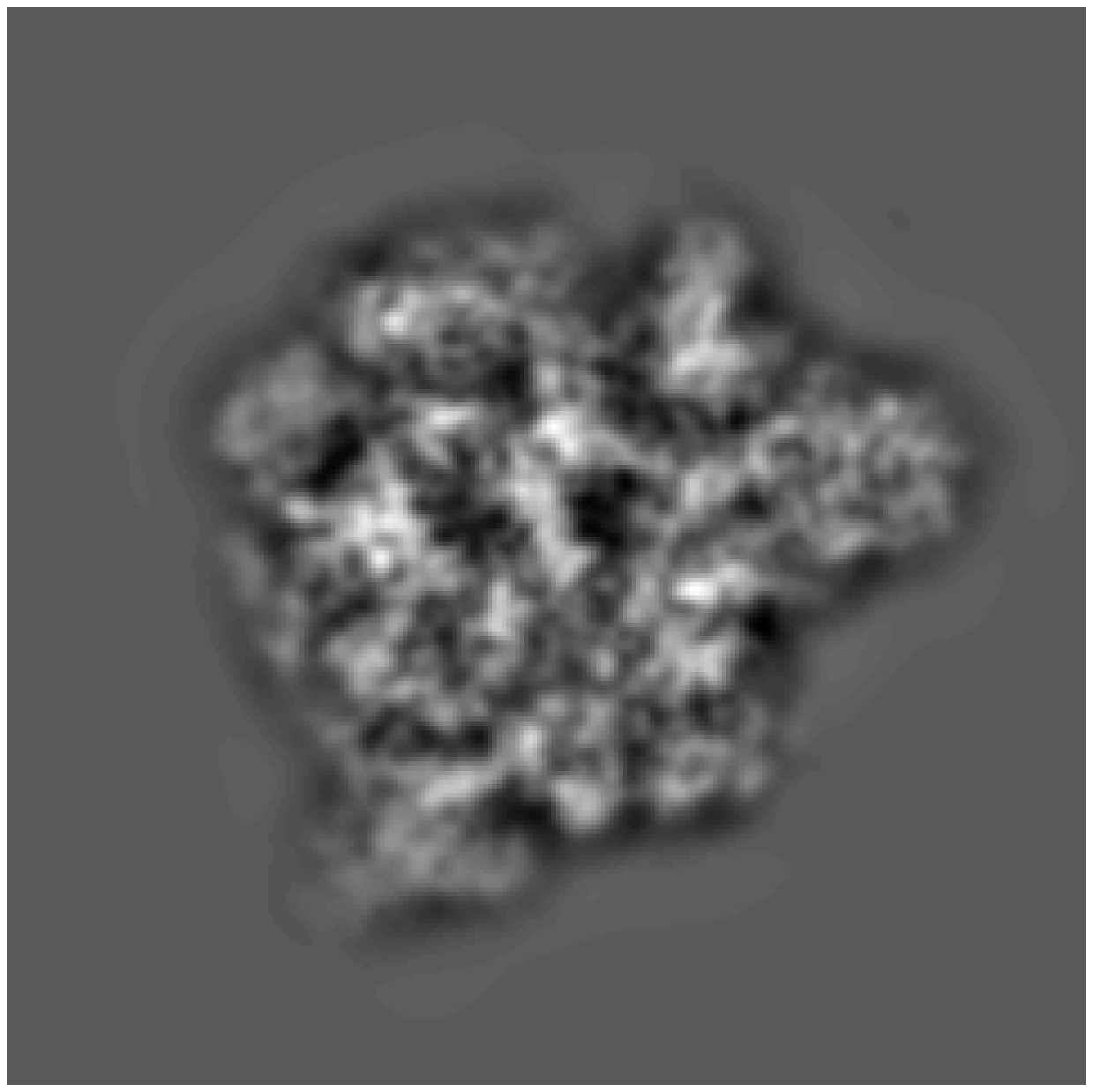}
\label{subfig:clean}
}
\subfloat[noisy]{
\includegraphics[width=0.3\columnwidth]{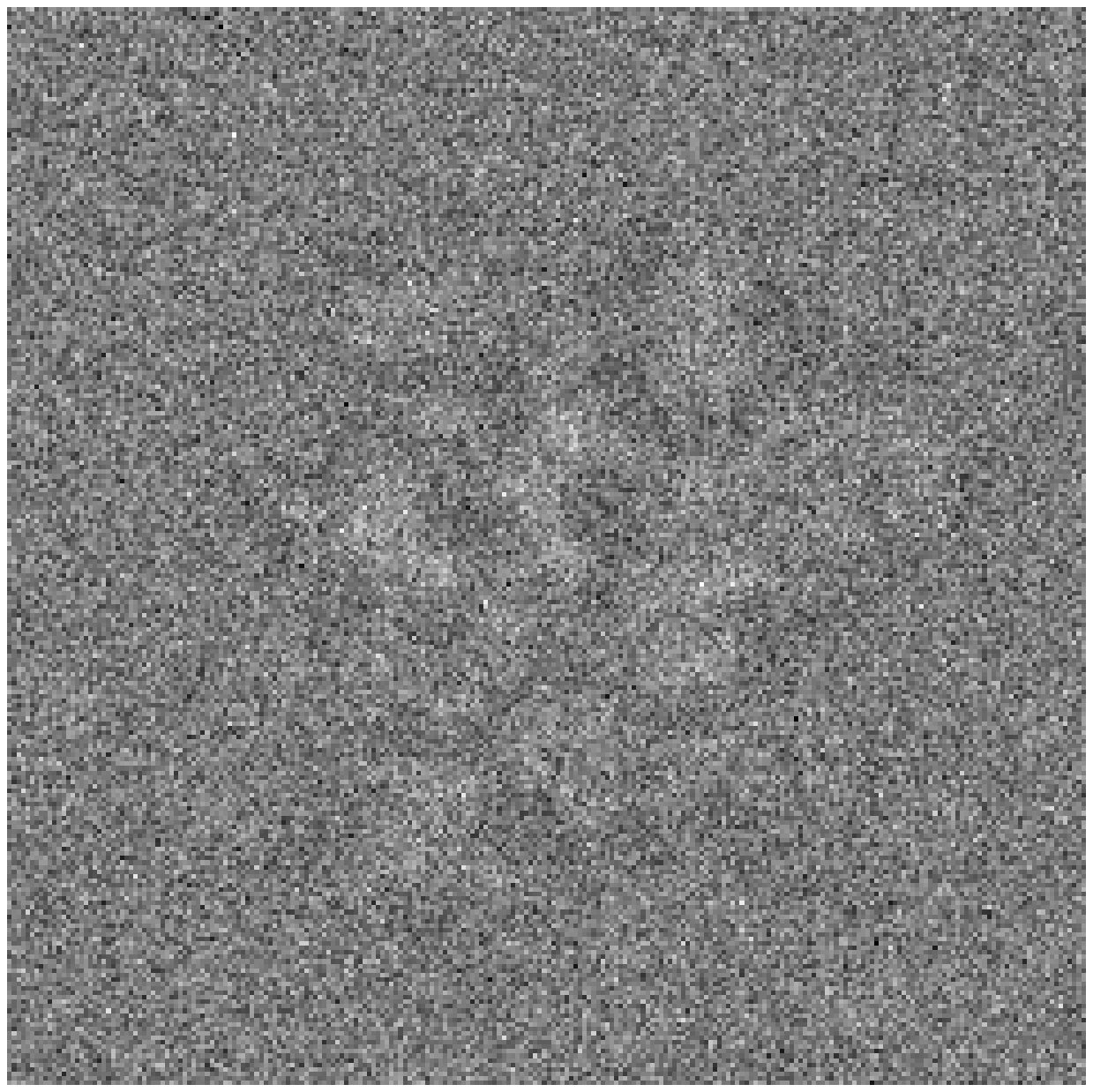}
\label{subfig:noisy}
}
\subfloat[PCA]{
\includegraphics[width=0.3\columnwidth]{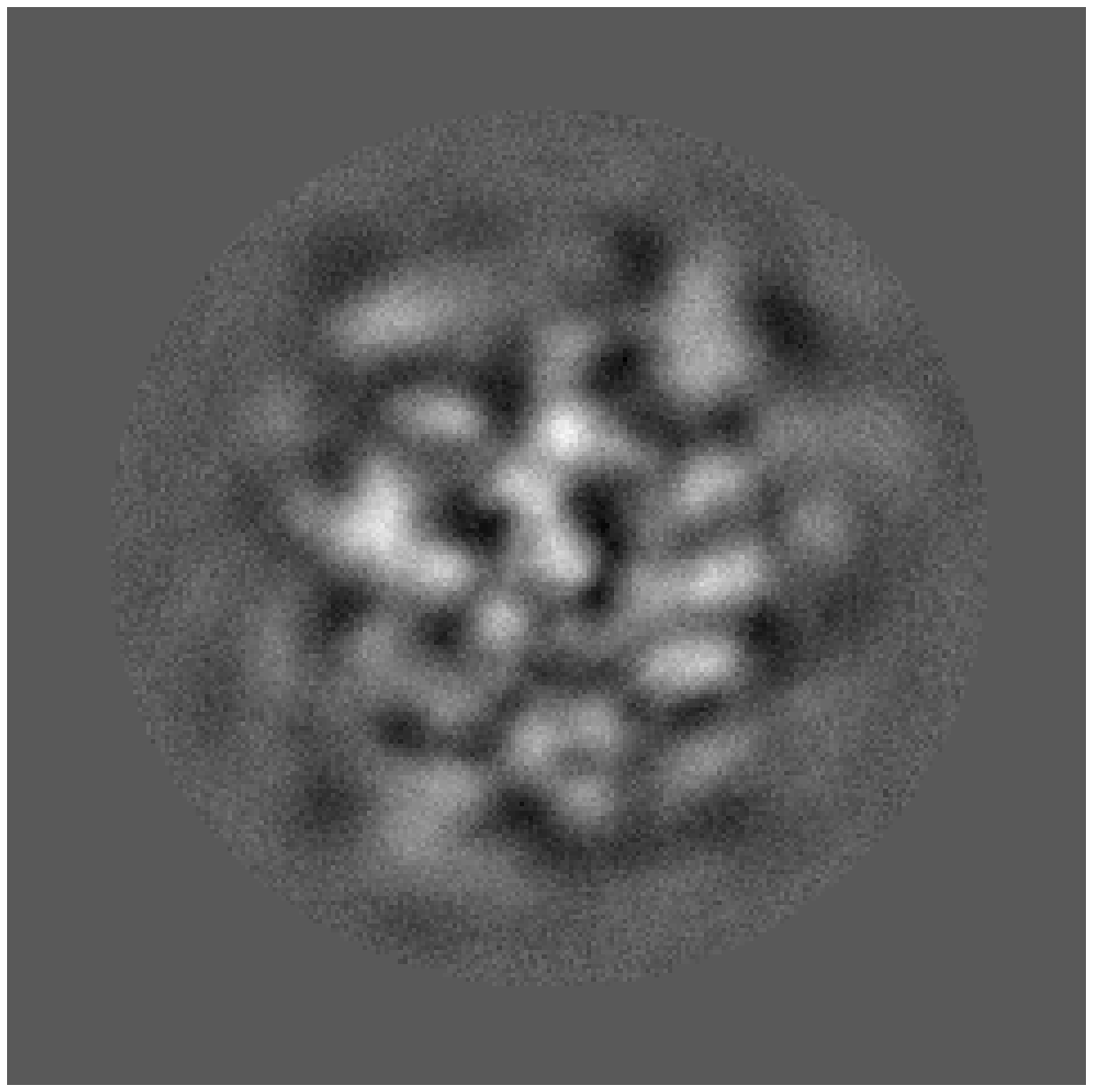}
\label{subfig:PCA}
}\\
\subfloat[Curvelet]{
\includegraphics[width=0.3\columnwidth]{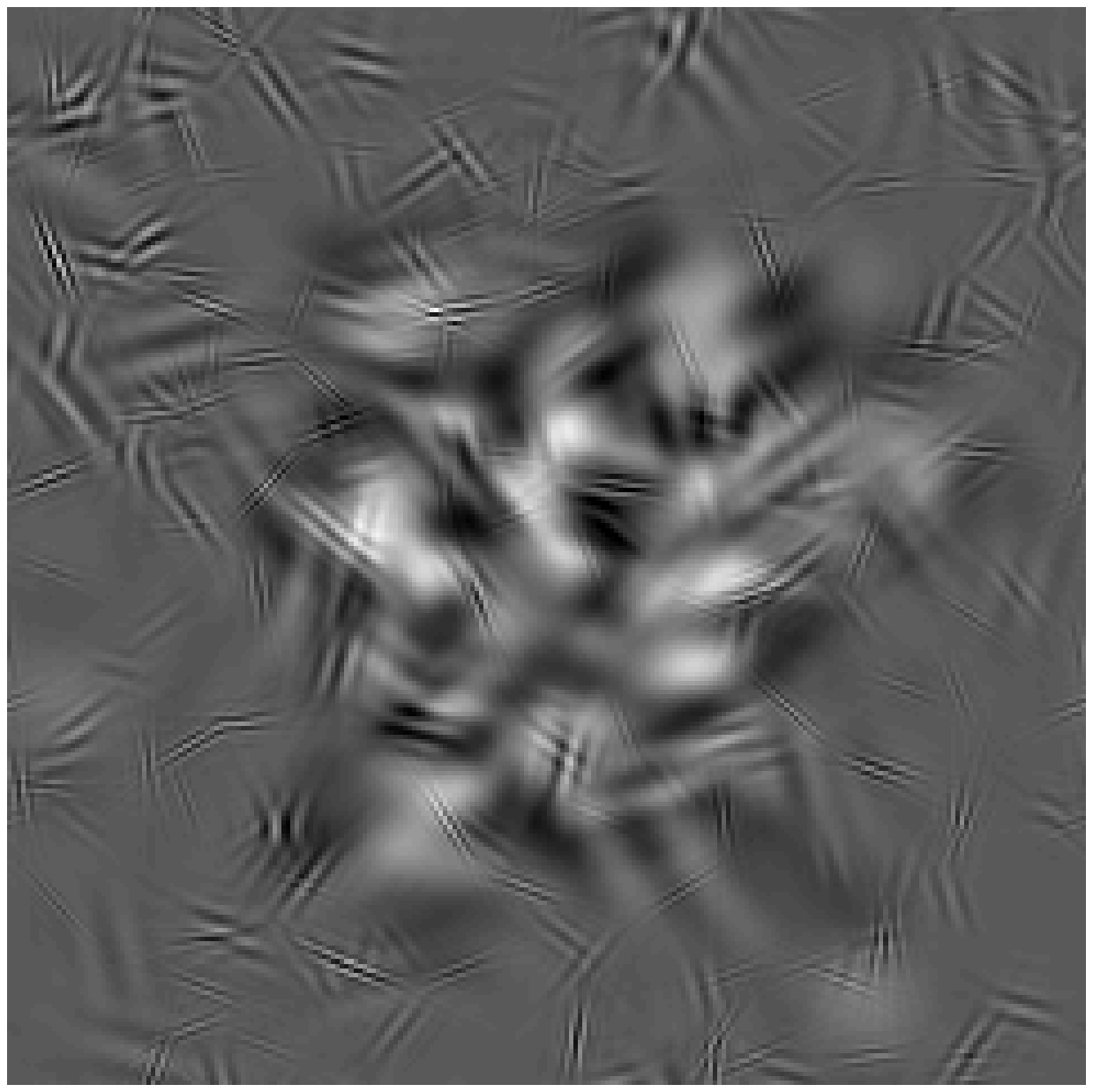}
\label{subfig:curvelet}
}
\subfloat[FBsPCA]{
\includegraphics[width=0.3\columnwidth]{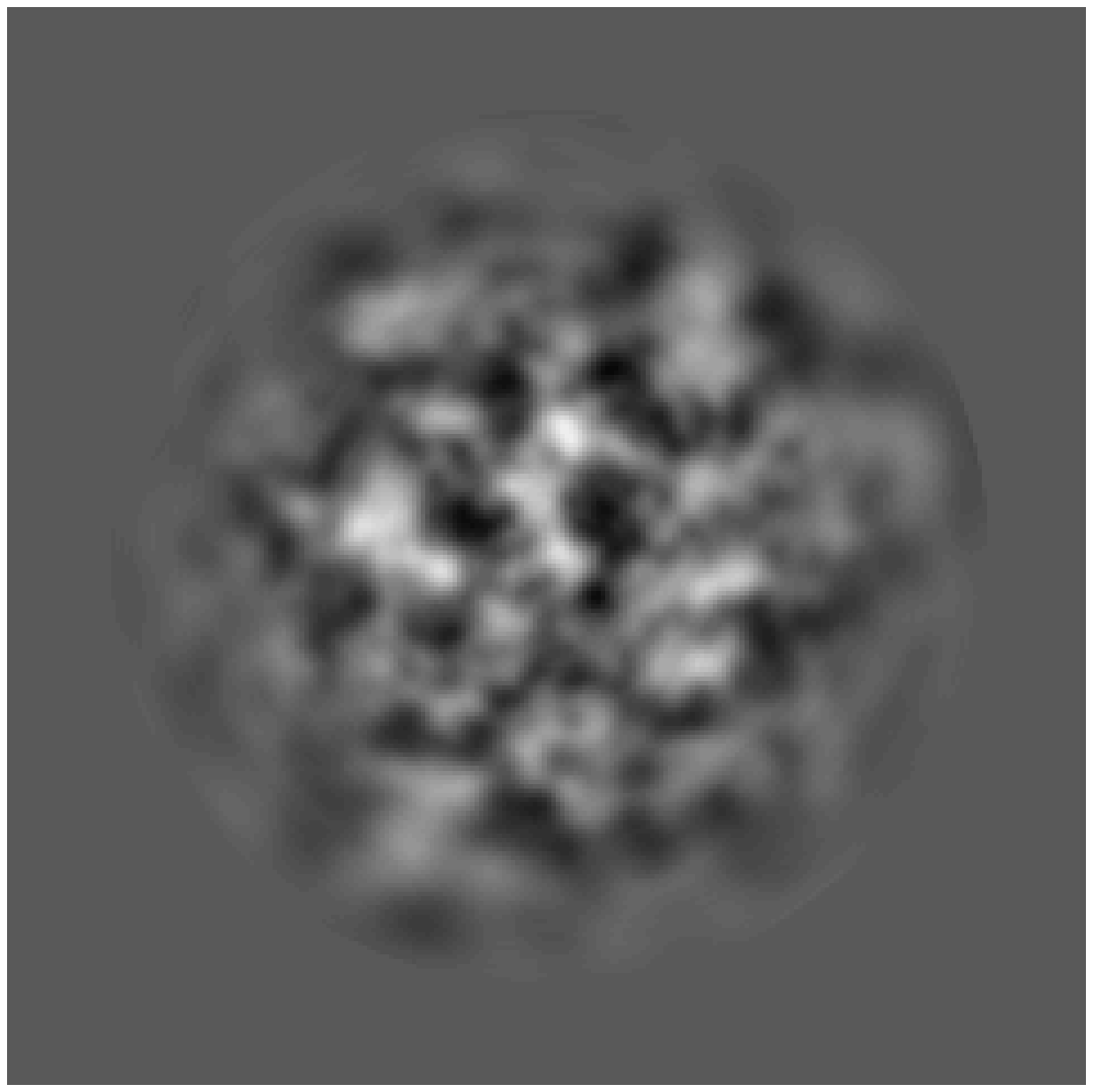}
\label{subfig:FBsPCA}
}
\subfloat[FFBsPCA]{
\includegraphics[width=0.3\columnwidth]{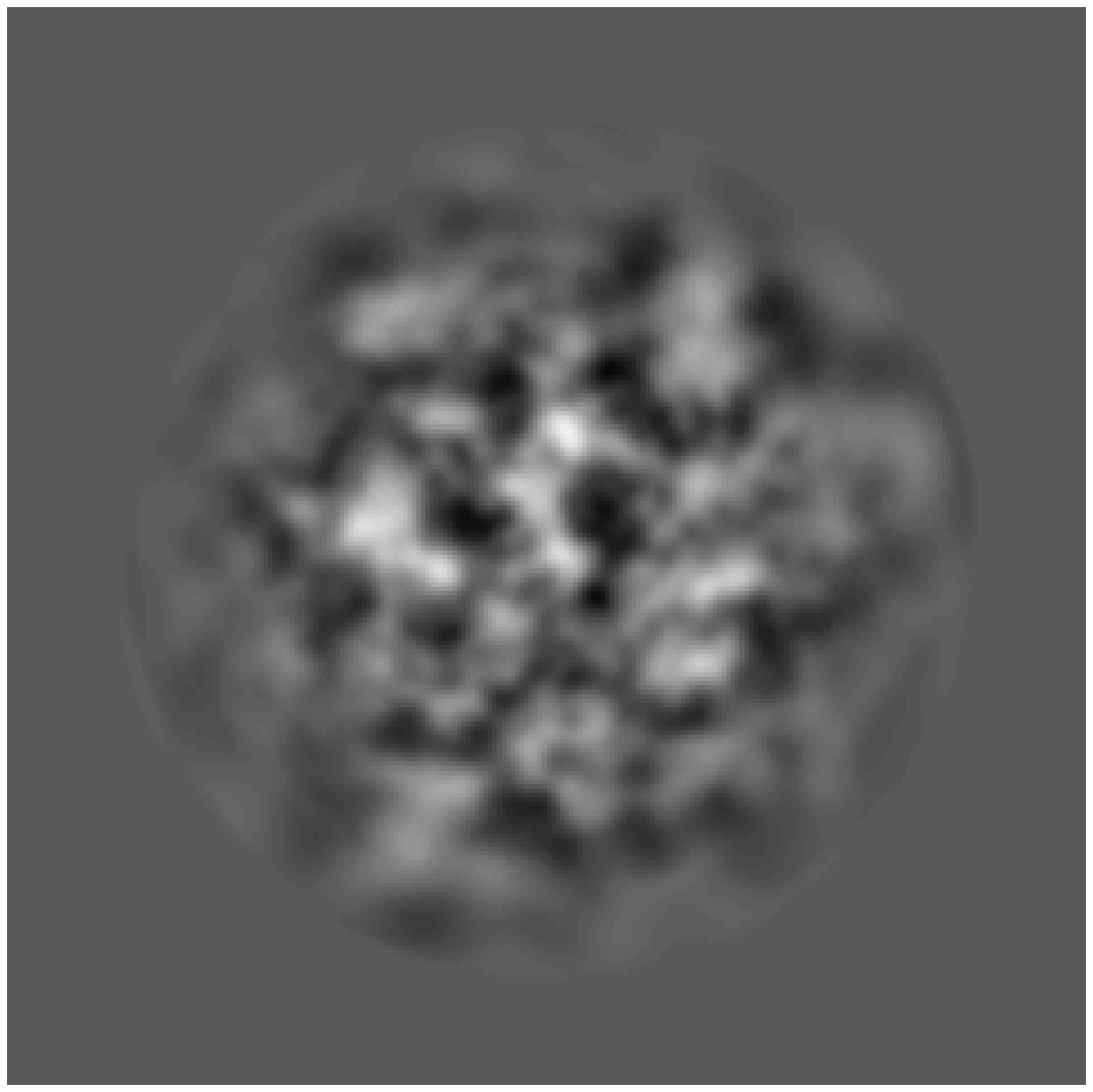}
\label{subfig:sPCA}
}
\end{center}
\caption{Denoising simulated projection images. \protect\subref{subfig:clean}~clean projection image, \protect\subref{subfig:noisy}~noisy projection image with SNR$=1/30$, \protect\subref{subfig:PCA}~denoised projection image using traditional PCA, \protect\subref{subfig:curvelet}~denoised projection image using Curvelet transform, complex block thresholding and cycle spinning, \protect\subref{subfig:FBsPCA}~denoised image using FBsPCA, and~\protect\subref{subfig:sPCA}~denoised image using FFBsPCA.}
\label{fig:denoise}
\end{figure}

\begin{figure}[htb]
\begin{center}
\subfloat[clean]{
\includegraphics[width=0.3\columnwidth]{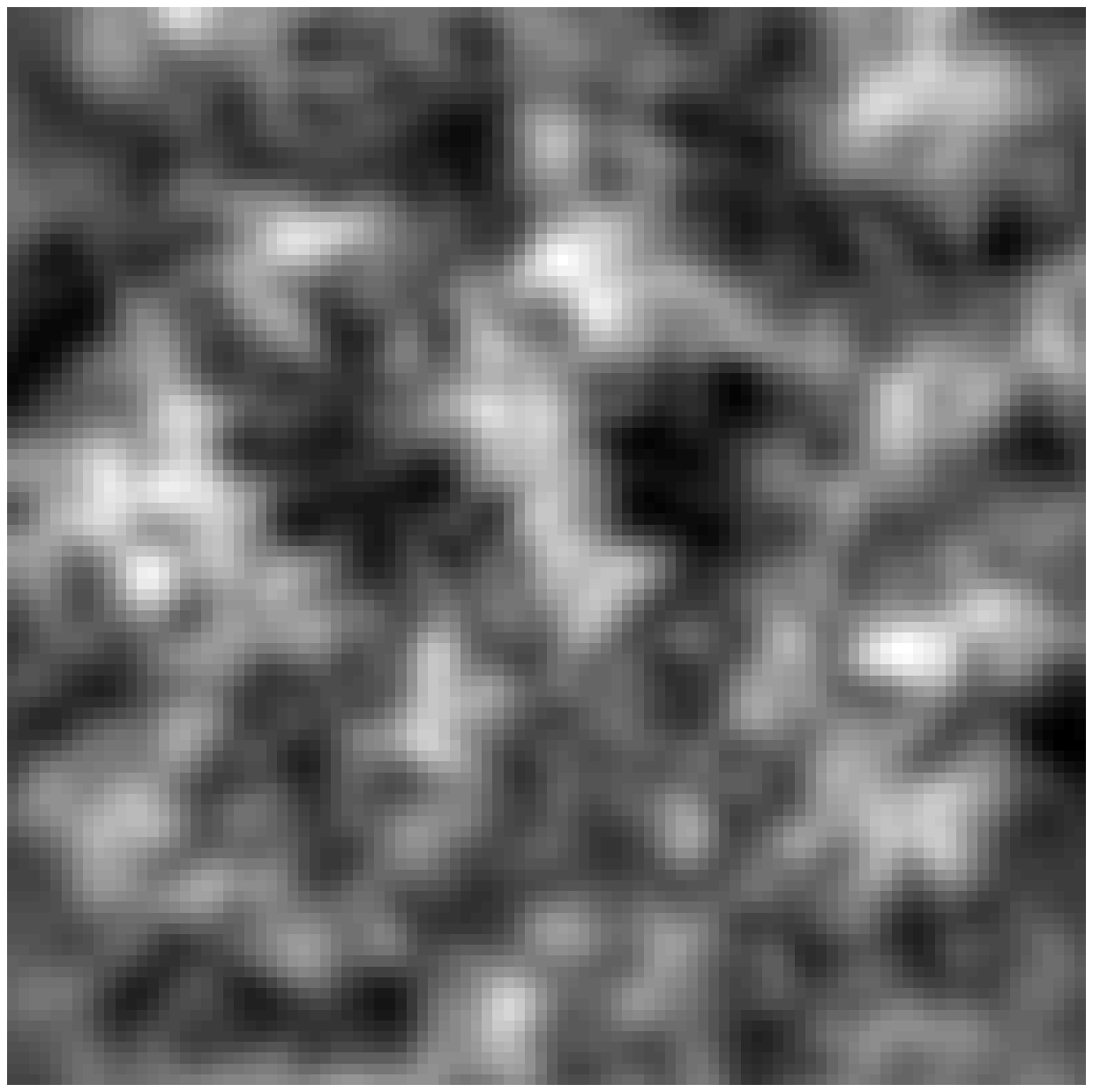}
\label{subfig:clean2}
}
\subfloat[noisy]{
\includegraphics[width=0.3\columnwidth]{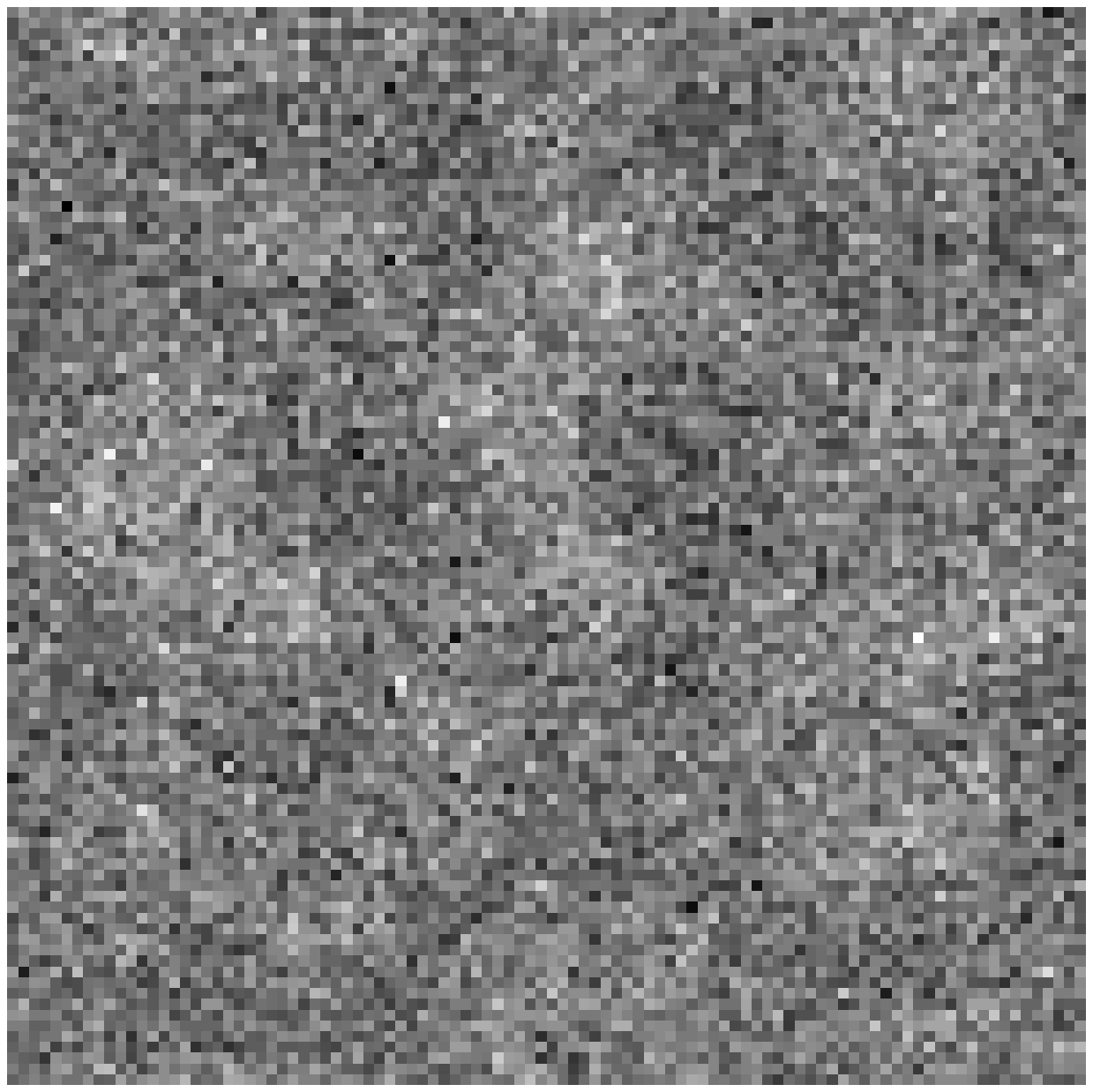}
\label{subfig:noisy2}
}
\subfloat[PCA]{
\includegraphics[width=0.3\columnwidth]{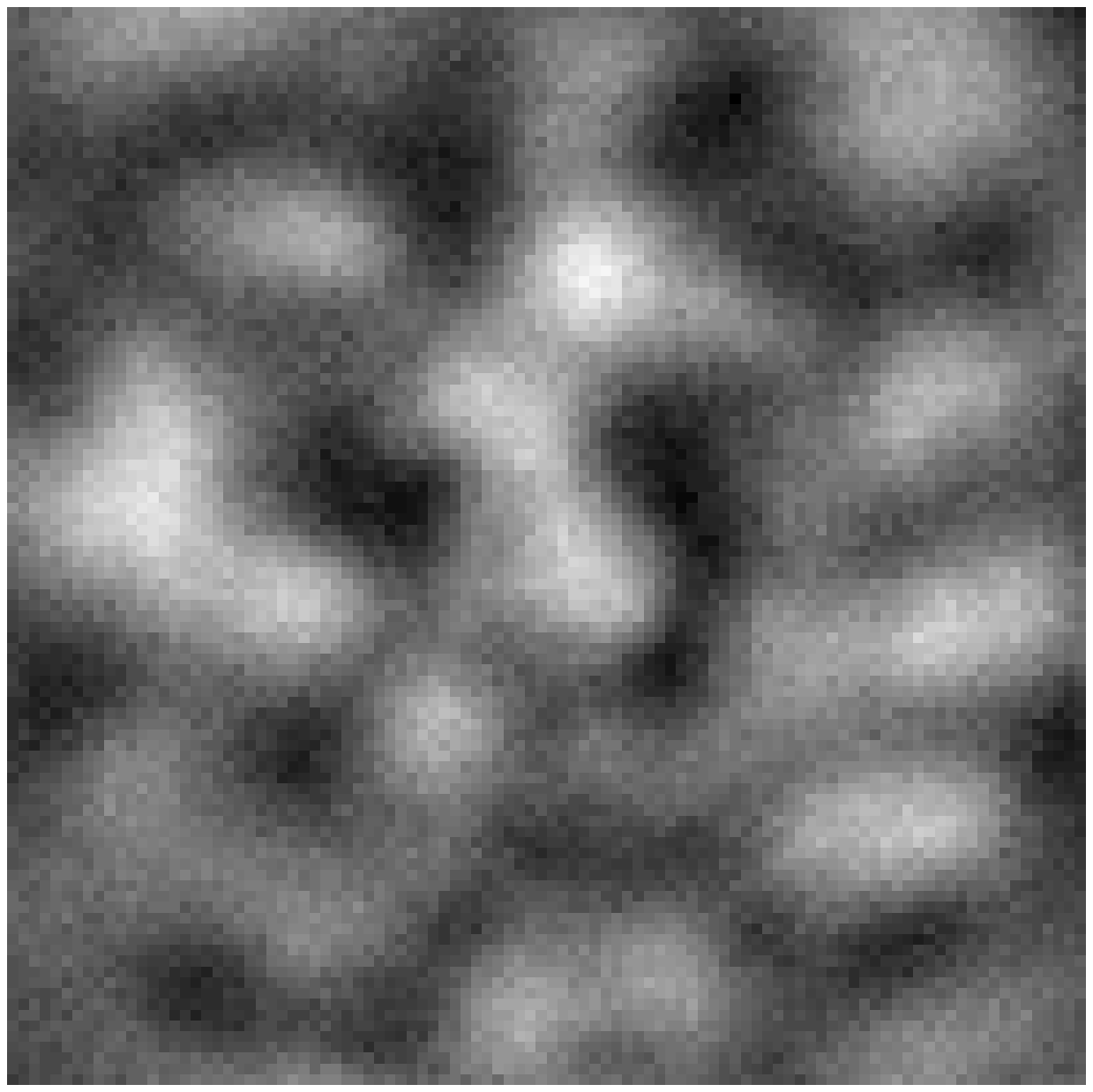}
\label{subfig:PCA2}
}\\
\subfloat[Curvelet]{
\includegraphics[width=0.3\columnwidth]{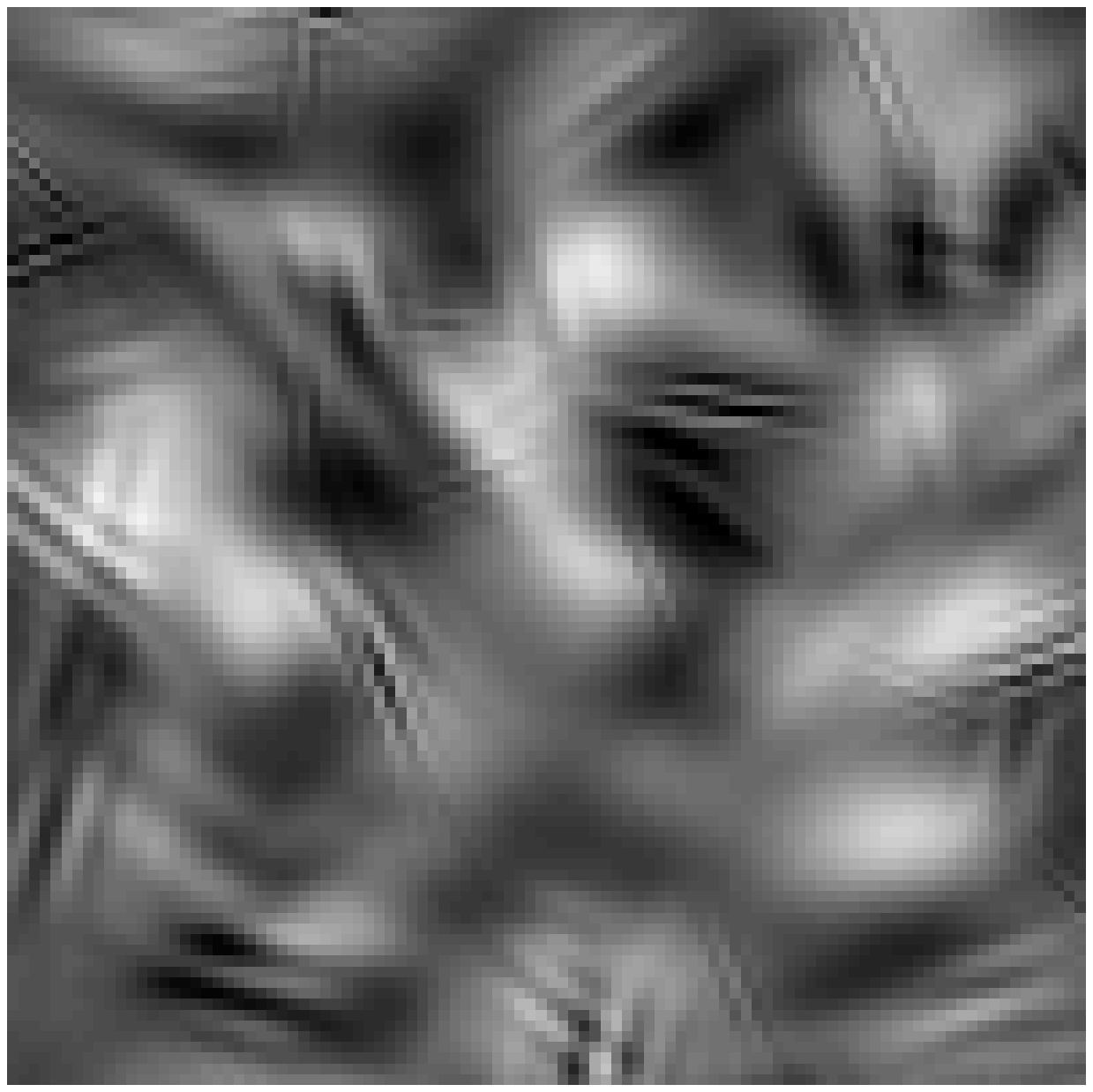}
\label{subfig:curvelet2}
}
\subfloat[FBsPCA]{
\includegraphics[width=0.3\columnwidth]{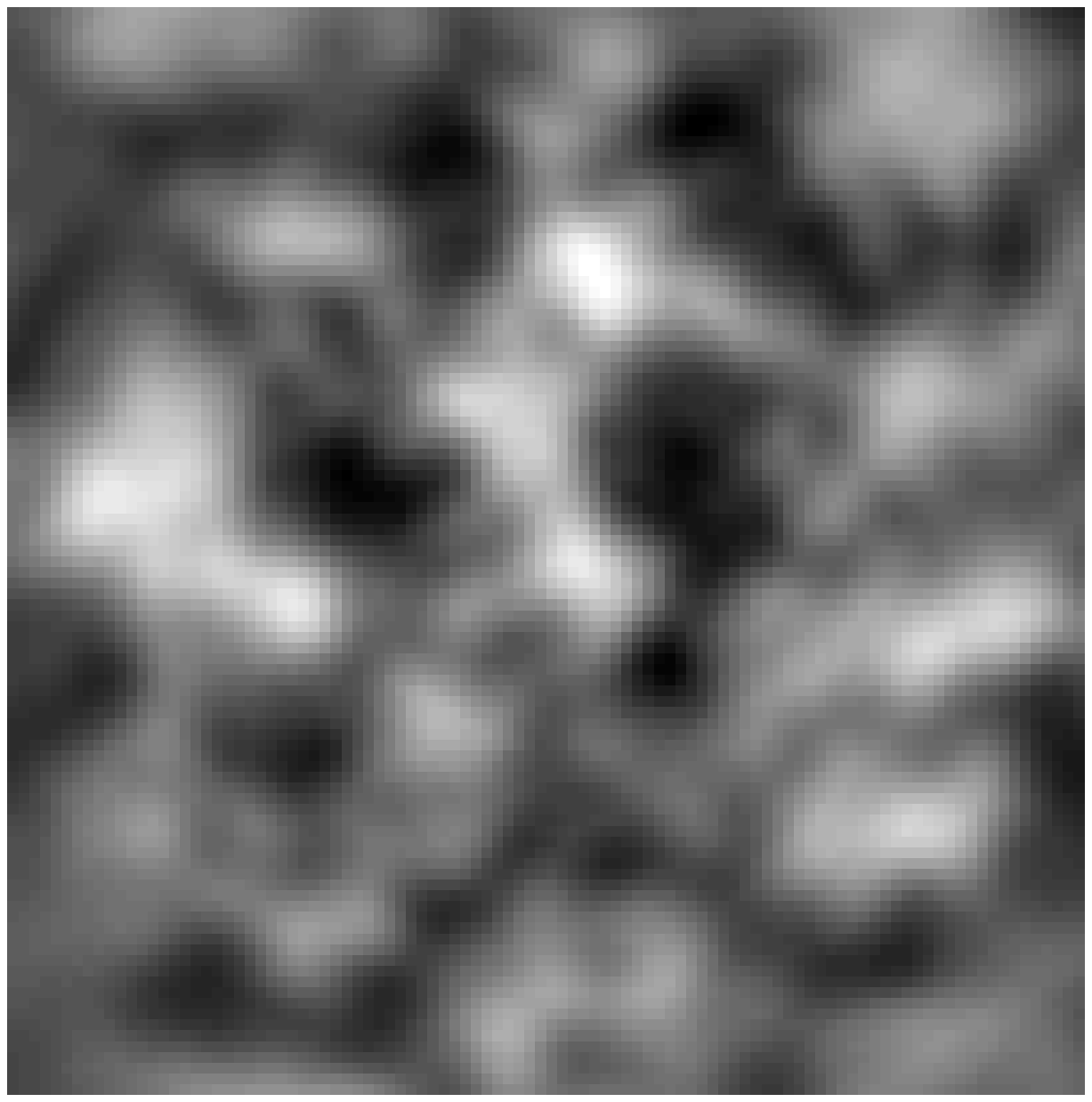}
\label{subfig:FBsPCA2}
}
\subfloat[FFBsPCA]{
\includegraphics[width=0.3\columnwidth]{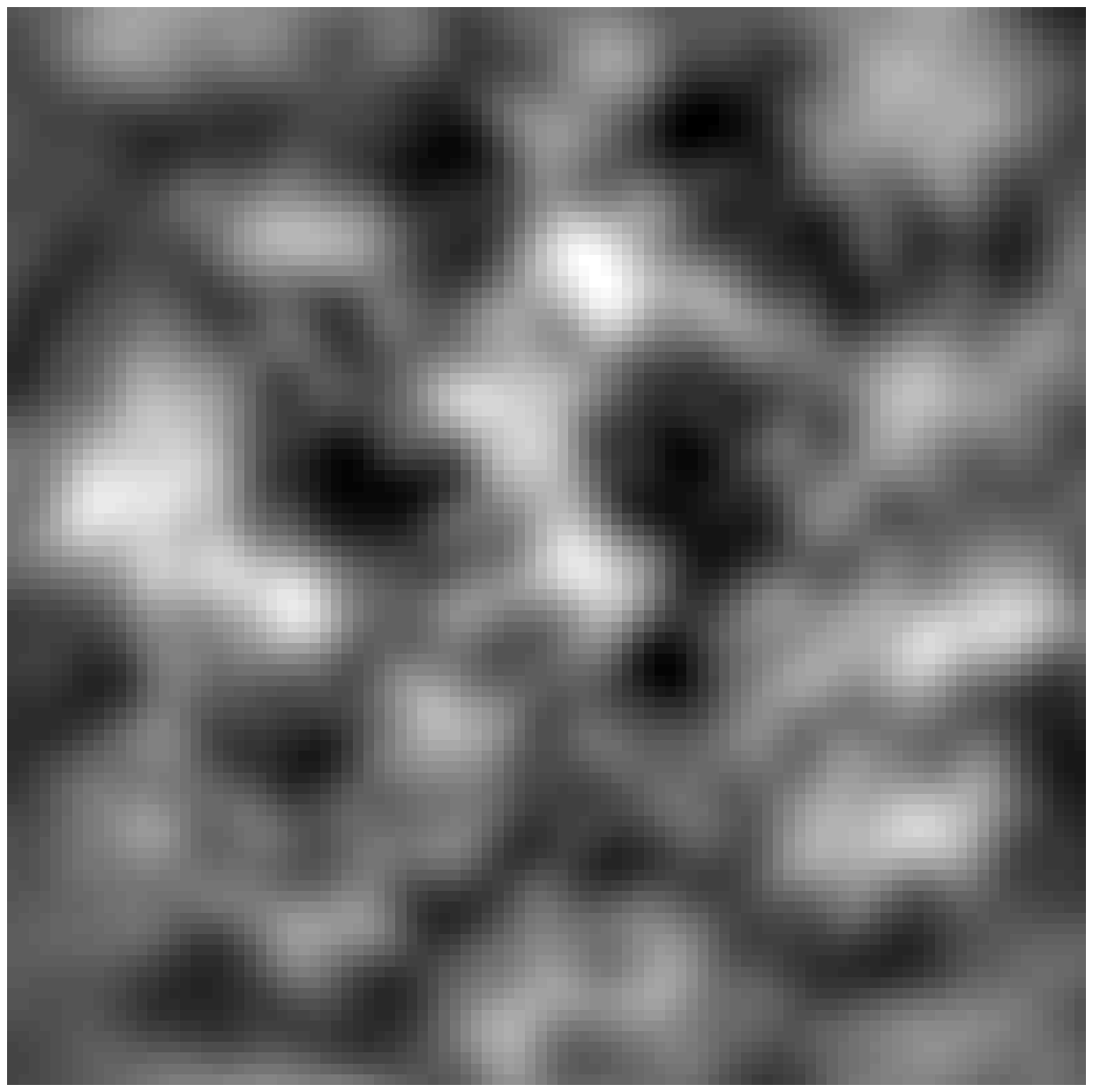}
\label{subfig:sPCA2}
}
\end{center}
\caption{Enlarged view of $100 \times 100$ pixels box at the center of the images in Figure~\ref{fig:denoise}.}
\label{fig:denoise2}
\end{figure}

In our simulation, each noisy projection image $I$ is obtained by contaminating the clean image $I_c$ with additive white Gaussian noise of variance $\sigma^2 = 9$. Given the noise level, we would like to automatically select the appropriate principal components to compress and denoise the noisy images. Since the transformation $T^*$ is nearly unitary, the coefficient matrices can be modeled approximately as $A^{(k)} = A^{(k)}_c + \epsilon^{(k)}$, where $\epsilon^{(k)}$ is white Gaussian noise with variance $\sigma^2$ and $A^{(k)}_c$ is the coefficient matrix for the clean images.  
In the case when there is no signal, that is $A^{(k)}_c = 0$, all eigenvalues of the covariance matrix $C^{(k)}$ from Eqs.~\eqref{eq:ck} and~\eqref{eq:ck_0} converge to $\sigma^2$ as $n$ goes to infinity, while $p_k$ is fixed. When $A^{(k)}_c \neq 0$, components with eigenvalues larger than $\sigma^2$ correspond to the underlying clean signal. In the non-asymptotic regime of a finite number of images, the eigenvalues of the sample covariance matrix from white Gaussian noise spread around $\sigma^2$. The empirical density of the eigenvalues can be approximated by the Mar{\v c}enko-Pastur distribution with parameter $\gamma_k$,  where $\gamma_0=\frac{p_0}{n}$ and $\gamma_k=\frac{p_k}{2n}$ for $k>0$ and the eigenvalues of $C^{(k)}$ are supported on $[ \lambda^{(k)}_- , \lambda^{(k)}_+]$, with $\lambda^{(k)}_\pm = \sigma^2(1 \pm \sqrt{\gamma_k})^2$. The principal components corresponding to eigenvalues larger than $\lambda^{(k)}_+$ correspond to signal information beyond noise level. Therefore, with the estimated noise variance $\hat{\sigma}^2$, we denote by $\lambda^{(k)}_1 \geq \lambda^{(k)}_2 \geq \dots \geq \lambda^{(k)}_{p_k}$ the eigenvalues of the covariance matrix $C^{(k)}$, and select the components with eigenvalues
\begin{equation}
\lambda^{(k)}_l > \hat{\sigma}^2(1+\sqrt{\gamma_k})^2, \quad l = 1, \ldots, p_k.
\label{eq:compselect}
\end{equation}
Various ways of selecting principal components from noisy data have been proposed.
We refer to~\cite{Kritchman} for an automatic procedure for estimating the noise variance and the number of components beyond the noise level.
For the simulated ribosomal subunit projections images, there are 966 steerable principal radial components above the threshold in Eq.~\eqref{eq:compselect}, whereas considerably fewer principal components (391) with the traditional PCA were selected.

Moreover, we filter the expansion coefficients to get better denoising. To first order approximation, when $n\gg p_k$, the noise simply shifts all eigenvalues upward by $\sigma^2$ and this calls for soft thresholding of the sample covariance eigenvalues: $(\lambda - \sigma^2)_+$. To correct for the finite sample effect, we can apply more sophisticated shrinkage to the eigenvalues, such as the methods proposed in~\cite{Wu13, Donoho14}. Specifically, we applied the shrinkage method in~\cite{Wu13} to the coefficients computed by FFBsPCA, FBsPCA, and PCA. Because we were able to use more principal components with FFBsPCA, we recovered finer details of the clean projection images, comparing Fig.~\ref{subfig:PCA2} and Fig.~\ref{subfig:sPCA2}.

\begin{figure}
\begin{center}
\includegraphics[width=0.8\columnwidth]{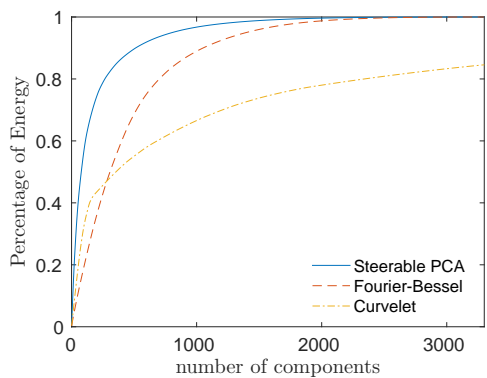}
\end{center}
\caption{Cumulative variance of FFBsPCA, Fourier-Bessel and Curvelet expansion coefficients of simulated clean ribosome projection images as in Fig.~\ref{fig:clean}.}
\label{fig:cumulative_variance}
\end{figure}

In addition to using data-adaptive bases, we also used a non-isotropic directional multiscale transform, i.e., Curvelet transform~\cite{Candes2005} with complex block thresholding and cycle spinning, to denoise the images. An example of a denoised image using PCA, Curvelet, FBsPCA, and FFBsPCA is shown in Fig.~\ref{fig:denoise}. The steerable PCA basis captures the variance of the clean dataset with fewer components than non-adaptive bases, such as Fourier-Bessel basis or Curvelets (see Fig.~\ref{fig:cumulative_variance}).
\begin{table}[htb]
\begin{center}
\begin{tabular}{|c|c|c|c|c|}
\hline
& \multicolumn{4}{ |c| }{MSE ($10^{-5}$)} \\
\hline
& Curvelet & PCA  & FBsPCA & FFBsPCA  \\
\hline
Image 1 & 1.38 & 1.10  & 0.77 & \textbf{0.77}  \\
\hline
Image 2 & 1.63 & 1.29  & \textbf{0.95} & 0.96 \\
\hline
Image 3 & 1.58 & 1.17  & 0.85 & \textbf{0.85} \\
\hline
\end{tabular}
\end{center}
\caption{
MSE of denoised images using PCA, Curvelets, FBsPCA and FFBsPCA, all computed using pixels within $R = 98$.}
\label{tab:denoise_a}
\end{table}

\begin{table}[htb]
\begin{center}
\begin{tabular}{|c|c|c|c|c|}
\hline
 & \multicolumn{4}{ |c| }{PSNR (dB)} \\
\hline
 & Curvelet & PCA & FBsPCA & FFBsPCA  \\
\hline
Image 1   & 18.10  & 19.06 & 20.62 & \textbf{20.63} \\
\hline
Image 2  & 17.90 & 18.93  & \textbf{20.26} & 20.23\\
\hline
Image 3   & 18.68  & 19.99& 21.35 & \textbf{21.35}\\
\hline
\end{tabular}
\end{center}
\caption{PSNR of denoised images using PCA, Curvelets, FBsPCA and FFBsPCA, all computed using pixels within $R = 98$.}
\label{tab:denoise_b}
\end{table}
We computed the mean squared error (MSE) and Peak SNR (PSNR) to quantify the denoising effects in Tab.~\ref{tab:denoise_a} and Tab.~\ref{tab:denoise_b}. Comparing with the traditional PCA, FFBsPCA reduced the MSE by more than $25\%$ and increased the PSNR by over 1.3 dB. When the images are of low SNR, Curvelets are unable to outperform data adaptive bases, such as PCA, FBsPCA and  FFBsPCA (see Tab.~\ref{tab:denoise_a} and Tab.~\ref{tab:denoise_b}). This experiment shows that FFBsPCA is an efficient and effective procedure for denoising large image datasets.

\begin{table}[h!]
\begin{center}
\begin{tabular}{|c|c|c|c|c|c|c|}
\hline
 \multicolumn{2}{ |c|}{}& \multicolumn{3}{ |c| }{PSNR (dB)} \\
\hline
  max shifts (pixels) &  R (pixels) &  Image 1 &  Image 2  & Image 3   \\
\hline
0  & 98 & 21.61 & 21.16 & 22.34 \\
\hline
5  & 99 & 21.53 & 21.26 & 22.32 \\
\hline
10 & 102 & 21.42 & 21.13 & 22.11 \\
\hline
15 & 107 & 21.59 & 21.31 & 22.18 \\
\hline
20 & 110 & 21.31 & 21.30 & 22.07 \\
\hline
\end{tabular}
\end{center}
\caption{FFBsPCA denoising of images with maximum shifts of 0, 5, 10, 15, and 20 pixels. PSNRs are computed with pixels within $R = 110$. The estimated compact support $R$ increases with maximum shift.}
\label{tab:denoise_shifted}
\end{table}
Finally, we show that steerable PCA denoising is robust to small shifts. We simulated clean data with random shifts in the $\pm x$ and $\pm y$ directions with maximum shifts equal to $0$ (centered images), $5$, $10$, $15$, and $20$ pixels. The clean images 
are corrupted with additive white Gaussian noise of variance~9. As shown in Tab.~\ref{tab:denoise_shifted}, the denoising performance using FFBsPCA (measured in PSNR) is almost unaffected. The denoising results for centered images in Tab.~\ref{tab:denoise_b} and Tab.~\ref{tab:denoise_shifted} are slightly different because we used different support sizes to evaluate PSNRs.



\section{Conclusion}
\label{sec:concl}
In this paper we presented a fast Fourier-Bessel steerable PCA method that reduces the computational complexity with respect to the size of the images so that it can handle larger images. The complexity of the new algorithm is $O(nL^3 + L^4)$ compared with $O(nL^4)$ of the steerable PCA introduced in~\cite{Zhao13}. The key improvement is through mapping the images to a polar Fourier grid using NUFFT and evaluating the Fourier-Bessel expansion coefficients by angular 1D FFT and accurate radial integration. 

This work has been mostly motivated by its application to cryo-EM single particle reconstruction. Besides compression and denoising of the experimental images required for 2D class averaging~\cite{Zhao14} and common-lines based 3D ab-initio modeling, FFBsPCA can also be applied in conjunction with Kam's approach~\cite{Kam80} that requires the covariance matrix of the 2D images~\cite{Bhamre15}. The method developed here can also be extended to perform fast principal component analysis of a set of 3D volumes and their rotations. For this purpose, the Fourier-Bessel basis is replaced with the spherical-Bessel basis, and the expansion coefficients can be evaluated by performing the angular integration using a fast spherical harmonics transform~\cite{Tygert06} followed by radial integration.

Our numerical experiments show that an adaptive basis is necessary for denoising images with very low SNR. Steerable PCA is able to recover more signal components than PCA and achieves better denoising results. It is definitely possible to improve the denoising obtained by just using steerable PCA. For example, we can have more sophisticated dictionary denoising schemes, in which part of the dictionary is made of the steerable principal components and another part of the dictionary is made of wavelets. As these methods require the computation of steerable PCA, computing steerable PCA fast would be useful also for more advanced denoising schemes.

Finally, we remark that the Fourier-Bessel basis can be replaced in our framework with other suitable bases, for example, the 2D prolate spheroidal wave functions (PSWF) on a disk \cite{Slepian}. The 2D prolates also have a separation of variables form which makes them convenient for steerable PCA. A possible advantage of using 2D prolates is that they are optimal in terms of the size of their support.

\section*{Acknowledgment}
This research was supported by Award Number R01GM090200 from the NIGMS (A. S. and Y. S.), Award Number LTR DTD 06-05-2012 from the Simons Foundation (A. S.), the Moore Foundation DDD investigator award (A. S.), and Grant Number 578/14 from the Israel Science Foundation (Y. S.). The authors would like to thank Leslie Greengard, Michael O'Neil, and Alex Townsend for valuable discussions.

\ifCLASSOPTIONcaptionsoff
  \newpage
\fi

\bibliographystyle{IEEEtran}
\bibliography{FBsPCA}

%

\begin{IEEEbiography}[{\includegraphics[width=1in,height=1.25in,clip,keepaspectratio]{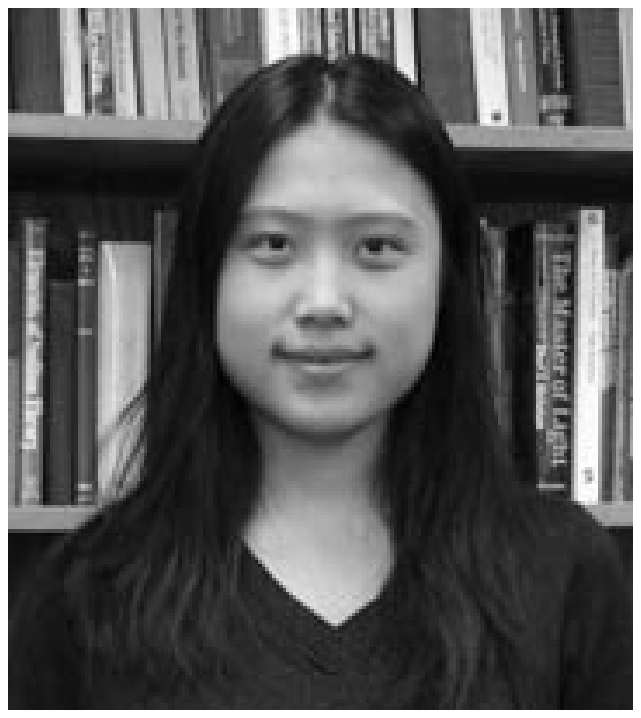}}]{Zhizhen Zhao}
received her B.A. and M.Sc. degrees in Physics from Trinity College, Cambridge University in 2008. Zhao received her  Ph.D degree in Physics from Princeton University in 2013. Since September 2013, she has been with the Courant institute of Mathematical Sciences at New York University. She is currently a Courant Instructor at NYU. Her research interests include applied and computational harmonic analysis, signal processing, data analysis and the applications in structural biology and atmospheric and oceanic sciences. 
\end{IEEEbiography}

\begin{IEEEbiography}[{\includegraphics[width=1in,height=1.25in,clip,keepaspectratio]{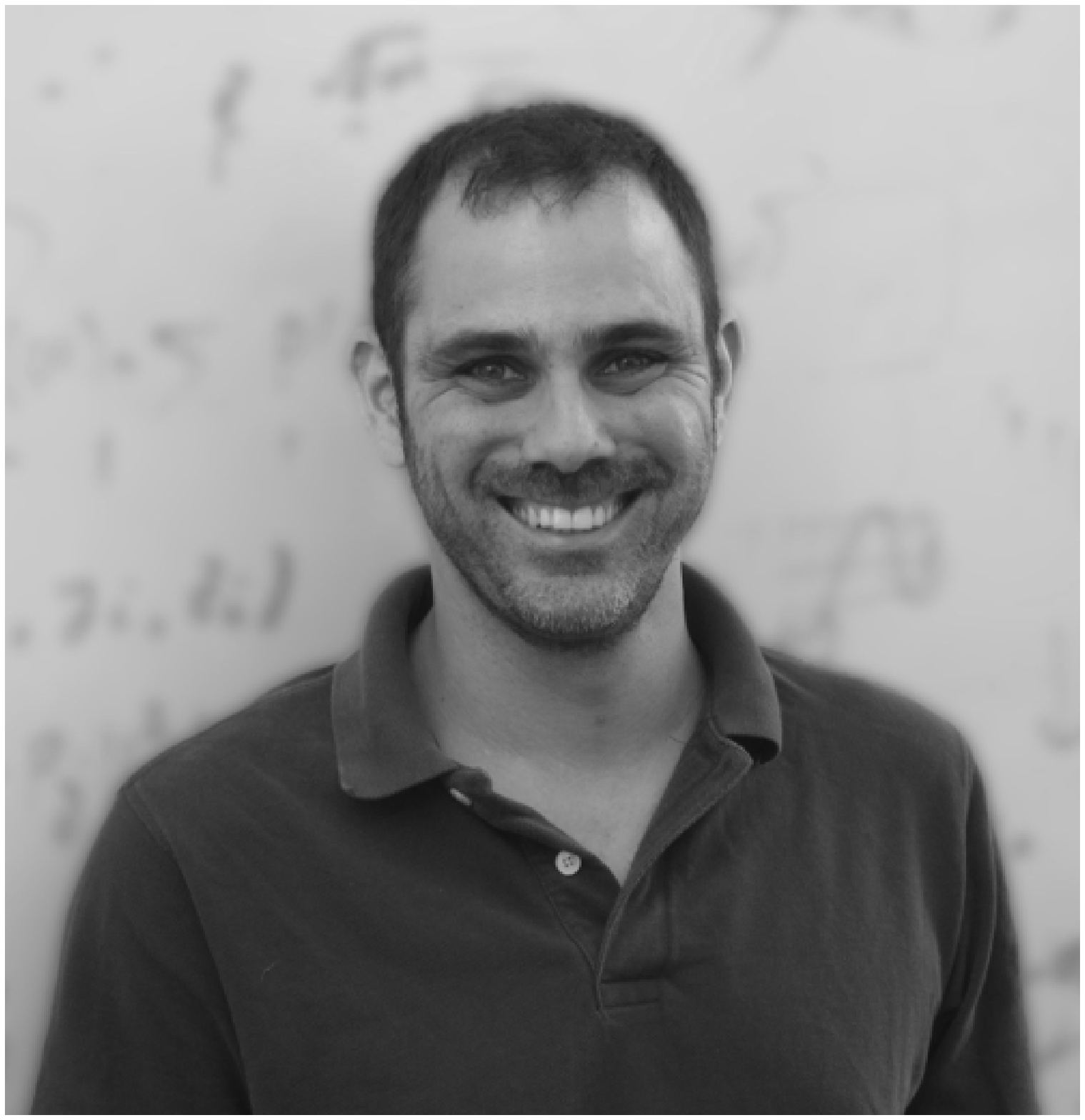}}]{Yoel Shkolnisky}
received his B.Sc. degree in mathematics and computer science and his M.Sc. and Ph.D. degrees in computer science, from Tel-Aviv University, Tel-Aviv, Israel, in 1996, 2001, and 2005, respectively. From July 2005 to July 2008, he was a Gibbs Assistant Professor in applied mathematics at the Department of Mathematics, Yale University, New Haven, CT. Since October 2009 he has been with the Department of Applied Mathematics, School of Mathematical Sciences, Tel-Aviv University. His research interests include computational harmonic analysis, scientific computing, and data analysis.
\end{IEEEbiography}

\begin{IEEEbiography}[{\includegraphics[width=1in,height=1.25in,clip,keepaspectratio]{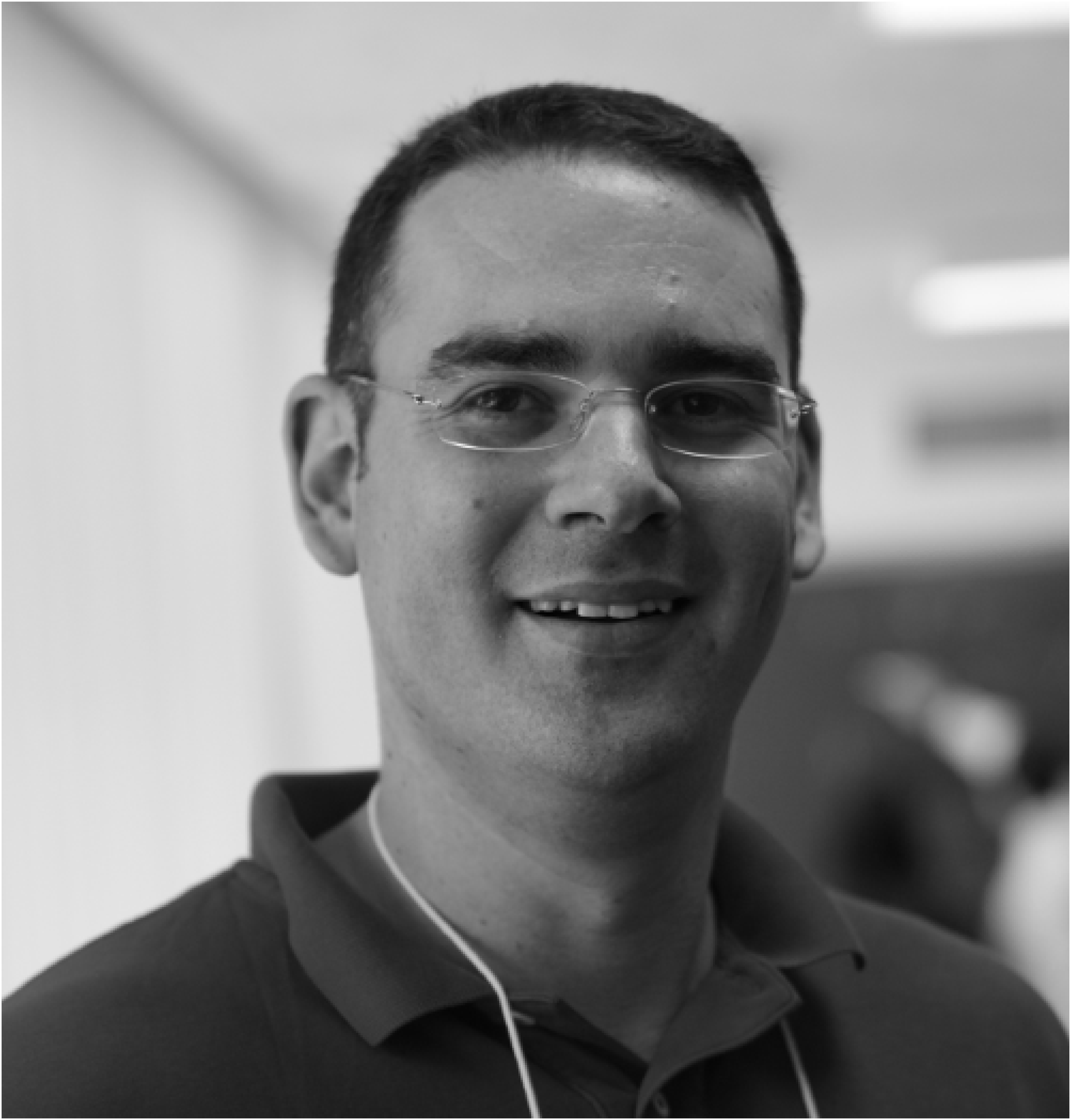}}]{Amit Singer}
is a Professor of Mathematics and a member of the Executive Committee of the Program in Applied and Computational Mathematics (PACM) at Princeton University.
He joined Princeton as an Assistant Professor in 2008. From 2005 to 2008 he was a
Gibbs Assistant Professor in Applied Mathematics at the Department of Mathematics,
Yale University. Singer received the BSc degree in Physics and Mathematics and the
PhD degree in Applied Mathematics from Tel Aviv University (Israel), in 1997 and
2005, respectively. He served in the Israeli Defense Forces during 1997-2003. He was
awarded the Moore Investigator in Data-Driven Discovery (2014), the Simons
Investigator Award (2012), the Presidential Early Career Award for Scientists and
Engineers (2010), the Alfred P. Sloan Research Fellowship (2010) and the Haim
Nessyahu Prize for Best PhD in Mathematics in Israel (2007). His current research in
applied mathematics focuses on theoretical and computational aspects of data
science, and on developing computational methods for structural biology.
\end{IEEEbiography}








\end{document}